\documentclass{article}

\PassOptionsToPackage{numbers, compress}{natbib}

\usepackage[preprint]{neurips_2026}

\usepackage{array}
\usepackage[utf8]{inputenc}
\usepackage[T1]{fontenc}
\usepackage{hyperref}
\usepackage{url}
\usepackage{booktabs}
\usepackage{amsfonts}
\usepackage{nicefrac}
\usepackage{microtype}
\usepackage{xcolor}
\definecolor{reviewtext}{rgb}{0.2,0.2,0.2}
\usepackage{graphicx}
\usepackage{subcaption}
\usepackage{booktabs}
\usepackage{float}
\usepackage{multirow}
\usepackage{tikz}
\usepackage{lmodern}

\usetikzlibrary{shapes.geometric, arrows.meta, positioning, calc, fit, backgrounds}

\usepackage{etoolbox}

\usepackage[most]{tcolorbox}
\usepackage{textalpha}
\tcbuselibrary{listings,breakable}
\usepackage{cuted}

\title{PRAIB: Peer Review AI Benchmark of Behaviour of LLM-Assisted Reviewing}

\author{
  Krzysztof Żurawicki$^{\star}$ \quad 
  Julia Farganus$^{\star}$ \quad 
  Arkadiusz Gaweł$^{\star}$ \And 
  Mateusz Bystroński \quad 
  Tomasz Jan Kajdanowicz \\
  \addlinespace[0.5em]
  Department of Artificial Intelligence \\
  Wrocław University of Science and Technology \\
  \addlinespace[0.3em]
  \texttt{\{krzysztof.zurawicki, mateusz.bystronski\}@pwr.edu.pl} \\
  \addlinespace[0.5em]
  \small $^{\star}$Equal contribution.
}

\begin{document}

\maketitle

\begin{abstract}
The growing number of submitted papers has motivated the exploration of Large Language Models (LLMs) as a means to support and augment the peer review process, particularly in terms of improving its speed and scalability.
Yet, it remains unknown whether LLMs \emph{engage} with scientific manuscripts in the same manner as human reviewers, or whether they merely produce review-looking text.
To address this, we introduce the Peer Review AI Benchmark (PRAIB), a novel framework comprising thoroughly defined metrics that measure review specificity, style, and behavior of engagement.
To complement the PRAIB framework, we conduct a large-scale empirical study leveraging a dataset of $11,000$ reviews generated by five proprietary and open-source models for $1,000$ ICLR and NeurIPS papers. 
Spanning the 2021--2025 period, these machine-generated reviews are compared against original human feedback across diverse prompting strategies to identify systematic behavioral divergences. 
Our analysis reveals that the generated reviews diverge significantly from feedback provided by human reviewers: LLM ratings are less variable, positively biased, and overconfident, and their cross-reference patterns are model-dependent and distinct from human norms.
Furthermore, when evaluated through PRAIB, we observe that LLMs tend to generate longer, more complex reviews, yet frequently overlook the atomic weaknesses noted by human reviewers.
By characterizing \emph{where} and \emph{how} LLMs reviewing behavior departs from human norms, PRAIB provides the community with a diagnostic tool for identifying which aspects of the review process LLMs can reliably support today and which require further development before deployment.
\end{abstract}

\section{Introduction}
\label{sec:introduction}

As large language models are increasingly used to generate or assist with
peer reviews~\cite{liang2024monitoring}, a central question
emerges: how should we evaluate whether an LLM-generated review is actually good? 
Existing approaches either reduce this to numerical rating
prediction~\cite{sahu2025reviewertoo, jin2024agentreview}, rely on
LLM-as-a-Judge pipelines~\cite{garg-etal-2025-revieweval}, or measure token-level statistics to detect AI-generated text~\cite{kobak2025delving}. 
Existing peer review benchmarks typically evaluate machine-generated feedback through dimensions such as helpfulness, verifiability, \cite{sadallah2025good} or deficiency \cite{du2024llms}. While these metrics quantify quality, they often overlook the critical reference point of human alignment; a review may score highly on objective scales yet remain stylistically artificial or diverge sharply from human norms. We argue that "behavioral patterns" are as essential as quality, particularly given the recent proliferation of automated review generation systems. By assessing the inherent, zero-shot capabilities of LLMs -- modeling the behavior of a "lazy" reviewer -- we establish a foundational baseline for more sophisticated systems.

\begin{figure*}[t]
\centering
\includegraphics[width=1.0\linewidth]{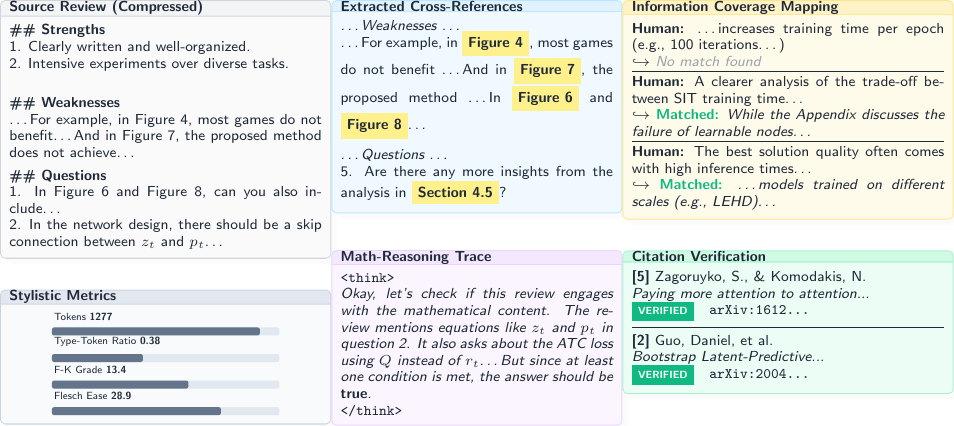}
\caption{\textbf{Overview of the evaluation framework.} \textit{Source Review} represents the generated text; \textit{Extracted Cross-References} highlights entities retrieved via custom regex; \textit{Stylistic Metrics} quantifies review complexity and length; \textit{Math-Reasoning Trace} provides an interpretable rationale for binary classifications; \textit{Information Coverage} employs an LLM judge to map extracted atomic strengths and weaknesses against human baselines; and \textit{Citation Verification} validates formatted references against external databases.}
\label{fig:our_schema}
\end{figure*}

Our investigation is reinforced by changing trends at ICLR and NeurIPS  across the pre- and post-2023 period (see Figure~\ref{fig:readability_trends}), implying that these developments could stem from the increasing use of LLMs for review augmentation and modification.
Ultimately, a high-quality review must not only satisfy predefined metrics but also mirror the way humans engage with scientific content. Our proposed framework helps identify "gamification," where a model might produce an unnaturally verbose or complex review that covers the same technical ground as a much more compact human critique. 
We define specificity as the degree of engagement with a manuscript’s technical core, characterized by explicit references to concrete elements -- such as figures, equations, and theorems -- and the provision of citations that are factually verifiable against external bibliographic databases. 
In contrast, a readable review balances linguistic clarity with structural complexity to ensure that feedback remains professional and actionable. 
To operationalize these attributes, we introduce a multi-dimensional framework that quantifies behavioral patterns using psycholinguistic metrics, such as Token Type Ratio (TTR) and Flesch Reading Ease (FRE), alongside novel measures of mathematical engagement and cross-referencing. 
We further evaluate the information coverage of model-generated strengths and weaknesses by benchmarking them against human references.
Rather than utilizing conventional statistics for rating differences, which may be inappropriate for such categorical data, we prioritize assessing the alignment between human and machine evaluations through agreement metrics. Our contribution can be divided into several parts:
\begin{itemize}
    \item \textbf{A Multi-dimensional Evaluation Framework} --- We introduce a framework to characterize the behavioral patterns of automated peer reviews, extending beyond surface-level linguistics. This approach integrates novel metrics for mathematical engagement and psycholinguistic complexity alongside factuality verification against external bibliographic records and atomic informational coverage (see Fig. \ref{fig:our_schema}).

    \item \textbf{Comparative Study} --- We conduct a systematic analysis of human reviewing patterns by partitioning the dataset into two distinct eras: pre- and post-LLM emergence. This enables a granular investigation into how academic discourse and review quality have evolved following the integration of generative AI tools.

    \item \textbf{Systematic SOTA Benchmarking} --- Systematic SOTA Benchmarking: We conduct evaluation of five open-source and proprietary LLMs, utilizing prompting strategies aligned with official conference guidelines. This provides a direct comparison between model outputs and "gold standard" human expert feedback.

    \item \textbf{Human-AI Alignment Assessment} --- By employing Krippendorff’s $\alpha$, we quantify the level of agreement between OpenReview data and LLM-generated reviews, offering empirical evidence on the current limitations and AI capabilities in replicating human-level expert judgment. Source code is available at \href{https://anonymous.4open.science/r/PRAIB-1EDE/}{the anonymous repository}.
\end{itemize}

\section{Related work}

\paragraph{LLMs in Peer Review.} 

LLMs in peer review range from standalone assistants to autonomous agents designed to mitigate human bias, inconsistency, and oversight.
Frameworks like AgentReview \cite{jin2024agentreview} and ReviewerToo \cite{sahu2025reviewertoo} model the social dynamics of the process, simulating authors, reviewers, and chairs to reduce groupthink.
Other systems focus on multi-agent interactions, employing domain-specific reviewer agents \cite{liutowards} or providing end-to-end pipelines for both paper and review generation \cite{lu2024ai}.
Evaluation strategies for these systems vary widely.
Some measure predictive accuracy for final acceptance decisions \cite{jin2024agentreview,sahu2025reviewertoo} or alignment with human scores \cite{lu2024ai}.
Others use "LLM-as-a-judge" to evaluate review "specificity" \cite{sahu2025reviewertoo,garg-etal-2025-revieweval} or human preference \cite{bougie-watanabe-2025-generative}, though this risks bias toward longer texts.
Text-level evaluations include measuring overlap with human-identified strengths and weaknesses \cite{liutowards} or relying on expensive manual labeling of segment shortages \cite{du2024llms}.
Finally, performance is heavily influenced by instruction methods, which range from targeted fine-tuning \cite{idahl-ahmadi-2025-openreviewer} to advanced prompting strategies, such as few-shot with reflection \cite{lu2024ai} or sentiment-specific (positive, negative, neutral) prompts \cite{demetrio2025gen}.

\paragraph{Open Review Analysis}
Due to its extensive open-access history, OpenReview (particularly ICLR conference) is the primary data source for peer review analysis.
Longitudinal studies using this data \cite{jung2025drives} reveal that post-ChatGPT reviews have become significantly more complex and harder to read.
Furthermore, \cite{liang2024monitoring} estimate that up to 16.9\% of top AI conference reviews show signs of AI modification.
The common use of LLMs is confirmed by "vocabulary shift" techniques \cite{gray2024chatgpt,kobak2025delving} that detect AI signatures without explicit training corpora, confirming that generative AI has fundamentally changed the scholarly review ecosystem.

\paragraph{Benchmarks.}
Benchmarks for LLM-generated reviews span from static datasets to multi-turn dialog frameworks.
Static resources emphasize volume or specific structures, such as Gen-Review's \cite{demetrio2025gen} 81k single-venue reviews, AgentReviewers' \cite{pmlr-v267-lu25p} standardized "Strengths and Weaknesses," and large-scale datasets focused solely on AI detection rather than content quality \cite{yu2025your}.
Frameworks like ReviewBench \cite{gao2025reviewagents} evaluate these static outputs using general NLP metrics (Inverse Self-BLEU, ROUGE) and BERT-based sentiment analysis, which often miss technical nuances and yield spurious results.
To address limitations of static datasets, recent works model peer review as a dynamic process.
Work done in \cite{tan2024peer} captures multi-turn interactions between authors, reviewers, and decision-makers, while \cite{zhang2025re} diversifies beyond ICLR-centric data with extensive rebuttal discussions.
Despite structural advances, evaluation remains a critical bottleneck.
Both paradigms rely heavily on n-gram overlap (ROUGE/BLEU), general semantic similarity (LLM-as-a-judge), or Mean Absolute Error (MAE).
These metrics are frequently inadequate, as they overlook domain-specific specificity and incorrectly treat categorical venue scores as continuous variables.

\section{Data Acquisition and Preparation}

In order to analyze the influence of LLMs on the peer review process, we systematically extract data from OpenReview, a primary platform for open-access peer reviews in machine learning.
In our data extraction process, we consider the differences between versions 1 and 2 of the OpenReview API and the integrity of the collected metadata across multiple venues.
We collected peer reviews for the following conferences: 
\begin{itemize}
    \item ICLR (International Conference on Learning Representations) years: 2013-2025,
    \item NeurIPS (Conference on Neural Information Processing Systems) years: 2021-2025.
\end{itemize}

We retrieved the data using the Python API access to OpenReview through the \verb|openreview-py| library. 
We utilized specific invitation identifiers 
to isolate official conference tracks. 
For each conference, we collected all publicly accessible papers. 
First, we extracted article-specific metadata, such as titles, abstracts, and keywords. 
Next, using the unique identifier for each submission, we retrieved the entire discussion history, including comments, reviews, and meta-reviews. 
These data were subsequently standardized to ensure consistency across different venues.
For each entry, we recorded the text, metadata, author identity, and paper assessment details.
The structure of reviews changed year by year, posing a particular challenge for standardizing the fields. 
Over the years, the structure became more fine-grained; this is particularly notable for ICLR, where we have access to data spanning over 10 years. Due to the low paper count for years before and including 2018, we grouped all papers for ICLR in a single year group. Dataset count is provided in Appendix \ref{sec:appendix_analysis}, Table \ref{tab:iclr_nips_counts}. Although rejected papers account for only ~5\% of the available NeurIPS dataset, we address this imbalance by employing a balanced sampling strategy, ensuring equal representation of accepted and rejected papers in our benchmark.

To process the manuscripts, we first convert the PDF files to images using \texttt{pypdfium2} \cite{pypdfium2} (Apache-2.0 license) and subsequently extract the text into Markdown format. Specifically, we employ the DeepSeek-OCR model \cite{wei2025deepseek} (MIT license) due to its strong capability in parsing complex mathematical formulas. Utilizing a specialized vision-language model rather than a standard PDF parser significantly improves the extraction fidelity of equations and tables, while simultaneously reducing the token footprint of the resulting Markdown. Because full manuscripts and their appendices can be exceedingly long, we truncate each parsed document at a maximum of 50,000 tokens. This threshold is sufficient to encompass the main body of the paper; any resulting truncation typically affects only the later sections of the appendix, which peer reviewers are generally not obligated to read.

\section{Evaluation Methodology} \label{sec:methodology}

Our evaluation protocol is designed to compare human and LLM-generated reviews along multiple complementary dimensions.
We focus not only on surface-level stylistic characteristics, but also on whether the generated reviews exhibit paper-specific engagement, calibrated scoring behavior, and agreement patterns comparable to those of human reviewers.
To do this, we define a set of metrics that capture textual complexity, content specificity, and reviewer alignment.

\subsection{Textual Complexity and Readability}
\label{subsec:readability}

We quantify the linguistic properties of reviews using complementary metrics on full texts:

\begin{itemize}
    \item \textbf{Token count (TKC)} -- the total number of word tokens, capturing review length.
    \item \textbf{Type-Token Ratio (TTR)} -- the ratio of unique word types to total tokens, reflecting vocabulary richness. 
    \item \textbf{Flesch Reading Ease (FRE)} -- a readability score where higher values correspond to easier-to-read text, with scores typically in range from 0 to 100 \cite{flesch1948new}.
    \item \textbf{Flesch-Kincaid Grade Level (F-K Grade)} -- an estimate of the U.S.\ school grade level required to comprehend the text, where higher values denote greater syntactic complexity ~\cite{kincaid1975derivation}.
\end{itemize}

These metrics have been used to measure the readability of scientific papers \cite{shannon2024improving, jung2025drives}. 
The exact formulae are provided in Appendix, Table \ref{tab:readabilityformulae}. We examine both the temporal evolution of these metrics for human reviews (in Appendix \ref{sec:appendix_analysis}, Table \ref{tab:combined_metrics_yearly}) and later on the distributional differences between LLM generated reviews for a selected subset of papers.

\begin{figure*}[ht]
\begin{subfigure}{0.48\textwidth}
  \centering
  \includegraphics[width=0.9\linewidth]{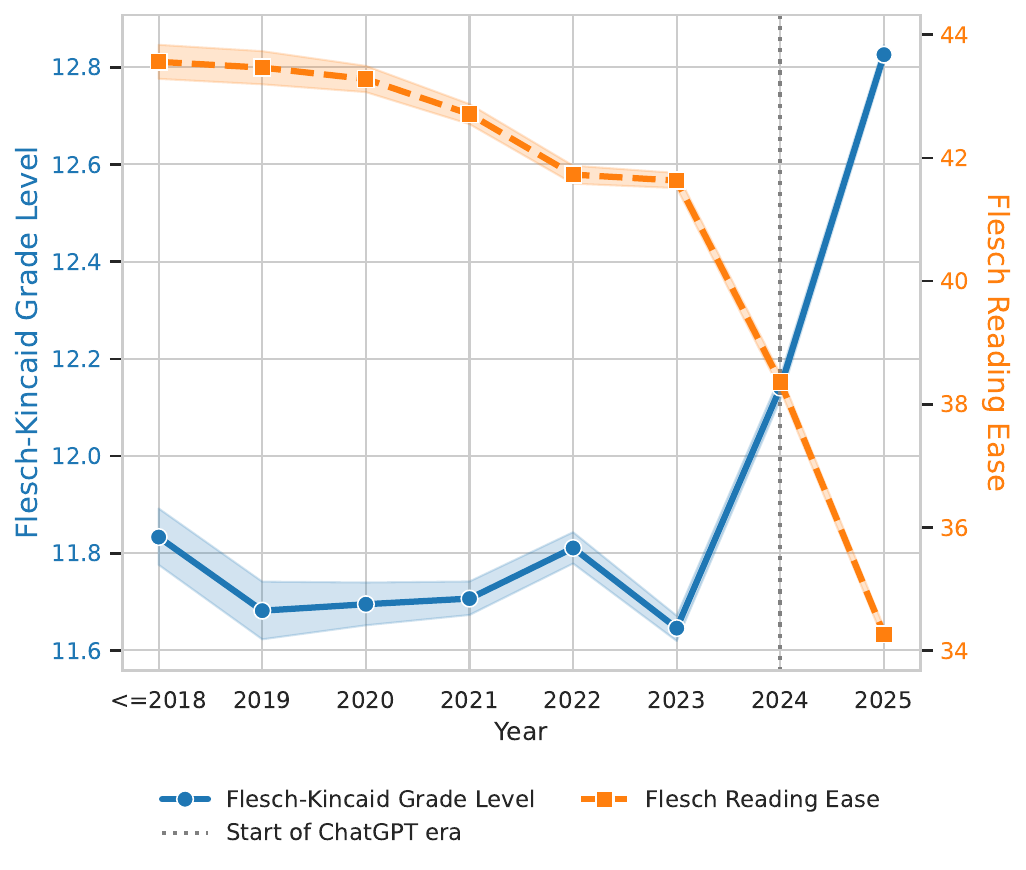}
    \caption{\textbf{
  ICLR readability metrics trend.
  }}
  \label{fig:iclr_readability_metrics_trend}
\end{subfigure}
\hfill
\begin{subfigure}{0.48\textwidth}
  \centering
  \includegraphics[width=0.9\linewidth]{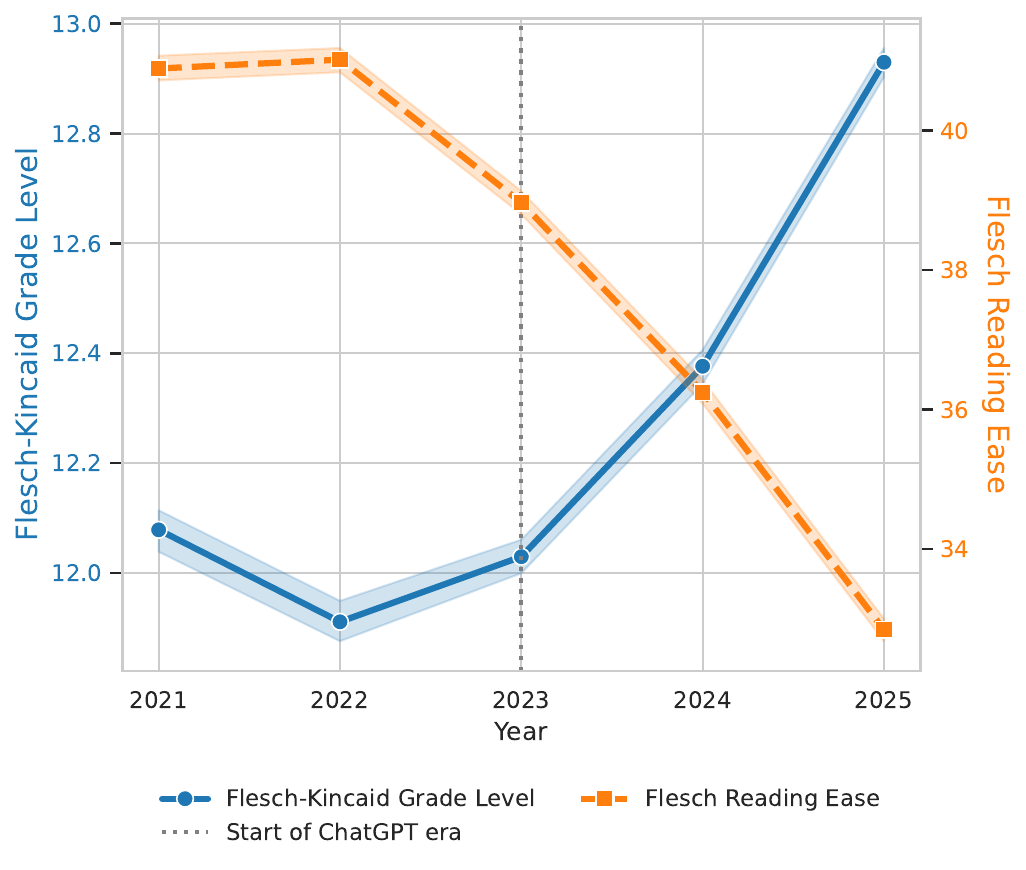}
  \caption{\textbf{
  NeurIPS readability metrics trend.
  }}
  \label{fig:neurips_readability_metrics_trend}
\end{subfigure}
\caption{Timeline of the peer review dataset, spanning both the pre- and post-LLM eras. The public release of ChatGPT (November 30, 2022) serves as the dividing line between baseline human reviews and the post-adoption cycles, highlighting key milestones such as the ICLR 2024 review release (November 10, 2023) and NeurIPS 2024 author notifications (August 22, 2023).}
\label{fig:readability_trends}
\end{figure*}

\subsection{Content Engagement}
\label{subsec:specificity}

Existing evaluations predominantly focus on measuring text frequencies to detect LLM-generated content \cite{liang2024monitoring, kobak2025delving}. While this approach may indicate the proportion of LLM-edited text, it is too shallow to assess substantive text changes in the context of peer review. Through our analysis, we identified specific areas where LLMs may struggle: (a) referencing exact locations within a manuscript, as spatial layout is often lost during PDF-to-Markdown conversion; (b) engaging meaningfully with mathematical content, which requires tracking complex notation across different sections; and (c) producing valid external citations in pre-defined formats, where the commonly used marker "et al." \cite{liang2024monitoring} does not provide a way to further fact-check such citations. Consequently, we evaluate the depth of review engagement along three axes:

\begin{itemize}
    \item \textbf{References to paper content (Xref.)} -- we extract mentions of structural elements (figures, tables, sections, equations, theorems, definitions, lemmas, corollaries, page numbers) using a comprehensive set of regular expressions (see Appendix \ref{papercontentregex}). The count of such references per review reflects the degree of specificity.
   \item \textbf{Mathematical content detection} -- we employ a secondary LLM (\texttt{Qwen/Qwen3.5-8B}) to classify whether each review contains substantive engagement with mathematical content. We define an interaction as either a direct reference to a specific equation or a formal result. To validate this approach, independent experts manually annotated, by authors, a subset of 30 papers to establish a ground-truth reference; the LLM attained 96.6\% accuracy on these golden examples. Using this validated classifier, we compute the ratio of reviews that engaged with the content mathematically. The exact prompt is detailed in Appendix \ref{sec:math_content_detection}.
    \item \textbf{External citations} -- we identify references formatted according to standard academic citation styles (MLA, APA, IEEE) using targeted regular expressions. To ensure validity, each extracted citation is subsequently verified against an external bibliographic database (exact procedure and regex can be found in Appendix \ref{appendix:citations}).
\end{itemize}

We demonstrate that cross-referencing serves as an interpretable proxy for grounding specificity. By applying the RevUtil model \cite{sadallah2025good} -- a GPT-4o-based evaluator fine-tuned on human-annotated data -- to the 2025 subset, we observe correlation scores of $0.393$ for generated reviews and $0.298$ for human reviews. Regarding mathematical engagement, we contend that while LLM feedback may appear helpful, it risks "gamifying" the review process if it lacks grounding in formal formulations and analysis. This distinction is critical, as a reviewer who engages with specific technical sections is more likely to base their critique on the manuscript's actual content rather than subjective perception. Furthermore, departing from existing approaches that simply count "et al." occurrences, our framework identifies potential hallucinations by verifying citations against external databases, including Crossref, OpenAlex, and DBLP. Collectively, these metrics provide diagnostic insights into how LLM engagement patterns diverge from human norms, while the qualitative evaluation of these individual dimensions is left for future work. 

\subsection{Reviewer Agreement}
\label{subsec:agreement}

We assess the degree of agreement among reviewers using several complementary measures:

\begin{itemize}
    \item \textbf{Krippendorff's $\alpha$} -- we evaluate inter-annotator agreement using an ordinal implementation of Krippendorff's $\alpha$ ~\cite{Krippendorff2011ComputingKA}. This metric robustly corrects for chance agreement while accounting for the ordered nature of the rating scales. A detailed mathematical formulation is provided in Appendix~\ref{sec:krippendorfss_alpha}.
    \item \textbf{Math engagement} -- we report conditional detection rates, showing how frequently an LLM engages with mathematical content given that the original human review did (or did not). This highlights the alignment in mathematical focus between human and AI reviewers.
\end{itemize}

While previous studies typically measure the variance between ratings \cite{jung2025drives, kargaran2025insights}, we note that the underlying distribution of ratings changes from year to year, as depicted on Figure \ref{fig:rating_labels}. To address this, we employ Krippendorff's $\alpha$ which is better suited for ordinal scales and naturally accommodates a variable number of reviewers per paper.

\section{Models and Experimental Setup}

We evaluate a diverse selection of open-source and proprietary models under various prompting strategies. Our set of state-of-the-art models includes efficient open-source models (\texttt{google/gemma-3-12b-it} \cite{gemma_2025}, \texttt{Qwen/Qwen3.5-9B} \cite{qwen3.5}), a domain-specialized model (\texttt{maxidl/Llama-OpenReviewer-8B} \cite{idahl-ahmadi-2025-openreviewer}), and large-scale frontier models (DeepSeek-V3 \cite{deepseekai2024deepseekv3technicalreport} and \texttt{gpt-5-2025-08-07}). We adopt a zero-shot, single-prompt evaluation setting. While this may not fully capture the iterative, multi-turn interactions a human reviewer might have with an LLM, it allows us to assess the models' intrinsic capabilities and ensures a fair, standardized comparison.

We designed two distinct prompting templates (full texts are provided in Appendix \ref{sec:prompts}): a baseline "lazy"  template and detailed conference-specific templates. The lazy template is a simple, conference-agnostic prompt designed to test a model's inherent ability to independently structure a peer review. In contrast, the conference-specific templates strictly enforce the exact rating and confidence labels utilized by the venues in a given year. We emphasize this distinction because previous studies frequently assume a static 1-10 rating scale, which poorly reflects the actual historical distribution of formats (as illustrated in Figure~\ref{fig:rating_labels}, Appendix~\ref{sec:appendix_analysis}). To ensure comparable baselines, we evaluate OpenReviewer \cite{idahl-ahmadi-2025-openreviewer} using both our standardized templates and its original training prompt.

Finally, while we intended to keep the inserted guidelines as faithful to the official conference instructions as possible, formatting adaptations were necessary to enforce structured outputs. During parsing, we extract only the pre-defined fields (summary, strengths, weaknesses, questions, and limitations/additional feedback), systematically discarding any conversational filler or unprompted text produced by the models. Concrete examples of model outputs can be found in Appendix \ref{sec:review_examples}.

\null

\section{Longitudinal Analysis of Human Reviews}
\label{sec:longitudinal}
To provide context for our metrics, we assess how peer reviews have evolved over time, specifically comparing the periods before and after the introduction of ChatGPT. 
In this analysis, we focus on the stylistic characteristics and specificity of the review content. 
We hypothesize that these changing patterns can be attributed to the increasing use of LLMs as a tool for proofreading or even the direct generation of reviews.

\textbf{The Pre-ChatGPT vs. Post-ChatGPT era.}
The introduction of ChatGPT marks a distinct shift in the stylistics of academic reviews, characterized by a clear trend in textual complexity (see Figure \ref{fig:readability_trends}). In the post-ChatGPT era (beginning with \textit{ICLR} 2024 and \textit{NeurIPS} 2023), reviews have become measurably more difficult to read. Metrics evaluating educational grade level and linguistic complexity display a sharp, continuous upward trajectory. This indicates a shift toward longer sentences and more complex vocabulary traits commonly associated with LLM-generated text. While this trend could potentially be influenced by confounding variables, such as an exponential increase in submissions or a demographic shift within the reviewer pool, such explanations are unlikely to be the primary cause. This conclusion aligns with the analysis in \cite{liang2024monitoring}, which observed a similar exponential growth in the frequency of LLM-typical adjectives. A related study focusing solely on ICLR \cite{jung2025drives} further supports these empirical claims. 
Although selecting only pre-ChatGPT papers was an option, we opted for the full 2021--2025 range to ensure our benchmark accounts for recent distributional shifts. 
Given the difficulty of identifying individual machine-generated reviews, we prioritize population-level statistics over single-review classification. 
By doing so, we can effectively measure whether the behavioral patterns of LLMs align with the mean trajectory of human reviewers over this five-year period.

\section{Benchmark Results: Human vs.\ LLM Reviewers}
\label{sec:results}

Table~\ref{tab:benchmark_extended} reports all metrics for the \texttt{extended\_prompt} configuration across five LLM reviewers and the human baseline. Each model -- conference cell comprises 500 reviews (100 papers $\times$ 5 years, 2021 -- 2025); values are aggregated across ICLR and NeurIPS unless noted (Benchmark dataset counts are reported in Appendix \ref{tab:paper_counts}, \ref{tab:generated_counts}). Krippendorff's $\alpha$ is computed per conference against the human reviewer pool (reported as ICLR/NeurIPS). All $\alpha$ values fall below the $0.667$ reliability threshold~\cite{Krippendorff2011ComputingKA}, so we interpret them comparatively.

\begin{table*}[t]
\centering
\caption{Benchmark results for the \texttt{extended\_prompt} configuration across all models. Cit.: mean verified/total citations; $\alpha_r$: Krippendorff's $\alpha$ (ICLR/NeurIPS). $P(+|+)$/$P(+|-)$: conditional math detection rates (ICLR/NeurIPS). Bold denotes values closest to human baseline.}
\label{tab:benchmark_extended}
\small
\setlength{\tabcolsep}{3pt}
\begin{tabular}{l rrrr rrr r rr}
\toprule
 & \multicolumn{4}{c}{\textbf{Read.}} & \multicolumn{3}{c}{\textbf{Spec.}} & \multicolumn{1}{c}{\textbf{Agree.}} & \multicolumn{2}{c}{\textbf{Math C.}} \\
\cmidrule(lr){2-5} \cmidrule(lr){6-8} \cmidrule(lr){9-9} \cmidrule(lr){10-11}
\textbf{Source}
 & TKC $\downarrow$ & TTR $\uparrow$ & FRE $\uparrow$ & FKG $\downarrow$
 & Cit. & Xref. & Math
 & $\alpha_r$
 & $P(+|+)$ & $P(+|-)$ \\
\midrule
Human & 532 & 0.460 & 39.1 & 12.1 & 0.039/0.11 & 2.26 & 0.427 & 0.34/0.24 & --- & --- \\
\midrule
DeepSeek & \textbf{637} & 0.413 & 13.7 & 14.9 & 0/0 & 0.84 & 0.286 & 0.14/0.12 & 0.37/0.43 & \textbf{0.23/0.23} \\
Gemma    & 898  & 0.387 & 15.0 & 15.7 & 0/0 & 0.74 & 0.329 & 0.21/0.14 & 0.41/0.50 & .25/.28 \\
GPT-5    & 1449 & 0.388 & 21.3 & 14.5 & 0/0 & \textbf{2.47} & 0.758 & \textbf{0.36/0.26} & \textbf{0.91/0.90} & 0.69/0.67 \\
OpenReviewer & 678 & \textbf{0.433} & \textbf{38.3} & \textbf{12.2} & \textbf{0.063}/0.190 & 2.77 & \textbf{0.403} & 0.32/0.22 & 0.48/0.52 & 0.35/0.35 \\
Qwen     & 1120 & 0.388 & 20.1 & 14.8 & 0/0 & 4.96 & 0.686 & 0.22/0.17 & 0.86/0.89 & 0.65/0.53 \\
\bottomrule
\end{tabular}
\end{table*}

\paragraph{Readability.}
LLM-generated reviews diverge sharply from human prose in surface-level complexity.
All general-purpose models produce text with Flesch Reading Ease scores between 13.7 (DeepSeek) and 21.3 (GPT-5), far below the human baseline of 39.1 -- a drop that moves the text from ``difficult'' into ``very difficult'' territory.
This is driven by longer sentences and more complicated vocabulary, as reflected in Flesch-Kincaid grade levels of 14.5--15.7 (vs.\ 12.1 for humans).
Review length also increases substantially: GPT-5 averages 1{,}449 tokens (2.7$\times$ the human mean, Appendix, Tab. \ref{tab:model_behavior_ratios}), and even the more restrained DeepSeek produces 637 tokens (+20\%).
Type-Token Ratios are uniformly lower for LLMs (0.387--0.413 vs.\ 0.460), indicating reduced vocabulary diversity, consistent with the well-known tendency of LLMs toward repetitive phrasing (longer texts dilute TTR).
The sole exception is OpenReviewer: with an FRE of 38.3, F-K grade of 12.2, and TTR of 0.433, it closely replicates the linguistic profile of human reviews, demonstrating the value of domain-specific fine-tuning over raw model scale.

\paragraph{Content specificity.}
We evaluate paper-specific engagement along three axes.
For \emph{cross-references} to manuscript elements (figures, tables, sections, equations), human reviews average 2.26 per review. OpenReviewer exceeds this (2.77), GPT-5 approximates it (2.47), and Qwen overshoots substantially (4.96). DeepSeek (0.84) and Gemma (0.74) produce far fewer, indicating limited structural engagement with the manuscript.  For \emph{external citations} almost all models failed to provide citations in valid format; only OpenReviewer produced some citations, from which 33\% could be verified, similarly to human (35\%). 
Nevertheless, this finding suggest that LLMs should be given explicit instruction to add formal citations that could be later fact-checked against sources. For \emph{mathematical engagement} -- the proportion of reviews substantively engaging with equations or formal results -- GPT-5 (75.8\%) and Qwen (68.6\%) substantially exceed the human rate (42.7\%), whereas DeepSeek (28.6\%) and Gemma (32.9\%) fall below it. OpenReviewer (40.3\%) is closest to the human baseline, suggesting that the overdetection of frontier models and the underdetection of smaller models bound the human distribution.

\paragraph{Scoring agreement.}
Krippendorff's $\alpha$ for ratings reveals a clear model hierarchy.
GPT-5 achieves the highest agreement with human panels ($\alpha_r = 0.36/0.26$ on ICLR/NeurIPS), matching or slightly exceeding the human-only baseline ($\alpha_r = 0.34/0.24$).
Gemma ranks second ($\alpha_r = 0.23/0.14$), followed by Qwen ($\alpha_r = 0.22/0.17$) and OpenReviewer ($\alpha_r = 0.32/0.22$).
DeepSeek exhibits the weakest alignment ($\alpha_r = 0.14/0.12$). Confidence score returned by the models exhibit positive bias and low variance, hence we don't apply Krippendorff's $\alpha$ and perform more in depth analysis in Appendix \ref{appendix:confidence_score}.

\paragraph{Mathematical focus alignment.}

The conditional detection rates $P(+|+)$ and $P(+|-)$ capture whether LLM-generated and human reviews tend to engage mathematically with the same papers. A high $P(+|+)$ means that when a human reviewer engages with the mathematical content of a paper, the LLM does so as well indicating shared sensitivity to theoretical depth. A high $P(+|-)$ means the LLM engages mathematically even on papers that human reviewers treated more informally, suggesting the model brings its own, broader propensity for technical engagement regardless of the human signal. GPT-5 and Qwen show the strongest co-occurrence with human mathematical engagement ($P(+|+) = 0.90$-$0.91$ and $0.86$-$0.89$, respectively), meaning they reliably respond to the same theoretical content that draws human attention. At the same time, both models also engage mathematically on papers humans did not flag in this way ($P(+|-) = 0.67$-$0.69$ for GPT-5, $0.53$–$.65$ for Qwen), which may reflect a tendency to highlight formal reasoning even where human reviewers chose not to. DeepSeek and Gemma are more selective, engaging mathematically primarily when the paper's formal content also draw a human response (with $P(+|+) \leq 0.50$, $P(+|-) \leq 0.28$). OpenReviewer sits between these extremes with a relatively balanced profile ($P(+|+) = 0.48$-$0.52$, $P(+|-) = 0.35$).
These patterns suggest that frontier LLMs can effectively complement human reviewers by consistently engaging with mathematical content and, in some cases, identifying formal structures that humans may overlook.

\paragraph{Prompt ablations.}
Comparing the extended and guidelines prompts reveals three consistent effects
(full results in Appendices ~\ref{appendix:stylistic_specificity_ablations}, ~\ref{appendix:confidence_score}, ~\ref{appendix:rating_dist_prompt}).
The extended prompt increases cross-reference counts (by $1.2\times$--$5.5\times$
depending on the model), produces longer but marginally more readable reviews
(token counts up 8--40\%, FRE up 2--5 points), and substantially affects rating
self-consistency: GPT-5 is the only model with meaningful agreement between the
two prompts ($\alpha=0.76$ on ICLR, $0.47$ on NeurIPS), while all other models
show near-zero or negative $\alpha$, indicating their scores are driven more by
prompt phrasing than by paper content.

\section{LLM-coverage computation}

Rather than employing an LLM-as-a-Judge approach to evaluate predefined dimensions, we measure information coverage by treating human-authored reviews as the ground truth. While we acknowledge potential LLM contamination within the human review pool, we observe that models often generate a higher number of strengths and weaknesses than their human counterparts. However, this increased volume -- while superficially suggesting a higher level of engagement -- does not necessarily imply that the models are attending to the same technical points prioritized by human experts.

\paragraph{Atomic Strength and Weakness Extraction.} We utilize original human reviews and LLM-generated reviews from the 2025 subset to perform atomic strength and weakness extraction. This temporal selection is designed to minimize potential data leakage, as we hypothesize that the majority of evaluated models possess knowledge cut-offs preceding 2024. 
By using the 2025 data, we ensure that the evaluation reflects the models' inherent reasoning and information coverage capabilities on novel scientific content rather than memorized text. 
Simple regular expression-based parsing is insufficient for this task, as LLMs frequently pack multiple distinct critiques into a single bullet point, and human reviewers often do not follow a standardized structure, frequently interweaving strengths and weaknesses within a single sentence.
We define an \textit{atomic} comment as one or more literally extracted sentences that convey a single, self-contained piece of information. 
To perform these extractions, we employed \texttt{Qwen/Qwen3.5-9b} using the specific prompts detailed in Appendix~\ref{sec:prompts}. The fidelity of this extraction process is validated by ROUGE-L scores, which remained uniformly high ($0.96$--$1.00$) across all human and model variants. These scores confirm the effectiveness of our approach in isolating distinct technical units while strictly preserving the original literal phrasing of the reviewers.
In the 2025 subset, human reviewers provided an average of $5.58$ atomic weaknesses and $3.74$ strengths, with accepted papers ($4.04$ strengths, $5.24$ weaknesses) receiving more strengths and fewer weaknesses compared to rejected papers ($3.45$ strengths, $5.93$ weaknesses). All evaluated LLMs consistently generated more atomic strengths than human reviewers, with ratios ranging from $1.32$ to $2.55$. While most models also significantly increased the number of weaknesses (ranging from $1.23$--$2.90$) -- most notably GPT-5-1, which provides nearly triple the human baseline (ratio 2.90) -- DeepSeek was the sole exception, identifying fewer mean weaknesses than the original human reviews. We provide more in-depth analysis in Appendix \ref{sec:appendix:sw_analysis}.
\paragraph{Information coverage.} Atomic comment coverage was evaluated via \texttt{google/gemma-3-12b-it} (Prompt \ref{lst:coverage_prompt}). Across all evaluated models, information coverage is significantly higher for atomic strengths than weaknesses, with mean atomic recall values of $0.89$ and $0.79$, respectively. GPT-5-1 is the top-performing model in both categories, achieving an atomic recall of $0.93$ for weaknesses and $0.94$ for strengths. In contrast, DeepSeek provides the lowest coverage for weaknesses ($0.70$ recall), while OpenReviewer exhibits the lowest coverage for strengths ($0.78$ recall). These results suggest that while LLMs are highly effective at identifying the technical merits noted by humans, capturing the full breadth of human-identified weaknesses remains a more significant challenge for most models. We report the overall statistics in Table \ref{tab:atomic_stats_full}.

\begin{table*}[t]
\centering
\caption{Coverage (Atomic Recall) and item counts for strengths and weaknesses, comparing individual models against the aggregated average and human baseline.}
\label{tab:atomic_stats_full}
\footnotesize
\setlength{\tabcolsep}{3pt}
\begin{tabular}{l cccccc c}
\toprule
\textbf{Metric} & \textbf{DeepSeek} & \textbf{Gemma} & \textbf{GPT-5} & \textbf{OpenRev.} & \textbf{Qwen} & \textbf{Model Avg.} & \textbf{Human} \\
\midrule
\multicolumn{8}{l}{\textit{Strength Dimension}} \\
Atomic Recall & 0.92 ± 0.21 & 0.90 ± 0.27 & 0.94 ± 0.19 & 0.78 ± 0.31 & 0.89 ± 0.24 & 0.89 ± 0.24 & --- \\
Count & 5.19 ± 1.10 & 6.66 ± 2.18 & 7.71 ± 2.31 & 4.03 ± 4.63 & 6.08 ± 1.87 & 5.93 ± 2.42 & 3.74 ± 2.14 \\
\midrule
\multicolumn{8}{l}{\textit{Weakness Dimension}} \\
Atomic Recall & 0.70 ± 0.30 & 0.77 ± 0.29 & 0.93 ± 0.20 & 0.73 ± 0.33 & 0.84 ± 0.26 & 0.79 ± 0.28 & --- \\
Count   & 4.33 ± 1.14 & 7.06 ± 3.24 & 10.22 ± 3.06 & 6.25 ± 7.04 & 6.93 ± 3.30 & 6.96 ± 3.56 & 5.58 ± 4.40 \\
\bottomrule
\end{tabular}
\end{table*}

\section{Discussion}
\label{sec:discussion}

Our findings demonstrate a significant behavioral divergence between Large Language Models (LLMs) and human authors. While LLMs typically produce longer and more structurally complex text, this increased verbosity does not consistently correlate with higher informational density. However, the high degree of mathematical engagement and cross-referencing observed in specific models suggests they could effectively augment human-led technical reviews. 
Furthermore, model sensitivity to prompt engineering -- where specialized instructions yield denser, mathematically rigorous, and cross-referenced outputs -- identifies a key target for future fine-tuning. A critical bottleneck remains the universal failure of general-purpose models to generate verifiable citations, which are fundamental to academic integrity. Consequently, future evaluation protocols must establish citation factuality as a baseline requirement for any LLM-assisted review system. 
Given these stylistic and factual gaps, LLMs are currently unsuitable as wholesale replacements for human expertise. Instead, we propose a synergistic pipeline: leveraging frontier models for mathematical analysis and fine-tuned models for stylistic drafting, while retaining human reviewers for final qualitative judgment and citation verification.

\section*{Limitations}

As original submissions may be edited and the review may not reflect the corresponding paper content, locating original submissions on pre-print repositories like arXiv is logistically unfeasible. Nevertheless, high coverage ratios (above $0.9$ for \texttt{GPT-5}) suggest that revised versions remain reliable proxies.
We also observe that LLMs do not always follow the given format of the review, inventing their own sections and scores.
This leads to a situation where, for around 1\% of reviews, we are unable to parse the reviews to extract scores and decide to drop them from the analysis.
Our method accounts only for citations structured according to well known formats but unfortunately LLMs as well as reviewers do not always follow standardized citations format because unawareness or mistake.
To verify citation, we use publicly available APIs, that methods may not contain all possible domains and resources.
As we only focus on verifying the existence of referenced resource, future research could concentrate on confirming the usefulness of these resources, evaluating whether it is reasonable to cite them and expanding analysis to larger spectrum of possible resources. Moreover, the struggle to provide valid citations could stem from zero-shot setting and unavailability of external sources.
Our framework is limited by the zero-shot setting, LLM-based mathematical engagement classification, and venue coverage (ICLR, NeurIPS). 
Nonetheless, the metrics and benchmark introduced here provide a principled, extensible foundation for measuring progress in automated peer review -- one that we believe will prove essential as LLM adoption in the review process continues to accelerate.

\bibliographystyle{plainnat}
\bibliography{bibliography}

\clearpage

\appendix
\section{OpenReview Data}
\label{sec:appendix_analysis}

\begin{table*}
\centering
\caption{\textbf{Dataset statistics for ICLR and NeurIPS.} The collected counts closely match external reports \cite{iclrstatistics,neuripsstatistics}. Note that due to the NeurIPS' policy on public data availability only accepted papers are completely collected.}
\label{tab:iclr_nips_counts}
\setlength{\tabcolsep}{6pt}
\renewcommand{\arraystretch}{1.05}
\begin{tabular}{lrrrrrrrrr}
\toprule
Year & $\le$2018 & 2019 & 2020 & 2021 & 2022 & 2023 & 2024 & 2025 & Total \\
\midrule
\hline
\multicolumn{10}{c}{\textbf{ICLR}} \\
\hline
Papers & 1485 & 1579 & 2580 & 2993 & 3386 & 4931 & 7361 & 11670 & 35985 \\
\textit{Accept} & 672 & 502 & 687 & 860 & 1095 & 1574 & 2261 & 3708 & 11359 \\
\textit{Reject} & 741 & 917 & 1526 & 1755 & 1580 & 2285 & 3486 & 5019 & 17309 \\
\textit{Withdrawn} & 72 & 160 & 367 & 378 & 711 & 1072 & 1614 & 2943 & 7317 \\
Reviews & 5250 & 4763 & 7766 & 11492 & 13123 & 18532 & 27857 & 46737 & 135520 \\
\midrule
\hline
\multicolumn{10}{c}{\textbf{NeurIPS}} \\
\hline
Papers & - & - & - & 2768 & 2824 & 3394 & 4236 & 5540 & 18762 \\
\textit{Accept} & - & - & - & 2632 & 2671 & 3218 & 4035 & 5287 & 17842 \\
\textit{Reject} & - & - & - & 136 & 153 & 176 & 201 & 254 & 920 \\
Reviews & - & - & - & 10736 & 10330 & 15171 & 16644 & 22373  & 74472 \\
\bottomrule
\end{tabular}
\end{table*}

\begin{table*}
\centering
\caption{Readability metrics for each year for both conferences.}
\label{tab:combined_metrics_yearly}
\resizebox{\textwidth}{!}{
\begin{tabular}{lrrrrrrrr|rrrrr}
\toprule
 & \multicolumn{8}{c}{\textbf{ICLR}} & \multicolumn{5}{c}{\textbf{NeurIPS}} \\
\cmidrule(lr){2-9} \cmidrule(lr){10-14}
\textbf{Metric} & $\leq$2018 & 2019 & 2020 & 2021 & 2022 & 2023 & 2024 & 2025 & 2021 & 2022 & 2023 & 2024 & 2025 \\
\midrule
\textbf{TKC} & 396.34 & 469.64 & 476.39 & 563.96 & 600.12 & 562.36 & 499.44 & 524.41 & 546.84 & 506.99 & 497.08 & 473.47 & 559.54 \\
\textbf{TTR} & 0.52 & 0.49 & 0.48 & 0.45 & 0.43 & 0.44 & 0.47 & 0.47 & 0.46 & 0.46 & 0.47 & 0.48 & 0.48 \\
\textbf{FRE} & 43.56 & 43.47 & 43.28 & 42.71 & 41.73 & 41.63 & 38.36 & 34.27 & 40.89 & 41.01 & 38.97 & 36.25 & 32.85 \\
\textbf{F-K Grade} & 11.83 & 11.68 & 11.70 & 11.71 & 11.81 & 11.65 & 12.14 & 12.83 & 12.08 & 11.91 & 12.03 & 12.38 & 12.93 \\
\textbf{Xref.} & 1.87 & 2.42 & 2.45 & 2.83 & 2.93 & 2.49 & 2.35 & 2.06 & 2.12 & 2.06 & 1.92 & 1.78 & 1.77 \\
\bottomrule
\end{tabular}
}
\end{table*}

Table~\ref{tab:combined_metrics_yearly} reports longitudinal readability metrics for human reviews across both venues. The most notable trend is a steady decline in Flesch Reading Ease — from 43.56 (ICLR $\leq$2018) to 34.27 (ICLR 2025) and from 40.89 to 32.85 over the NeurIPS period — accompanied by a rising Flesch--Kincaid grade level, together indicating that reviews have become progressively more syntactically complex over time.

Table~\ref{tab:iclr_nips_counts} summarises the scale of the dataset, covering 35{,}985 ICLR submissions from 2018 to 2025 and 18{,}762 NeurIPS submissions from 2021 to 2025, yielding a total of 135{,}520 and 74{,}472 reviews respectively. Both venues show consistent year-on-year growth in submission volume, with ICLR nearly quadrupling between 2019 and 2025.

\section{Pre-ChatGPT and Post-ChatGPT Supplementary Analysis} \label{appendix:statistical-analysis}

\begin{figure*}[ht]
\centering
\begin{subfigure}{0.95\textwidth}
  \centering
  \includegraphics[width=1.0\linewidth]{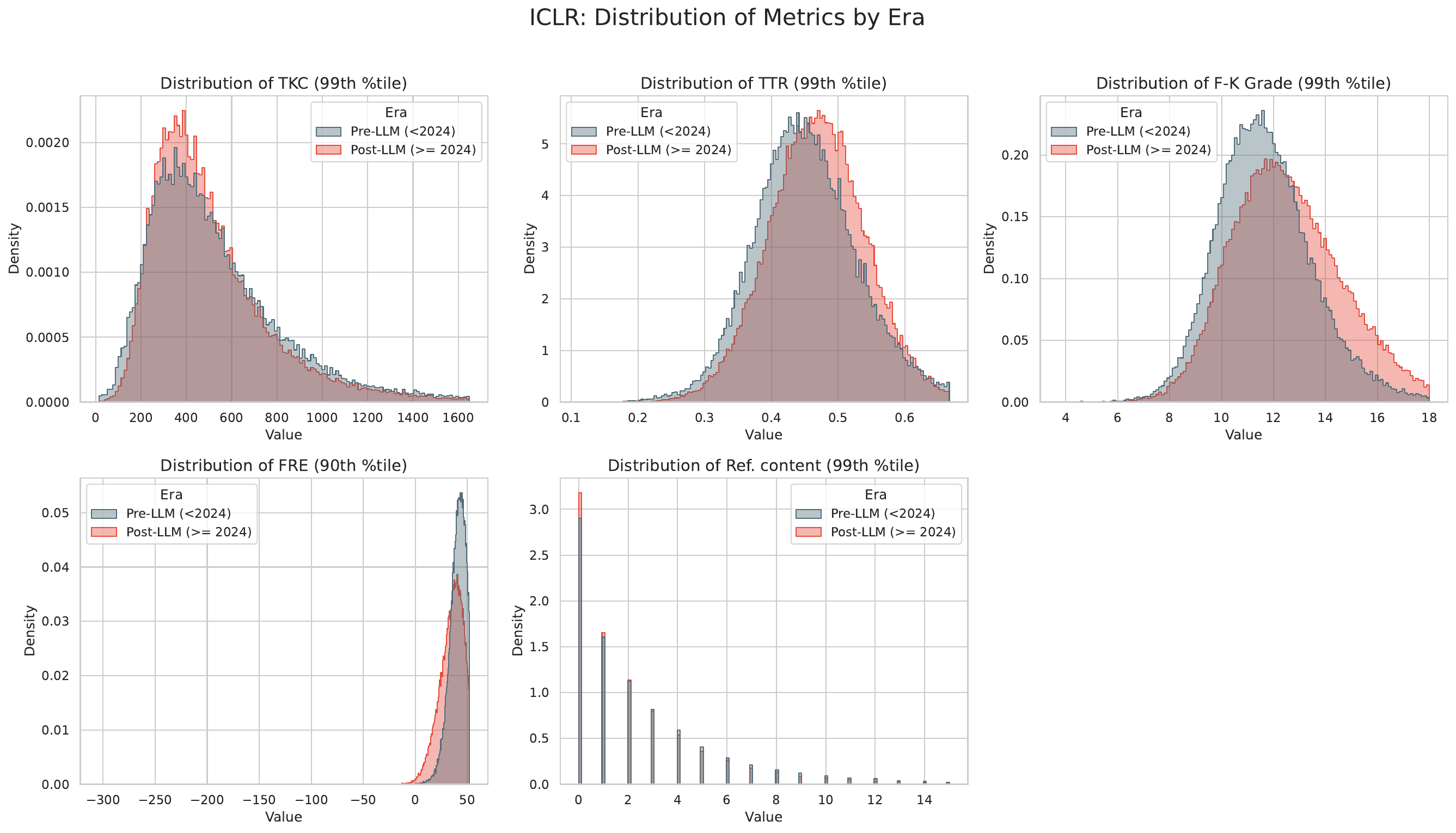}
  \caption{\textbf{ICLR metrics distribution}. All metrics have 99nth percentile, except FRE with 90th due to outliers. Reference to content are displayed as categorical variables. \textit{Ref. content} stands for \textit{Xref.} used in the main article text.}
  \label{fig:iclr_rating_labels}
\end{subfigure}
\hfill
\begin{subfigure}{0.95\textwidth}
  \centering
  \includegraphics[width=1.0\linewidth]{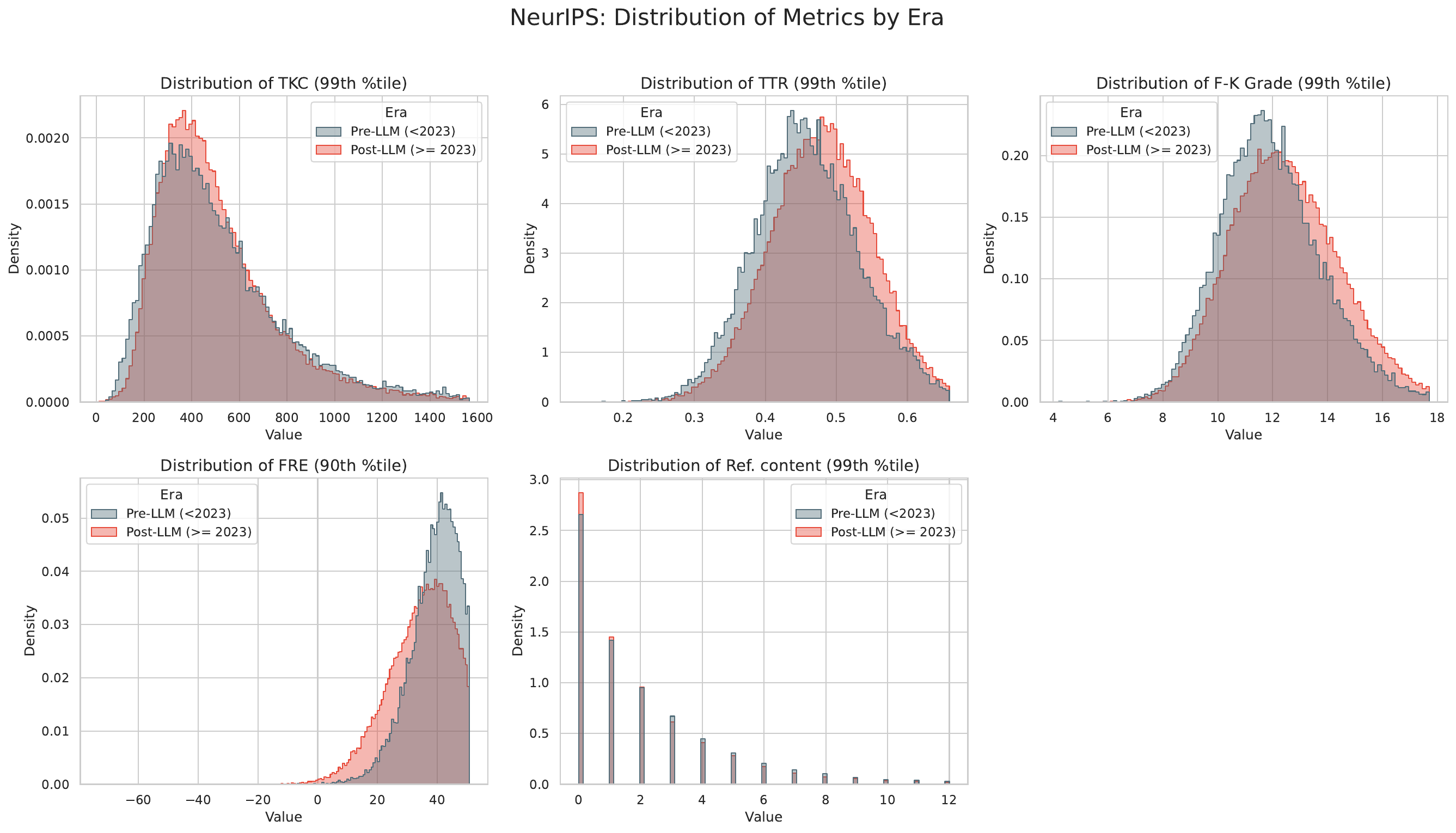}
  \caption{{\textbf{NeurIPS metrics distribution}. Historical distribution of ratings and stylistic markers; note that data is primarily limited to accepted papers for this venue.}}
  \label{fig:neurips_rating_labels}
\end{subfigure}
\caption{Historical distributions of numerical ratings for ICLR and NeurIPS. These distributions highlight how scoring standards and evaluation formats have shifted across different conference cycles.}
\label{fig:distribution_pre_post_llms}
\end{figure*}

The increase in the Flesch-Kincaid Grade Level in the post-LLM era suggests that review text has become theoretically 'more complex' or 'academic' in structure (Figure \ref{fig:distribution_pre_post_llms}). This can be explained by several factors characteristic of Large Language Model (LLM) outputs:

\begin{itemize}
    \item \textbf{Sentence Length:} LLMs tend to produce longer, more syntactically complete sentences with multiple dependent clauses. Since the F-K formula heavily weights the average number of words per sentence, these 'run-on' or complex structures directly inflate the grade level.
    \item \textbf{Syllable Density:} Models often use a more formal, 'standardized' academic vocabulary. They are less likely to use the informal shorthand or sentence fragments that human reviewers often use when writing quickly. A higher average of syllables per word contributes to a higher F-K score.
    \item \textbf{Formulaic Structure:} LLMs often follow a structured template (e.g., 'While the authors demonstrate X, it remains unclear whether Y \dots'). This leads to higher-level connective tissue in the writing that markers of readability interpret as advanced literacy.
    \item \textbf{Lack of Conciseness:} Human experts often use specialized jargon to be concise, whereas LLMs may use verbose explanations to ensure clarity, which paradoxically increases the 'grade level' by increasing word counts per sentence.
\end{itemize}

\textbf{Conference Comparison (ICLR vs. NeurIPS).}
Data from both conferences reveals significant similarities in how the release of ChatGPT has influenced review stylistics. Both venues exhibit the same pattern shift, offset naturally by their respective submission deadlines (as summarized in Table \ref{tab:llm_era_metrics_mean}). Although NeurIPS lacks extended pre-ChatGPT data, a slight decrease in the F-K Grade between 2021 and 2022 is followed by a distinct upward trajectory in the post-ChatGPT era, strongly suggesting the change is attributable to the model's release. Historically, while ICLR has maintained slightly higher reference content counts than NeurIPS, the relative changes in stylistics between the two venues remain consistent.

\begin{table}[t]
\centering
\caption{Mean Metrics Before and After the LLM Era across ICLR and NeurIPS conferences. \textbf{F-KG} stands for F-K Grade. Greater values for a given conference and a dimension are bolded.}
\label{tab:llm_era_metrics_mean}
\begin{tabular}{lrrrr}
\toprule
& \multicolumn{2}{c}{\textbf{ICLR}} & \multicolumn{2}{c}{\textbf{NeurIPS}} \\
\cmidrule(lr){2-3} \cmidrule(lr){4-5}
\textbf{Metric} 
& \multicolumn{1}{c}{Pre-LLM} & \multicolumn{1}{c}{Post-LLM} 
& \multicolumn{1}{c}{Pre-LLM} & \multicolumn{1}{c}{Post-LLM} \\
& \multicolumn{1}{c}{($<$ 2024)} & \multicolumn{1}{c}{($\ge$ 2024)} 
& \multicolumn{1}{c}{($<$ 2023)} & \multicolumn{1}{c}{($\ge$ 2023)} \\
\midrule
\textbf{TKC} & \textbf{538.28} ± 333.23 & 515.09 ± 297.18 & \textbf{527.30} ± 323.84 & 515.62 ± 280.89 \\
\textbf{TTR} & 0.46 ± 0.08 & \textbf{0.47} ± 0.07 & 0.46 ± 0.08 & \textbf{0.48} ± 0.07 \\
\textbf{F-KG} & 11.72 ± 1.94 & \textbf{12.57} ± 2.22 & 12.00 ± 1.94 & \textbf{12.51} ± 2.06 \\
\textbf{FRE} & \textbf{42.38} ± 9.12 & 35.80 ± 12.14 & \textbf{40.95} ± 9.38 & 35.61 ± 11.73 \\
\textbf{Xref.} & \textbf{2.58} ± 3.74 & 2.17 ± 3.09 & \textbf{2.09} ± 2.93 & 1.82 ± 2.60 \\
\bottomrule
\end{tabular}
\end{table}

\paragraph{Statistical test analysis.} 

This statistical summary supports the paper's findings that reviews in the post-ChatGPT era have become significantly more syntactically complex (indicated by the \textbf{F-K Grade} and \textbf{FRE} shifts) while showing a measurable decline in specific \textbf{grounding} via cross-references (Table \ref{tab:stat_comparison}).
\begin{table}[H]
\centering
\caption{Statistical comparison of review metrics (Pre- vs. Post-LLM era). Mean values are reported across conferences, with Mann-Whitney U test results indicating significant differences that validate the proposed hypothesis.}
\label{tab:stat_comparison}
\begin{tabular}{llcccc}
\toprule
\textbf{Conference} & \textbf{Metric} & \textbf{Pre-Mean} & \textbf{Post-Mean} & \textbf{$p$-value} & \textbf{Significant (0.05)} \\ 
\midrule
\textbf{ICLR} & TKC & 538.28 & 515.09 & $< 0.001$ & \textbf{Yes} \\
 & FRE & 42.38 & 35.80 & $< 0.001$ & \textbf{Yes} \\
 & F-K & 11.72 & 12.57 & $< 0.001$ & \textbf{Yes} \\
 & Xref. & 2.58 & 2.17 & $< 0.001$ & \textbf{Yes} \\
\midrule
\textbf{NeurIPS} & TKC & 527.30 & 515.62 & 0.223 & No \\
 & FRE & 40.95 & 35.61 & $< 0.001$ & \textbf{Yes} \\
 & F-K & 12.00 & 12.51 & $< 0.001$ & \textbf{Yes} \\
 & Xref. & 2.09 & 1.82 & $< 0.001$ & \textbf{Yes} \\
\bottomrule
\end{tabular}
\end{table}

\begin{figure*}[ht]
\centering
\begin{subfigure}{0.95\textwidth}
  \centering
  \includegraphics[width=1.0\linewidth]{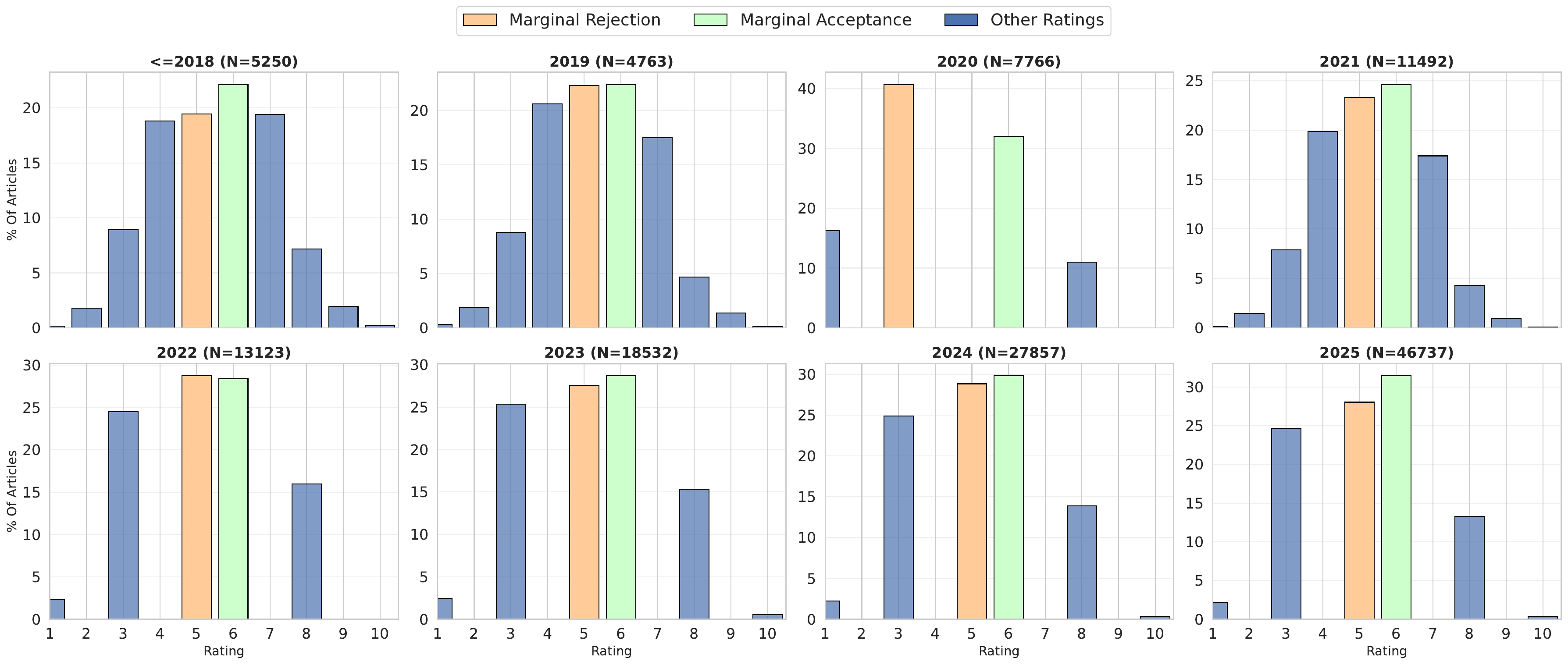}
  \caption{\textbf{ICLR ratings distribution} shows changes in labels used in each year. It is visible that some scores have been out of usage since 2021.}
  \label{fig:iclr_rating_labels}
\end{subfigure}
\hfill
\begin{subfigure}{0.95\textwidth}
  \centering
  \includegraphics[width=1.0\linewidth]{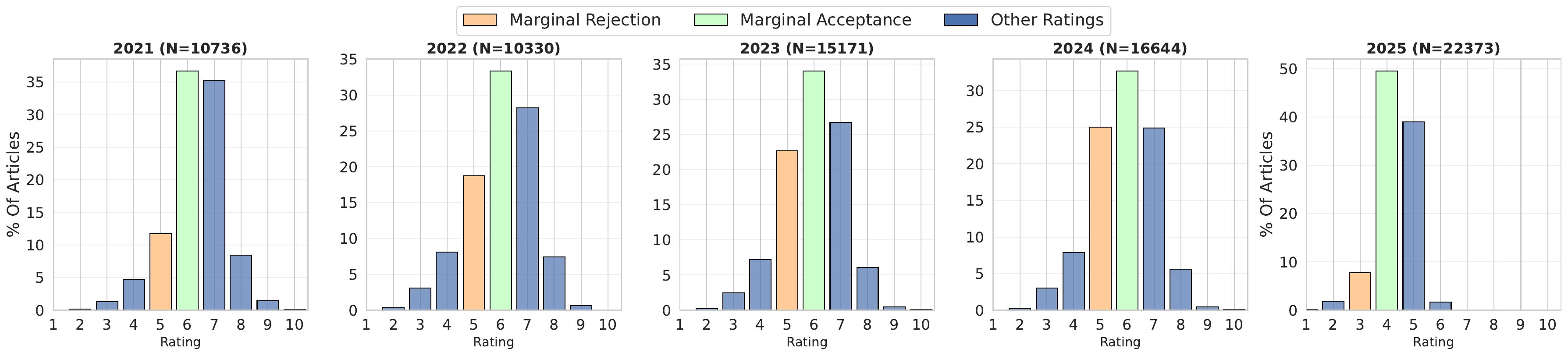}
  \caption{\textbf{NeurIPS ratings distribution} showing the changes in rating scales used across different years. Specifically, starting in 2025, the scale transitioned to a 1–6 range. Note that this distribution reflects only accepted papers due to data unavailability for rejected submissions.}
  \label{fig:neurips_rating_labels}
\end{subfigure}
\caption{ Ratings distribution through time for ICLR and NeurIPS. \textbf{N} denotes the number of reviews with scores. }
\label{fig:rating_labels}
\end{figure*}

\begin{table}[h]
\centering
\caption{Number of reviews sampled per conference and year.}
\label{tab:paper_counts}
\begin{tabular}{llr}
\toprule
\textbf{Conference} & \textbf{Year} & \textbf{\# Reviews} \\
\midrule
\multirow{5}{*}{ICLR}
  & 2021 & 377 \\
  & 2022 & 390 \\
  & 2023 & 376 \\
  & 2024 & 387 \\
  & 2025 & 398 \\
\cmidrule(lr){1-3}
\multirow{5}{*}{NeurIPS}
  & 2021 & 388 \\
  & 2022 & 369 \\
  & 2023 & 438 \\
  & 2024 & 381 \\
  & 2025 & 406 \\
\cmidrule(lr){1-3}
\textbf{Total} & & \textbf{3{,}910} \\
\bottomrule
\end{tabular}
\end{table}

\begin{table}[h!]
\centering
\caption{Number of generated reviews per conference, year, model, and
  prompt type. Each cell contains 100 reviews. Models: DeepSeek,
  Gemma, GPT-5, OpenReviewer, Qwen. Prompt types:
  \textsc{Ext} = \texttt{extended\_prompt},
  \textsc{Guide} = \texttt{guidelines\_prompt},
  \textsc{OR} = \texttt{open\_reviewer\_prompt}
  (OpenReviewer only).}
\label{tab:generated_counts}
\setlength{\tabcolsep}{5pt}
\begin{tabular}{llccccc}
\toprule
 & & \multicolumn{5}{c}{\textbf{Reviews per model (Ext / Guide [/ OR])}} \\
\cmidrule(lr){3-7}
\textbf{Conference} & \textbf{Year}
  & \textbf{DeepSeek} & \textbf{Gemma} & \textbf{GPT-5}
  & \textbf{OpenReviewer} & \textbf{Qwen} \\
\midrule
\multirow{5}{*}{ICLR}
  & 2021 & 100 / 100 & 100 / 100 & 100 / 100 & 100 / 100 / 100 & 100 / 100 \\
  & 2022 & 100 / 100 & 100 / 100 & 100 / 100 & 100 / 100 / 100 & 100 / 100 \\
  & 2023 & 100 / 100 & 100 / 100 & 100 / 100 & 100 / 100 / 100 & 100 / 100 \\
  & 2024 & 100 / 100 & 100 / 100 & 100 / 100 & 100 / 100 / 100 & 100 / 100 \\
  & 2025 & 100 / 100 & 100 / 100 & 100 / 100 & 100 / 100 / 100 & 100 / 100 \\
\cmidrule(lr){1-7}
\multirow{5}{*}{NeurIPS}
  & 2021 & 100 / 100 & 100 / 100 & 100 / 100 & 100 / 100 / 100 & 100 / 100 \\
  & 2022 & 100 / 100 & 100 / 100 & 100 / 100 & 100 / 100 / 100 & 100 / 100 \\
  & 2023 & 100 / 100 & 100 / 100 & 100 / 100 & 100 / 100 / 100 & 100 / 100 \\
  & 2024 & 100 / 100 & 100 / 100 & 100 / 100 & 100 / 100 / 100 & 100 / 100 \\
  & 2025 & 100 / 100 & 100 / 100 & 100 / 100 & 100 / 100 / 100 & 100 / 100 \\
\bottomrule
\end{tabular}
\end{table}

\newpage

\section{Regular expressions} \label{appendix:regular_expressions}

To find specified patten and in result information in analyzed reviews we use regular expressions.

\subsection{Paper content} \label{papercontentregex}

To extract references in reviews to specific elements of a manuscript (e.g., figures, equations, and theorems), we use the regular expression shown below.

\begin{lstlisting}[caption={Regular expression used to identify references to specific paper elements (e.g., figures, tables, equations, theorems).}, label={lst:reference_regex}]
reference_regex = (
    r"[\s(^\[]("
    r"("
    r"[Ff]ig|[Ff]igures?|S|[Ss]ec|[Ss]ections?|[Ss]sec|[Ss]ub[\s-]?sections?|"
    r"[Ee]q|[Ee]quations?|[Tt]ab|[Tt]ables?|[Dd]ef|[Dd]efinitions?|[Dd][Bb]|"
    r"[Dd]atabases?|[Dd][Ss]|[Dd]ata\s?sets?|[Tt]hm|[Tt]heorems?|[Ll]em|"
    r"[Ll]emmas?|[Cc]or|[Cc]orollar(y|ies)|[Pp]|[Pp]ages?|like(ly)?"
    r")"
    r"\.?\s?"
    r"("
    r"[0-9]+(\.[0-9]+)*|\([0-9]+(\.[0-9]+)*\)"
    r")"
    r"|in\s[0-9]+(\.[0-9]+)+"
    r")"
)
\end{lstlisting}

When designing a regular expression to capture references to specific elements within a review, we encounter a key challenge: reviewers do not follow a standardized convention for referring to such elements. As a result, references may appear in a wide variety of forms.

The proposed regular expression is designed to capture the majority of these cases, including the most common patterns observed in practice. Consequently, it enables the extraction of references to elements such as figures, sections, subsections, theorems, lemmas, pages, corollaries, equations, definitions, tables, and datasets.

\subsection{Citations}  \label{appendix:citations}

When constructing regular expressions to capture citations in reviews, we encountered a challenge similar to that observed for references to manuscript elements: reviewers often do not adhere to standardized citation formats. In many cases, citations are incomplete—for example, including only the author(s) and publication year without the article title.

We decided that the proposed regular expressions successfully capture citations in common formats such as MLA, APA, and IEEE, while also remaining robust to minor formatting inconsistencies.

\label{citationsregex}
\begin{lstlisting}[caption={Regular expressions utilized for the extraction of various citation formats from the review corpus.}, label={lst:regex_patterns}, mathescape=true]
MLA_regex = (
    r"(?P<authors>(([A-Z][a-z$\ddot{u}ddot{o}$]*.?,?[\s\-])*(and\s([A-Z][a-z$\ddot{u}\ddot{o}$]*.?,?\s){2})?)(et al\.\s)?)"
    r"\"?(?P<title>[A-Za-z0-9\s\:\-\(\),]*)\.\"?"
    r"\s([A-Za-z\s:-]*),(\svol\. \d*,)?\s(no\. \d*,\s)?"
    r"(?P<year>\d{4})"
    r"(,\spp\.\s\d*.\d*)?\."
)

APA_regex = (
    r"(?P<authors>(([A-Z][a-z$\ddot{u}\ddot{o}$]*\.?,?[\s\-])*(&\s([A-Z][a-z$\ddot{u}\ddot{o}$]*.?,?\s){2})?)(et al\.\s)?)"
    r"(?P<year>\(\d{4}\))\."
    r"(?P<title>[A-Za-z0-9\s\:\-\(\),]*)\."
)

IEEE_citation = (
    r"\[\d*\]\s"
    r"(?P<authors>(([A-Z][a-z$\ddot{u}\ddot{o}$]*.?,?[\s\-])*(and\s([A-Z][a-z$\ddot{u}\ddot{o}$]*.?,?\s){2})?)(et al\.,?\s)?)"
    r"\"?(?P<title>[A-Za-z0-9\s\:\-\(\),]*),?\.?\"?"
    r"\s.*\(?"
    r"(?P<year>\d{4})\)?\."
)
\end{lstlisting}

\section{Krippendorff's \texorpdfstring{$\alpha$}{alpha}}
\subsection{Mathematical Formulation of Krippendorff's \texorpdfstring{$\alpha$}{alpha}}

\label{sec:krippendorfss_alpha}

To robustly measure inter-annotator agreement between human reviewers and LLMs, we compute Krippendorff's $\alpha$ with an ordinal level of measurement. This metric accommodates varying numbers of annotators per paper (missing data) and penalizes disagreements based on the ordered nature of the rating scale. 

The general form of Krippendorff's $\alpha$ is defined as:
\begin{equation*}
    \alpha = 1 - \frac{D_o}{D_e}
\end{equation*}
where $D_o$ represents the observed disagreement among annotators, and $D_e$ represents the expected disagreement by chance. 

Let $N$ be the total number of papers, and $c$ and $k$ represent distinct rating categories. We define $m_u$ as the number of valid ratings assigned to paper $u$, and $n_{uc}$ as the number of times rating $c$ is assigned to paper $u$. The total frequency of category $c$ across the entire dataset is $n_c$, and the total number of all valid ratings is $n$.

The observed disagreement $D_o$ evaluates differences in ratings assigned to the exact same paper:
{\small
\begin{equation*}
    D_o = \frac{1}{\sum_{u=1}^N m_u(m_u - 1)} \sum_{u=1}^N \sum_{c} \sum_{k} n_{uc} n_{uk} \delta^2(c, k)
\end{equation*}
}

The expected disagreement $D_e$ estimates the baseline divergence if annotators were to assign ratings randomly, constrained only by the marginal distribution of the rating categories:
\begin{equation*}
    D_e = \frac{1}{n(n - 1)} \sum_{c} \sum_{k} n_c n_k \delta^2(c, k)
\end{equation*}

Because our rating scale is ordered, we apply the \textbf{ordinal distance function} $\delta^2(c, k)$. Unlike nominal or interval scales, the ordinal distance between two ratings $c$ and $k$ (where $c \le k$) depends on the cumulative frequency of all ratings that fall between them in the dataset:
\begin{equation*}
    \delta^2_{\text{ordinal}}(c, k) = \left( \sum_{g=c}^{k} n_g - \frac{n_c + n_k}{2} \right)^2
\end{equation*}
This ensures that the penalty for a disagreement (e.g., a human rating of 4 versus an LLM rating of 6) dynamically scales with the empirical density of the intermediate ratings, providing a highly sensitive measure of alignment for peer review scores.

We compute the measure for each year $y_k$ separately (due to the different rating distributions). We then average the score to obtain a single value for a given conference:
\begin{equation*}
    \alpha = \frac{1}{m}\sum_{k=1}^{m}\alpha_{y_m}
\end{equation*}

\subsection{Krippendorff's \texorpdfstring{$\alpha$}{alpha} for Rating Agreement: Human-Only vs.\ LLM-Augmented Panels}

\label{sec:appendix:krippendorff_rating}

Table~\ref{tab:kipp_all} report Krippendorff's
$\alpha$ (ordinal) for \texttt{num\_rating} computed in two conditions:
\emph{human only} ($\alpha_H$), using only the original OpenReview panel, and
\emph{with LLM} ($\alpha_{H+L}$), where the LLM reviewer is added as an
additional annotator. All values reported here use the \texttt{extended\_prompt}.
A drop from $\alpha_H$ to $\alpha_{H+L}$ indicates that the LLM's ratings
diverge from the human consensus, reducing overall panel agreement; a value
of $\alpha_{H+L} \geq \alpha_H$ indicates that the LLM is at least as
consistent with the panel as the human reviewers themselves.

\paragraph{Baseline human agreement.}
Human-only $\alpha$ varies by conference and year. ICLR panels are generally
more consistent ($\alpha_H = 0.320$--$0.367$) than NeurIPS panels
($\alpha_H = 0.176$--$0.261$), reflecting either tighter review guidelines
or a more homogeneous reviewer pool at ICLR. Both venues show a mild downward
trend across years, with 2022 NeurIPS being a notable low point
($\alpha_H = 0.176$). All human-only values fall in the ``slight to fair''
agreement range under standard Krippendorff interpretations, consistent with
the well-documented low inter-rater reliability in academic peer review \cite{bornmann2010reliability}.

\paragraph{GPT-5: the only model that matches or exceeds human agreement.}
GPT-5 is the sole model whose $\alpha_{H+L}$ is consistently at or above
$\alpha_H$. On ICLR it exceeds the human baseline in 2021 ($0.390$ vs.\
$0.367$) and remains within $0.004$ of it in 2024 ($0.354$ vs.\ $0.345$),
falling only slightly below in later years. On NeurIPS the pattern is even
clearer: $\alpha_{H+L}$ exceeds $\alpha_H$ in all five years (e.g.\ $0.297$
vs.\ $0.261$ in 2021; $0.253$ vs.\ $0.176$ in 2022). This means GPT-5 ratings
are, on average, \emph{more} consistent with the human panel than an additional
human reviewer would be--a remarkable result that suggests GPT-5's rating
behaviour under the extended prompt is well-aligned with ICLR and NeurIPS
reviewer norms.

\paragraph{OpenReviewer: near-human agreement, conference-dependent.}
OpenReviewer achieves $\alpha_{H+L}$ close to $\alpha_H$ on both venues.
On ICLR it slightly reduces agreement in most years (e.g.\ $0.332$ vs.\
$0.367$ in 2021; $0.267$ vs.\ $0.320$ in 2025), but on NeurIPS it exceeds the
human baseline in 2022 ($0.188$ vs.\ $0.176$) and 2024 ($0.266$ vs.\ $0.241$).
The drop on ICLR is modest ($\Delta \leq 0.053$), substantially smaller than
for DeepSeek or Qwen, confirming that domain-specific fine-tuning produces
ratings that integrate reasonably well into a human panel.

\paragraph{Gemma and Qwen: moderate disruption.}
Both models reduce $\alpha$ relative to the human baseline by a consistent
margin. On ICLR, Gemma's $\alpha_{H+L}$ ranges from $0.141$ (2021) to $0.244$
(2023), corresponding to drops of $0.089$-$0.226$ below $\alpha_H$. Qwen
shows a similar pattern ($0.148$-$0.263$ on ICLR, drops of $0.091$--$0.219$).
On NeurIPS both models are closer to the human baseline in absolute terms,
partly because $\alpha_H$ itself is lower. The average reduction across all
years is approximately $0.12$-$0.13$ for both models on ICLR and $0.06$-$0.09$
on NeurIPS, indicating moderate but not catastrophic misalignment.

\paragraph{DeepSeek: largest and most consistent drop.}
DeepSeek produces the most severe reductions in panel agreement. On ICLR its
$\alpha_{H+L}$ falls between $0.104$ and $0.164$, compared to human-only values
of $0.320$-$0.367$-an average drop of approximately $0.21$. On NeurIPS the
absolute values are lower ($0.072$-$0.153$) but the proportional reduction
relative to $\alpha_H$ is equally severe ($\Delta = 0.104$-$0.153$). The
2022 NeurIPS result ($\alpha_{H+L} = 0.072$) is the lowest value in the
entire table, indicating near-chance agreement when DeepSeek is included in
the panel. This is consistent with DeepSeek's extreme concentration at score
8 (Table~\ref{tab:rating_diff_human}): a near-constant reviewer provides no
discriminative information and suppresses the ordinal agreement signal.

\paragraph{Temporal trends.}
Across both conferences, $\alpha_{H+L}$ tends to decrease from 2021 to 2025
for most models, mirroring a similar (though smaller) decline in $\alpha_H$.
This may reflect increasing diversity in the human reviewer pool over time,
or changes in paper difficulty distribution. GPT-5 is the exception, showing
relatively stable $\alpha_{H+L}$ across years on NeurIPS.

\begin{table}[H]
\caption{Krippendorff's $\alpha$ (ordinal) for \texttt{num\_rating},
extended prompt. $\alpha_{H}$: human reviewers only;
$\alpha_{H+L}$: human panel augmented with the LLM reviewer;
$\Delta = \alpha_{H+L} - \alpha_{H}$, negative values indicate the LLM
reduces panel agreement.}
\label{tab:kipp_all}
\centering
\small
\begin{tabular}{ll r rr}
\toprule
\textbf{Conference} & \textbf{Year} & $\alpha_{H}$ & $\alpha_{H+L}$ & $\Delta$ \\
\midrule
\multicolumn{5}{l}{\textit{DeepSeek}} \\
\quad ICLR    & 2021 & $0.367$ & $0.122$ & $-0.245$ \\
\quad ICLR    & 2022 & $0.359$ & $0.164$ & $-0.195$ \\
\quad ICLR    & 2023 & $0.333$ & $0.135$ & $-0.198$ \\
\quad ICLR    & 2024 & $0.345$ & $0.154$ & $-0.191$ \\
\quad ICLR    & 2025 & $0.320$ & $0.104$ & $-0.216$ \\
\quad NeurIPS & 2021 & $0.261$ & $0.145$ & $-0.116$ \\
\quad NeurIPS & 2022 & $0.176$ & $0.072$ & $-0.104$ \\
\quad NeurIPS & 2023 & $0.257$ & $0.104$ & $-0.153$ \\
\quad NeurIPS & 2024 & $0.241$ & $0.114$ & $-0.127$ \\
\quad NeurIPS & 2025 & $0.257$ & $0.153$ & $-0.104$ \\
\midrule
\multicolumn{5}{l}{\textit{Gemma}} \\
\quad ICLR    & 2021 & $0.367$ & $0.141$ & $-0.226$ \\
\quad ICLR    & 2022 & $0.359$ & $0.243$ & $-0.116$ \\
\quad ICLR    & 2023 & $0.333$ & $0.244$ & $-0.089$ \\
\quad ICLR    & 2024 & $0.345$ & $0.228$ & $-0.117$ \\
\quad ICLR    & 2025 & $0.320$ & $0.207$ & $-0.113$ \\
\quad NeurIPS & 2021 & $0.261$ & $0.145$ & $-0.116$ \\
\quad NeurIPS & 2022 & $0.176$ & $0.096$ & $-0.080$ \\
\quad NeurIPS & 2023 & $0.257$ & $0.161$ & $-0.096$ \\
\quad NeurIPS & 2024 & $0.241$ & $0.091$ & $-0.150$ \\
\quad NeurIPS & 2025 & $0.257$ & $0.226$ & $-0.031$ \\
\midrule
\multicolumn{5}{l}{\textit{GPT-5}} \\
\quad ICLR    & 2021 & $0.367$ & $0.390$ & $+0.023$ \\
\quad ICLR    & 2022 & $0.359$ & $0.376$ & $+0.017$ \\
\quad ICLR    & 2023 & $0.333$ & $0.360$ & $+0.027$ \\
\quad ICLR    & 2024 & $0.345$ & $0.354$ & $+0.009$ \\
\quad ICLR    & 2025 & $0.320$ & $0.324$ & $+0.004$ \\
\quad NeurIPS & 2021 & $0.261$ & $0.297$ & $+0.036$ \\
\quad NeurIPS & 2022 & $0.176$ & $0.253$ & $+0.077$ \\
\quad NeurIPS & 2023 & $0.257$ & $0.256$ & $-0.001$ \\
\quad NeurIPS & 2024 & $0.241$ & $0.240$ & $-0.001$ \\
\quad NeurIPS & 2025 & $0.257$ & $0.262$ & $+0.005$ \\
\midrule
\multicolumn{5}{l}{\textit{OpenReviewer}} \\
\quad ICLR    & 2021 & $0.367$ & $0.332$ & $-0.035$ \\
\quad ICLR    & 2022 & $0.359$ & $0.332$ & $-0.027$ \\
\quad ICLR    & 2023 & $0.333$ & $0.351$ & $+0.018$ \\
\quad ICLR    & 2024 & $0.345$ & $0.311$ & $-0.034$ \\
\quad ICLR    & 2025 & $0.320$ & $0.267$ & $-0.053$ \\
\quad NeurIPS & 2021 & $0.261$ & $0.250$ & $-0.011$ \\
\quad NeurIPS & 2022 & $0.176$ & $0.188$ & $+0.012$ \\
\quad NeurIPS & 2023 & $0.257$ & $0.215$ & $-0.042$ \\
\quad NeurIPS & 2024 & $0.241$ & $0.266$ & $+0.025$ \\
\quad NeurIPS & 2025 & $0.257$ & $0.199$ & $-0.058$ \\
\midrule
\multicolumn{5}{l}{\textit{Qwen}} \\
\quad ICLR    & 2021 & $0.367$ & $0.148$ & $-0.219$ \\
\quad ICLR    & 2022 & $0.359$ & $0.263$ & $-0.096$ \\
\quad ICLR    & 2023 & $0.333$ & $0.242$ & $-0.091$ \\
\quad ICLR    & 2024 & $0.345$ & $0.223$ & $-0.122$ \\
\quad ICLR    & 2025 & $0.320$ & $0.204$ & $-0.116$ \\
\quad NeurIPS & 2021 & $0.261$ & $0.169$ & $-0.092$ \\
\quad NeurIPS & 2022 & $0.176$ & $0.159$ & $-0.017$ \\
\quad NeurIPS & 2023 & $0.257$ & $0.202$ & $-0.055$ \\
\quad NeurIPS & 2024 & $0.241$ & $0.129$ & $-0.112$ \\
\quad NeurIPS & 2025 & $0.257$ & $0.203$ & $-0.054$ \\
\bottomrule
\end{tabular}
\end{table}

\section{Stylistic and specificity measures} \label{appendix:stylistic_specificity_ablations}

\begin{table}[ht]
\centering
\caption{Readability formulae employed in this study. \textit{Legend:} $W$ -- total words, $S$ -- total sentences, $Sy$ -- total syllables.}
\label{tab:readabilityformulae}
\renewcommand{\arraystretch}{1.5} 
\small
\begin{tabular}{ll}
\toprule
\textbf{Metric} & \textbf{Formula} \\ 
\midrule
\textbf{FRE} & $206.835 - 1.015 \left( \frac{W}{S} \right) - 84.6 \left( \frac{Sy}{W} \right)$ \\
\textbf{F-K Grade} & $0.39 \left( \frac{W}{S} \right) + 11.8 \left( \frac{Sy}{W} \right) - 15.59$ \\
\bottomrule
\end{tabular}
\vspace{0.15cm} 
\end{table}

\paragraph{Effect of prompt type on review characteristics.}
Across all models, switching from \texttt{guidelines\_prompt} to
\texttt{extended\_prompt} consistently increases review length, reduces lexical
diversity, and raises cross-reference counts, while readability and mathematical
content show smaller, model-dependent shifts
(Tables ~\ref{tab:prompt_comparison}, ~\ref{tab:prompt_comparison_delta}).

The effect on \textbf{length} is the most pronounced and model-dependent
difference. GPT-5 produces the largest absolute increase ($\Delta\text{TKC} =
+413$ tokens), followed by Qwen ($+253$) and Gemma ($+219$), while
OpenReviewer is nearly unaffected ($+53$). This suggests that models with
stronger instruction-following capabilities are more sensitive to the additional
context supplied by the extended prompt, generating proportionally more content.
DeepSeek and Gemma remain well around the human baseline ($531$ tokens) even
under the extended prompt ($637$ and $898$ tokens respectively), whereas GPT-5
and Qwen substantially exceed it ($1449$ and $1120$ tokens).

\textbf{Lexical diversity} (TTR) decreases under the extended prompt for every
model, with DeepSeek showing the largest drop ($-0.052$) and OpenReviewer the
smallest ($-0.007$). Because longer texts mechanically tend toward lower TTR,
part of this effect is an artefact of the length increase; nonetheless, the
consistent direction of the change indicates that the extended prompt does not
elicit more varied vocabulary, contrary to what one might expect from richer
instructions.

\textbf{Readability} (Flesch Reading Ease, FRE) improves marginally under the
extended prompt for all models, with Qwen benefiting most ($+4.5$ points) and
OpenReviewer least ($+1.3$ points). Even so, all LLM-generated reviews remain
far below the human baseline ($39.1$), with DeepSeek and Gemma scoring below
$15$ regardless of prompt---indicating highly complex, difficult-to-read prose
that is atypical of the human reviews in the dataset.

\textbf{Cross-references} increase notably for Qwen ($+2.39$) and GPT-5
($+0.88$) under the extended prompt, again with OpenReviewer the least affected
($+0.42$). Only OpenReviewer and human reviewers produce cross-reference counts
in a comparable range ($2.16$--$2.77$), while DeepSeek and Gemma generate far
fewer under both prompts, suggesting these models are less inclined to cite
related work regardless of instructions.

\textbf{Mathematical content} shows the smallest prompt sensitivity
overall. DeepSeek ($+0.09$) and Gemma ($+0.16$) exhibit the largest relative
increases, whereas Qwen and GPT-5---which already flag a high proportion of
reviews as math-heavy under the guidelines prompt ($0.67$ and $0.70$
respectively)---show almost no further change ($+0.01$ and $+0.06$). This
ceiling effect implies that the math-detection signal is driven primarily by
model identity rather than prompt wording.

Taken together, the results indicate that the \texttt{extended\_prompt}
amplifies surface-level stylistic features (length, cross-references) more than
it alters substantive review content (math coverage, citation practice). The
relative insensitivity of OpenReviewer across all metrics is consistent with its
fine-tuned, domain-specific training, which likely overrides generic prompt
instructions to a greater degree than the general-purpose models.

\begin{table}[H]
\centering
\caption{Model Performance Metrics by Prompt Type (Mean $\pm$ SD)}
\label{tab:prompt_comparison}
\resizebox{\textwidth}{!}{
\begin{tabular}{llccccccc}
\toprule
\textbf{Model} & \textbf{Prompt Type} & \textbf{Tokens} & \textbf{TTR} & \textbf{Flesch Ease (FRE)} & \textbf{Xref.} & \textbf{Citations} & \textbf{Verified} & \textbf{Math} \\
\midrule
OpenReview & Human & $531.57 \pm 301.17$ & $0.460 \pm 0.075$ & $39.14 \pm 10.98$ & $2.26 \pm 3.17$ & --- & --- & $0.43 \pm 0.50$ \\
\midrule
\multirow{2}{*}{DeepSeek} & Extended & $637.05 \pm 124.01$ & $0.413 \pm 0.057$ & $13.74 \pm 7.35$ & $0.84 \pm 1.61$ & $0.00 \pm 0.00$ & $0.00 \pm 0.00$ & $0.29 \pm 0.45$ \\
 & Guidelines & $469.56 \pm 74.31$ & $0.465 \pm 0.033$ & $10.95 \pm 7.13$ & $0.15 \pm 0.72$ & $0.00 \pm 0.00$ & $0.00 \pm 0.00$ & $0.20 \pm 0.40$ \\
\midrule
\multirow{2}{*}{Gemma} & Extended & $897.50 \pm 84.85$ & $0.387 \pm 0.020$ & $14.97 \pm 5.56$ & $0.74 \pm 1.14$ & $0.00 \pm 0.00$ & $0.00 \pm 0.00$ & $0.33 \pm 0.47$ \\
 & Guidelines & $678.58 \pm 58.28$ & $0.410 \pm 0.020$ & $11.67 \pm 5.69$ & $0.25 \pm 0.64$ & $0.00 \pm 0.00$ & $0.00 \pm 0.00$ & $0.17 \pm 0.38$ \\
\midrule
\multirow{2}{*}{GPT-5-1} & Extended & $1449.23 \pm 221.97$ & $0.388 \pm 0.026$ & $21.28 \pm 6.07$ & $2.47 \pm 2.96$ & $0.00 \pm 0.00$ & $0.00 \pm 0.00$ & $0.76 \pm 0.43$ \\
 & Guidelines & $1035.88 \pm 142.06$ & $0.434 \pm 0.030$ & $18.48 \pm 6.72$ & $1.59 \pm 2.24$ & $0.00 \pm 0.00$ & $0.00 \pm 0.00$ & $0.70 \pm 0.46$ \\
\midrule
\multirow{3}{*}{\shortstack[l]{Open\\Reviewer}} & Extended & $677.94 \pm 396.46$ & $0.433 \pm 0.081$ & $38.33 \pm 10.35$ & $2.77 \pm 3.69$ & $0.19 \pm 0.96$ & $0.06 \pm 0.43$ & $0.40 \pm 0.49$ \\
 & Guidelines & $625.09 \pm 375.99$ & $0.440 \pm 0.079$ & $37.00 \pm 11.18$ & $2.35 \pm 3.44$ & $0.14 \pm 0.59$ & $0.05 \pm 0.28$ & $0.38 \pm 0.48$ \\
 & Original & $588.87 \pm 336.51$ & $0.440 \pm 0.071$ & $39.04 \pm 10.57$ & $2.16 \pm 3.56$ & $0.12 \pm 0.61$ & $0.04 \pm 0.29$ & $0.27 \pm 0.44$ \\
\midrule
\multirow{2}{*}{Qwen} & Extended & $1120.34 \pm 168.27$ & $0.388 \pm 0.030$ & $20.13 \pm 6.05$ & $4.96 \pm 3.53$ & $0.00 \pm 0.00$ & $0.00 \pm 0.00$ & $0.69 \pm 0.46$ \\
 & Guidelines & $867.27 \pm 115.51$ & $0.424 \pm 0.028$ & $15.58 \pm 6.09$ & $2.57 \pm 2.57$ & $0.00 \pm 0.00$ & $0.00 \pm 0.00$ & $0.67 \pm 0.47$ \\
\bottomrule
\end{tabular}
}
\end{table}

\begin{table*}[ht]
\caption{Comparison of prompting strategies. Values represent the absolute change ($\Delta$) in average metric scores when switching from the \texttt{guidelines\_prompt} to the \texttt{extended\_prompt}. Positive values indicate an increase under the extended prompt.}
\label{tab:prompt_comparison_delta}
\centering
\small
\begin{tabular}{l rrrrrr}
\toprule
\textbf{Model} & $\Delta$ \textbf{TKC} & $\Delta$ \textbf{TTR} & $\Delta$ \textbf{FRE} & \textbf{Xref.} & $\Delta$ \textbf{Math} \\
\midrule
DeepSeek & +167.4 & -0.052 & +2.8 &  +0.69 & +0.09 \\
Gemma    & +218.9 & -0.023 & +3.3 & +0.49 & +0.16 \\
GPT-5    & +413.3 & -0.046 & +2.8 &  +0.88 & +0.06 \\
OpenReviewer & +52.8 & -0.007 & +1.3 & +0.42 & +0.02 \\
Qwen     & +253.0 & -0.036 & +4.5 & +2.39 & +0.01 \\
\bottomrule
\end{tabular}
\end{table*}

\begin{table*}[ht]
\centering
\caption{Comparative behavioral metrics for LLM-generated reviews (expressed as ratios relative to the human baseline), including aggregated model statistics.}
\label{tab:model_behavior_ratios}
\small
\setlength{\tabcolsep}{5pt}
\begin{tabular}{l ccccc}
\toprule
\textbf{Model} & \textbf{Tokens} & \textbf{TTR} & \textbf{FRE} & \textbf{Xref.} & \textbf{Math} \\
\midrule
DeepSeek      & 1.20 & 0.90 & 0.35 & 0.37 & 0.67 \\
Gemma         & 1.69 & 0.84 & 0.38 & 0.33 & 0.77 \\
GPT-5         & 2.73 & 0.84 & 0.54 & 1.09 & 1.77 \\
OpenReviewer  & 1.28 & 0.94 & 0.98 & 1.22 & 0.94 \\
Qwen          & 2.11 & 0.84 & 0.51 & 2.20 & 1.61 \\
\midrule
\textbf{Mean $\pm$ Std} & \boldmath$1.80 \pm 0.63$ & \boldmath$0.87 \pm 0.04$ & \boldmath$0.55 \pm 0.25$ & \boldmath$1.04 \pm 0.76$ & \boldmath$1.15 \pm 0.50$ \\
\bottomrule
\end{tabular}
\end{table*}

\section{Confidence Score Analysis} \label{appendix:confidence_score}

We present a systematic analysis of confidence scores, characterizing the distributional divergence between Large Language Models (LLMs) and human baselines. Furthermore, we evaluate the stability of these scores across diverse prompting configurations to assess model calibration and robustness.

\subsection{Confidence Score Calibration: LLM vs.\ Human}
\label{sec:appendix:confidence_calibration}

Table~\ref{tab:conf_diff} reports the mean signed difference between each
LLM's confidence score and the median human confidence for the same paper
($\Delta = \text{LLM} - \text{human median}$). All values are positive,
confirming a universal \emph{overconfidence bias}: no model–prompt combination
produces confidence estimates that, on average, fall below the human median.

\paragraph{Cross-model patterns.}
Qwen is a clear outlier, exceeding human confidence by more than a full point
under the guidelines prompt (ICLR: $+1.12$; NeurIPS: $+1.15$), the largest
discrepancy observed. OpenReviewer is the second most over-confident model
($+0.53$--$+0.68$ across settings), despite its domain-specific fine-tuning.
DeepSeek and Gemma occupy an intermediate range ($+0.11$--$+0.40$), while
GPT-5 shows the most variable behavior: it is nearly calibrated under the
guidelines prompt on ICLR ($+0.04$), yet substantially over-confident under
the extended prompt on NeurIPS ($+0.39$).

\paragraph{Prompt sensitivity.}
The direction of the prompt effect is inconsistent across models, which is
notable given the uniform directional effects observed for stylometric metrics
(Section~\ref{sec:results}, Appendix ~\ref{appendix:stylistic_specificity_ablations}).
For Qwen the guidelines prompt produces \emph{noticeably} higher over-confidence
than the extended prompt ($+1.12$ vs.\ $+0.50$ on ICLR; $+1.15$ vs.\ $+0.86$
on NeurIPS), suggesting that concise instructions push the model toward extreme
confidence values. The opposite holds for GPT-5 on NeurIPS ($+0.15$ guidelines
vs.\ $+0.39$ extended) and for Gemma on NeurIPS ($+0.37$ guidelines vs.\
$+0.11$ extended). OpenReviewer and DeepSeek are relatively stable across
prompts ($|\Delta_{\text{ext}} - \Delta_{\text{guide}}| < 0.07$), implying
their confidence outputs are less sensitive to prompt wording.

\paragraph{Conference effect.}
NeurIPS reviews elicit slightly larger over-confidence biases than ICLR reviews
for most models, which may reflect that NeurIPS papers are perceived as
marginally harder to assess, leading LLMs to anchor at higher confidence as a
default.

\begin{table}[H]
\centering
\caption{Mean confidence difference (LLM -- human median) by conference,
model, and prompt type. Positive values indicate LLM over-confidence.
Rows are sorted by model and conference for readability.}
\label{tab:conf_diff}
\small
\begin{tabular}{llcc}
\toprule
\textbf{Model} & \textbf{Prompt} & \textbf{ICLR} & \textbf{NeurIPS} \\
\midrule
\multirow{2}{*}{DeepSeek}
  & Extended   & $+0.336$ & $+0.383$ \\
  & Guidelines & $+0.342$ & $+0.397$ \\
\midrule
\multirow{2}{*}{Gemma}
  & Extended   & $+0.287$ & $+0.114$ \\
  & Guidelines & $+0.322$ & $+0.366$ \\
\midrule
\multirow{2}{*}{GPT-5}
  & Extended   & $+0.277$ & $+0.394$ \\
  & Guidelines & $+0.038$ & $+0.152$ \\
\midrule
\multirow{3}{*}{OpenReviewer}
  & Extended        & $+0.593$ & $+0.603$ \\
  & Guidelines      & $+0.526$ & $+0.597$ \\
  & Original        & $+0.544$ & $+0.679$ \\
\midrule
\multirow{2}{*}{Qwen}
  & Extended   & $+0.504$ & $+0.858$ \\
  & Guidelines & $+1.123$ & $+1.152$ \\
\bottomrule
\end{tabular}
\end{table}

\subsection{Confidence Score Distributions Across Prompts}
\label{sec:appendix:conf_dist_prompt}

Table~\ref{tab:conf_dist} presents the normalized confidence-score distribution
for each model and prompt, pooled across both conferences. Cells encode
\textit{extended} / \textit{guidelines} (and additionally /
\textit{original} for OpenReviewer). The 1-5 ordinal scale collapses almost
entirely onto two values-4 and 5-for most models, indicating that LLMs
rarely express low or moderate confidence.

\paragraph{Mass concentration at score 4.}
For DeepSeek the distribution is essentially a point mass at 4 under both
prompts ($0.990$ / $0.998$). Gemma and GPT-5 also concentrate at 4, though
with non-trivial probability at 3: Gemma assigns $0.168$ / $0.031$ to score 3,
and GPT-5 assigns $0.033$ / $0.272$. This reveals a striking prompt effect
for GPT-5: the guidelines prompt shifts nearly $27\%$ of mass from score 4
down to score 3, yielding a substantially less confident and arguably better
calibrated distribution, consistent with its low mean difference under
guidelines on ICLR ($+0.04$).

\paragraph{Bimodal behaviour in Qwen and OpenReviewer.}
Qwen is the only model to place substantial probability at score 5 under the
guidelines prompt ($0.774$), while the extended prompt concentrates at score 4
($0.614$). This dramatic reversal ($\Delta \approx 0.55$ in mass between
scores 4 and 5) explains Qwen's large mean-difference gap between prompts:
the guidelines prompt almost deterministically pushes Qwen to maximum
confidence. OpenReviewer shows almost a bimodal distribution
($\approx 0.77$ at 4 and $\approx 0.23$ at 5), which is stable across all
three prompt conditions and is closest to the shape one might expect from a
involved reviewer.

\paragraph{Absence of low confidence scores.}
Scores 1 and 2 are virtually absent for all models ($\leq 0.002$), whereas
human reviewers use the full 1–5 scale. This truncation of the lower
confidence range is a qualitative difference between LLM and human behavior
that cannot be captured by mean-difference statistics alone.

\begin{table}[H]
\centering
\caption{Normalized confidence-score distribution per model and prompt
(extended / guidelines [/ original for OpenReviewer]) along with Human (OpenReview) distribution.
Values sum to 1.00 within each model--prompt combination.}
\label{tab:conf_dist}
\small
\setlength{\tabcolsep}{4pt}
\begin{tabular}{c c ccccc}
\toprule
\textbf{Score}
  & \textbf{Human}
  & \textbf{DeepSeek}
  & \textbf{Gemma}
  & \textbf{GPT-5}
  & \textbf{OpenReviewer}
  & \textbf{Qwen} \\
\midrule
1   & $0.008$ &  $0.001$ / $0.000$ & $0.000$ / $0.000$ & $0.000$ / $0.000$ & $0.000$ / $0.000$ / $0.000$ & $0.001$ / $0.001$ \\
2   & $0.080$ & $0.000$ / $0.000$ & $0.000$ / $0.000$ & $0.000$ / $0.000$ & $0.000$ / $0.002$ / $0.001$ & $0.000$ / $0.000$ \\
3   & $0.332$ & $0.009$ / $0.001$ & $0.168$ / $0.031$ & $0.033$ / $0.272$ & $0.001$ / $0.000$ / $0.001$ & $0.035$ / $0.002$ \\
4   & $0.460$ & $0.990$ / $0.998$ & $0.832$ / $0.962$ & $0.967$ / $0.728$ & $0.770$ / $0.805$ / $0.755$ & $0.614$ / $0.222$ \\
5   & $0.121$ & $0.000$ / $0.001$ & $0.000$ / $0.006$ & $0.000$ / $0.000$ & $0.229$ / $0.192$ / $0.243$ & $0.349$ / $0.774$ \\
\bottomrule
\end{tabular}
\end{table}

\paragraph{Summary.}
Confidence calibration is the metric most strongly affected by prompt choice,
but the direction and magnitude of the effect are highly model-specific.
GPT-5 and Qwen show the largest prompt-induced distributional shifts; DeepSeek
and OpenReviewer are the most stable. Universally, LLMs avoid low confidence
values, producing distributions that are right-skewed relative to human
reviewers and potentially misleading as signals of reviewer certainty (see Figure \ref{fig:confidence_distribution_plots}).

\section{Rating Score Analysis}
\label{appendix:rating_dist_prompt}

\subsection{Rating Distributions Across Prompts}
\label{sec:appendix:rating_dist_prompt}

Table~\ref{tab:rating_dist} reports the normalized rating distribution
(1--10 scale) for each model and prompt type, pooled across both conferences.
The data reveal significant, model-specific shifts in rating behavior when the
prompt changes, with no single direction of effect shared by all models.

\paragraph{DeepSeek: collapse onto score 8 under guidelines.}
Under the extended prompt, DeepSeek produces almost a bimodal  distribution with a
dominant peak at 8 ($0.586$) and a secondary cluster at 2 ($0.146$), suggesting
inconsistent rating behavior. Under the guidelines prompt this bimodality
almost entirely disappears: $93.1\%$ of ratings concentrate at score 8, with
negligible probability elsewhere ($\leq 0.058$ at score 7, zero at scores 1-6).
This is the most extreme prompt-induced collapse observed in the benchmark and
implies that the guidelines prompt eliminates DeepSeek's capacity
to discriminate between papers.

\paragraph{Gemma: shift from 6-7 to 7-8 under guidelines.}
Under the extended prompt Gemma distributes mass relatively broadly across
scores 6 ($0.367$), 7 ($0.332$), and 8 ($0.191$). The guidelines prompt
shifts this distribution upward: score 8 absorbs $0.568$ of the mass, score 7
remains dominant ($0.358$), and score 6 drops sharply to $0.050$. Gemma
thus becomes less rigorous under the guidelines prompt, concentrating nearly
$93\%$ of ratings in bins 7-8.

\paragraph{GPT-5: largest downward shift under guidelines.}
GPT-5 shows the opposite pattern to Gemma and DeepSeek. Under the extended
prompt its modal score is 6 ($0.316$), with meaningful coverage of scores
5 ($0.196$), 7 ($0.151$), and 3-4 ($0.221$ combined). Under the guidelines
prompt the distribution shifts markedly toward 7 ($0.337$), reducing coverage
of scores 3-5 and compressing variance. Like other models, the guidelines
prompt makes GPT-5 more lenient on average, though less extremely so than
DeepSeek or Gemma.

\paragraph{OpenReviewer: stable across all three prompts.}
OpenReviewer is the most stable model, maintaining a consistent modal peak at
score 6 ($0.267$--$0.315$) and score 5 ($0.227$--$0.241$) across all three
prompt conditions. The original prompt places slightly more mass in the lower
tail (scores 1--3: $0.225$ combined) compared to the extended and guidelines
prompts, but overall distributional differences are small. This stability
reinforces the pattern observed for confidence scores: domain-specific
fine-tuning largely overrides prompt-level instructions.

\paragraph{Qwen: noticeable bimodality under guidelines.}
Qwen produces perhaps the most striking prompt effect in the entire analysis.
Under the extended prompt its distribution is broad, with peaks at 8 ($0.436$),
7 ($0.193$), and 6 ($0.176$). Under the guidelines prompt $51.5\%$ of mass
moves to score 8 and $36.4\%$ to score 9---a near-bimodal distribution
concentrated at the top of the scale. Only $2.2\%$ of ratings fall at score 6
or below. This behavior mirrors the confidence collapse seen for Qwen under
the guidelines prompt: concise instructions systematically push the model
toward the upper extreme of any ordinal scale it is asked to use.

\begin{table}[H]
\centering
\caption{Normalised rating distributions per model and prompt type
(extended / guidelines [/ original for OpenReviewer]).
Values sum to 1.00 within each column.}
\label{tab:rating_dist}
\small
\setlength{\tabcolsep}{3.8pt}
\begin{tabular}{c ccccc}
\toprule
\textbf{Score}
  & \textbf{DeepSeek}
  & \textbf{Gemma}
  & \textbf{GPT-5}
  & \textbf{OpenReviewer}
  & \textbf{Qwen} \\
\midrule
1  & $0.003$ / $0.000$ & $0.000$ / $0.000$ & $0.002$ / $0.002$ & $0.009$ / $0.012$ / $0.024$ & $0.002$ / $0.001$ \\
2  & $0.146$ / $0.000$ & $0.000$ / $0.000$ & $0.025$ / $0.010$ & $0.010$ / $0.005$ / $0.034$ & $0.005$ / $0.002$ \\
3  & $0.008$ / $0.000$ & $0.002$ / $0.000$ & $0.100$ / $0.024$ & $0.131$ / $0.096$ / $0.167$ & $0.033$ / $0.002$ \\
4  & $0.013$ / $0.000$ & $0.092$ / $0.004$ & $0.121$ / $0.106$ & $0.078$ / $0.073$ / $0.104$ & $0.046$ / $0.010$ \\
5  & $0.066$ / $0.000$ & $0.015$ / $0.007$ & $0.196$ / $0.134$ & $0.227$ / $0.241$ / $0.227$ & $0.087$ / $0.004$ \\
6  & $0.056$ / $0.001$ & $0.367$ / $0.050$ & $0.316$ / $0.269$ & $0.300$ / $0.315$ / $0.267$ & $0.176$ / $0.022$ \\
7  & $0.122$ / $0.058$ & $0.332$ / $0.358$ & $0.151$ / $0.337$ & $0.118$ / $0.097$ / $0.067$ & $0.193$ / $0.079$ \\
8  & $0.586$ / $0.931$ & $0.191$ / $0.568$ & $0.087$ / $0.112$ & $0.124$ / $0.153$ / $0.104$ & $0.436$ / $0.515$ \\
9  & $0.001$ / $0.010$ & $0.000$ / $0.013$ & $0.001$ / $0.004$ & $0.001$ / $0.004$ / $0.002$ & $0.015$ / $0.364$ \\
10 & $0.000$ / $0.000$ & $0.000$ / $0.000$ & $0.000$ / $0.001$ & $0.001$ / $0.003$ / $0.004$ & $0.006$ / $0.001$ \\
\bottomrule
\end{tabular}
\end{table}

\begin{figure}[H]
\centering
\begin{subfigure}{0.95\textwidth}
  \centering
  \includegraphics[width=1.0\linewidth]{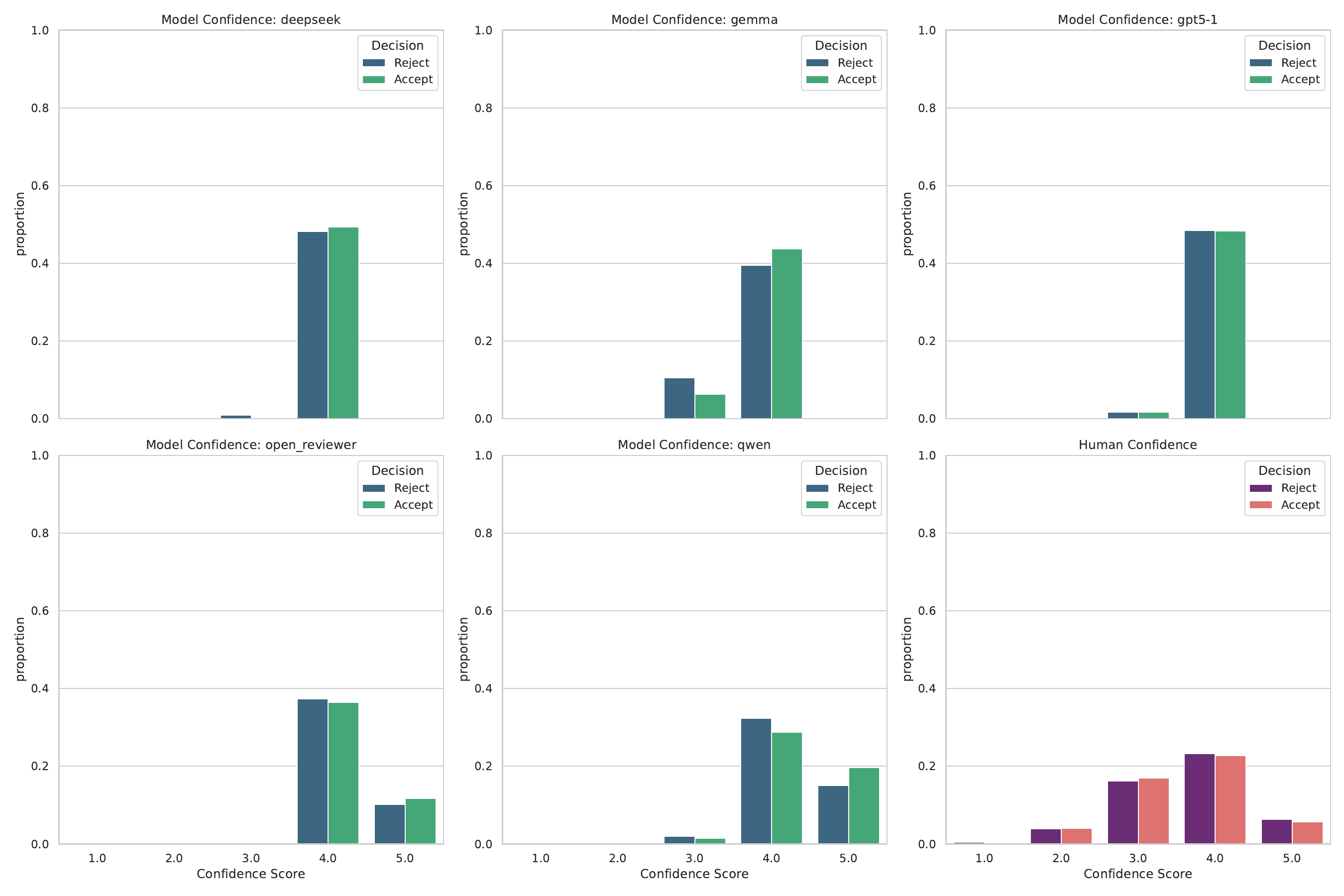}
  \caption{\textbf{Confidence distribution}. The bars are split into \textit{Accept} and \textit{Reject} categories (based on the final paper decision).}
  \label{fig:confidence_distribution_plots}
\end{subfigure}
\hfill
\begin{subfigure}{0.95\textwidth}
  \centering
  \includegraphics[width=1.0\linewidth]{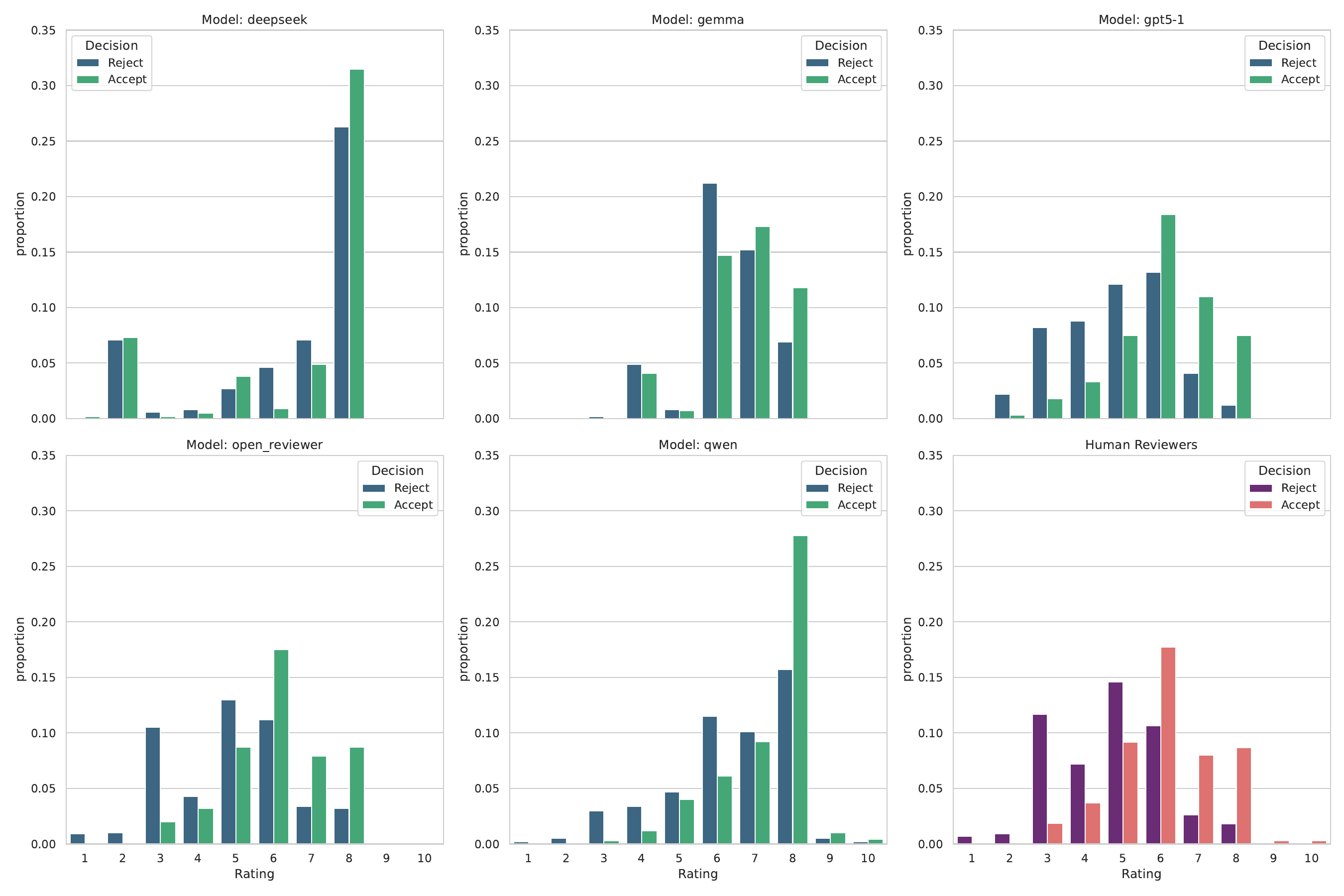}
  \caption{\textbf{Ratings distribution}. The bars are split into \textit{Accept} and \textit{Reject} categories (based on the final paper decision)}
  \label{fig:rating_bias_plots}
\end{subfigure}
\caption{Distributions of confidence and ratings reveal the substantial positive bias and differing mass assigned to papers that were either \textit{Accepted} or \textit{Rejected}.}
\label{fig:rating_confidence_plots}
\end{figure}

\subsection{Rating Distributions vs.\ Human Reviewers}
\label{sec:appendix:rating_diff_human}

Table~\ref{tab:rating_diff_human} reports the percentage-point difference
between each LLM's rating distribution under the extended prompt and the human
(OpenReview) baseline. The human distribution is concentrated in the mid-range:
scores 6 ($28.4\%$), 5 ($23.8\%$), and 3 ($13.5\%$) together account for
nearly two-thirds of all human ratings, with a secondary cluster at scores 4
($10.9\%$), 7 ($10.6\%$), and 8 ($10.5\%$). Scores at the extremes (1--2 and
9--10) are rare, together comprising under $3\%$ of human ratings (see Figure \ref{fig:rating_bias_plots}).

\paragraph{Shared pattern: hollowing out of the mid-range.}
All five models substantially under-use the human-dominant scores 3--6.
The largest shared deficits appear at score 5 ($-1.1$ to $-22.2$ pp) and
score 6 ($-10.8$ to $-22.8$ pp, except Gemma and OpenReviewer which modestly
over-use it). Score 3 is also universally avoided ($-0.5$ to $-13.3$ pp).
In place of this mid-range mass, all models shift probability upward into
scores 7--8, producing a systematic positive bias relative to human reviewers.

\paragraph{DeepSeek: extreme bimodal distortion.}
DeepSeek shows the most severe deviation from the human baseline. Its defining
feature is a $+48.1$ pp excess at score 8, which it assigns nearly five times
more often than human reviewers ($58.6\%$ vs.\ $10.5\%$). Simultaneously it
produces a large spike at score 2 ($+13.5$ pp) that has no parallel in any
other model, while deeply under-using scores 3--6 ($-9.6$ to $-22.8$ pp).
The result is a bimodal distribution concentrated at the extremes of 2 and 8,
which bears no resemblance to human rating behaviour and would render rankings
based on DeepSeek scores almost entirely uninformative.

\paragraph{Gemma: concentrated shift toward 7--8.}
Gemma eliminates nearly all probability from scores 2--5 ($-1.1$ to $-22.2$
pp) and reassigns it almost entirely to scores 7 ($+22.6$ pp) and 8 ($+8.7$
pp), with a modest over-use of score 6 ($+8.3$ pp). Unlike DeepSeek, Gemma
does not produce a lower-tail spike, so its distribution is unimodal but
strongly right-shifted. The total redistribution is large but more
interpretable than DeepSeek's bimodal pattern.

\paragraph{GPT-5: best-calibrated model, moderate upper shift.}
GPT-5 exhibits the smallest absolute deviations across the scale. It modestly
over-uses scores 6 ($+3.2$ pp), 7 ($+4.5$ pp), and 2 ($+1.5$ pp), while
under-using scores 3 ($-3.5$ pp) and 5 ($-4.2$ pp). The deviations at scores
7 and 8 are far smaller than for any other model ($+4.5$ and $-1.8$ pp
respectively), confirming that GPT-5 under the extended prompt is the closest
approximation to the human rating distribution. Its total variation
($\sum|\Delta| \approx 21$ pp) is roughly one-fifth that of DeepSeek.

\paragraph{OpenReviewer: closest to human overall.}
OpenReviewer shows the smallest deviations of all five models at nearly every
score. Its largest excesses are modest: $+1.97$ pp at score 8 and $+1.59$ pp
at score 6, and a small $+0.25$ pp at score 1 -- the only model to
\emph{over}-produce the lowest rating. It under-uses scores 3--5 ($-0.5$ to
$-3.1$ pp), but these deficits are an order of magnitude smaller than those
of DeepSeek or Gemma. OpenReviewer is thus the best-calibrated model in terms
of overall distributional fidelity to human reviewers, consistent with its
fine-tuning on real review data.

\paragraph{Qwen: strong upper-tail bias, second only to DeepSeek.}
Qwen places $+33.1$ pp excess mass at score 8 and $+8.8$ pp at score 7,
while strongly under-using scores 3 ($-10.2$ pp), 5 ($-15.1$ pp), and 6
($-10.8$ pp). Unlike DeepSeek it does not produce a lower-tail spike, but its
right-shift is nearly as extreme. Under the guidelines prompt (Table
\ref{tab:rating_dist}), Qwen pushes even further toward scores 8--9, making
it the model most sensitive to prompt choice in the upper tail.

\paragraph{Practical implications.}
The total variation distance from the human baseline -- approximated as
$\sum_r |\Delta(r)|$ -- ranks the models as: DeepSeek (${\approx}127$ pp)
$>$ Qwen (${\approx}87$ pp) $>$ Gemma (${\approx}79$ pp) $>$ GPT-5 (${\approx}21$ pp) $>$ OpenReviewer
(${\approx}10$ pp). The two-order-of-magnitude
gap between OpenReviewer and DeepSeek underscores that distributional alignment
with human reviewers is driven primarily by domain-specific training rather than
by model scale or general capability. Post-hoc score normalization or
distribution matching would be necessary before ratings from DeepSeek, Gemma,
or Qwen could substitute meaningfully for human assessments.

\begin{table}[H]
\caption{Rating distribution differences: LLM (extended prompt) minus human
baseline (percentage points). Positive values indicate the LLM over-produces
that rating relative to human reviewers; negative values indicate
under-production. The Human column gives the baseline frequency (\%).}
\label{tab:rating_diff_human}
\centering
\small
\begin{tabular}{c r rrrrr}
\toprule
\textbf{Score} & \textbf{Human (\%)}
  & \textbf{DeepSeek} & \textbf{Gemma} & \textbf{GPT-5}
  & \textbf{OpenReviewer} & \textbf{Qwen} \\
\midrule
1  & $0.69$  & $-0.39$ & $-0.69$ & $-0.49$ & $+0.25$ & $-0.49$ \\
2  & $1.05$  & $\mathbf{+13.54}$ & $-1.05$ & $+1.45$ & $\phantom{+}0.00$ & $-0.55$ \\
3  & $13.53$ & $-12.72$ & $-13.33$ & $-3.53$ & $-0.47$ & $-10.23$ \\
4  & $10.90$ & $-9.58$ & $-1.70$ & $+1.20$ & $-3.06$ & $-6.29$ \\
5  & $23.77$ & $-17.18$ & $-22.23$ & $-4.17$ & $-1.09$ & $-15.05$ \\
6  & $28.40$ & $-22.82$ & $+8.31$ & $+3.20$ & $+1.59$ & $-10.76$ \\
7  & $10.59$ & $+1.57$ & $+22.64$ & $+4.51$ & $+1.22$ & $+8.75$ \\
8  & $10.46$ & $\mathbf{+48.10}$ & $+8.66$ & $-1.76$ & $+1.97$ & $\mathbf{+33.12}$ \\
9  & $0.33$  & $-0.23$ & $-0.33$ & $-0.23$ & $-0.23$ & $+1.17$ \\
10 & $0.28$  & $-0.28$ & $-0.28$ & $-0.28$ & $-0.18$ & $+0.32$ \\
\midrule
$\sum|\Delta|$ & ---
  & $126.4$ & $79.2$ & $21.3$ & $10.3$ & $86.6$ \\
\bottomrule
\multicolumn{7}{l}{\footnotesize Bold entries mark deviations $>$13\,pp.
$\sum|\Delta|$: total variation distance from human baseline (pp).}
\end{tabular}
\end{table}

\section{Proxy metrics analysis}

\subsection{Validation of Cross-Reference Count as a Proxy for Grounding Specificity}
\label{sec:appendix:xrefs_validation}

The cross-reference count (\texttt{xrefs\_count}) is used throughout the
benchmark as a lightweight proxy for how specifically a review engages with
manuscript content. To validate this choice, we apply the RevUtil analytical
model~\cite{sadallah2025good} to individual (atomic) weakness comments extracted from both
human (OpenReview, 2025) and LLM-generated (2025, extended prompt) reviews,
scoring each comment on four dimensions: \emph{grounding specificity},
\emph{actionability}, \emph{helpfulness}, and \emph{verifiability}.
We then examine whether \texttt{xrefs\_count} -- computed purely from
surface-level pattern matching -- predicts the RevUtil grounding specificity
score.

\paragraph{Experimental setup.}
Weakness sections are split into atomic comments using a rule-based parser
handling fourteen list styles observed in OpenReview (numbered, bulleted,
dash-separated, and free-prose formats); comments shorter than 50 characters
are discarded. Each comment is scored independently by RevUtil and
per-review scores are obtained by averaging over all atomic comments.
Cross-reference counts are binned into four categories: $0$, $1$, $2$--$5$,
and ${>}5$ references. The choice is dictated by two reasons: (a) to avoid the influence of long-tailed distribution and (b) to reflect the categories of engagement with the paper content. We report the mean grounding specificity gain
relative to the zero-xrefs baseline (bin~$0$) for each model.

\paragraph{Results.}
Tables~\ref{tab:xrefs_bin_gen} and~\ref{tab:xrefs_bin_or} present the
bin-level results. For \textbf{human reviews}, the relationship is
strongly monotone: grounding specificity rises by $+0.319$ at bin~$1$,
$+0.526$ at bin~$2$--$5$, and $+0.833$ at bin~${>}5$ relative to reviews
with no cross-references. This large and consistent gradient confirms that,
in human-written text, cross-referencing manuscript elements is a reliable
surface signal of accurately grounded commentary.

For \textbf{LLM-generated reviews} the picture is model-dependent but
broadly consistent. OpenReviewer closely tracks the human pattern
($+0.763$, $+0.683$, $+0.716$ across bins~$1$, $2$--$5$, ${>}5$),
and Qwen likewise shows a strong monotone increase ($+0.377$, $+0.507$,
$+0.776$). GPT-5 exhibits a clear positive trend ($+0.033$, $+0.127$,
$+0.301$) albeit with smaller absolute gains. DeepSeek is the only model
with a negative value at bin~$1$ ($-0.019$), followed by a recovery at
higher bins ($+0.084$, $+0.139$), suggesting that single isolated
cross-references do not reliably indicate grounded commentary for this
model, possibly reflecting formulaic citation of paper elements without
substantive engagement. Gemma shows a non-monotone pattern ($+0.181$ at
bin~$1$ but only $+0.058$ at bin~$2$--$5$, with no observations at
bin~${>}5$), likely a consequence of sparse coverage in the higher bins.

\paragraph{Spearman correlations.}
Table~\ref{tab:spearman_corr} reports Spearman rank correlations between
the four RevUtil dimensions and the available surface features for both
human and LLM-generated reviews.
For \texttt{xrefs\_count}, the correlation with grounding specificity is
the strongest among all feature–dimension pairs in both datasets:
$\rho = 0.393$ for generated reviews and $\rho = 0.298$ for human reviews.
The density-normalized variant (\texttt{xrefs\_count\_density}) yields
similarly elevated values ($\rho = 0.334$ and $\rho = 0.270$ respectively),
confirming that the relationship is not driven purely by review length.
Crucially, the correlations between \texttt{xrefs\_count} and the other
three RevUtil dimensions are substantially weaker in both datasets:
actionability ($\rho = 0.047$ / $0.143$), helpfulness ($\rho = 0.138$ /
$0.057$), and verifiability ($\rho = 0.202$ / $0.155$). This dissociation
confirms that \texttt{xrefs\_count} tracks grounding specificity
selectively rather than acting as a generic quality signal.

\begin{table}[H]
\centering
\caption{Spearman rank correlations between RevUtil quality dimensions
and surface features, for LLM-generated (Gen) and human original (OR)
reviews (2025). Only \texttt{xrefs\_count} and
\texttt{xrefs\_count\_density} are shown.}
\label{tab:spearman_corr}
\small
\begin{tabular}{l cc cc}
\toprule
& \multicolumn{2}{c}{\textbf{xrefs\_count}}
& \multicolumn{2}{c}{\textbf{xrefs\_count\_density}} \\
\cmidrule(lr){2-3}\cmidrule(lr){4-5}
\textbf{Dimension} & Gen & OR & Gen & OR \\
\midrule
Actionability        & $0.047$          & $0.143$          & $0.002$          & $0.106$ \\
Grounding specificity& $\mathbf{0.393}$ & $\mathbf{0.298}$ & $\mathbf{0.334}$ & $\mathbf{0.270}$ \\
Helpfulness          & $0.138$          & $0.057$          & $0.082$          & $0.019$ \\
Verifiability        & $0.202$          & $0.155$          & $0.141$          & $0.097$ \\
\bottomrule
\end{tabular}
\end{table}

\paragraph{Human vs.\ LLM grounding specificity.}
Two observations stand out when comparing the two rows of
Table~\ref{tab:xrefs_bin_or}. First, LLMs score \emph{higher} than human
reviewers in absolute grounding specificity at every xrefs bin: the
aggregated LLM baseline (bin~$0$, $2.375$) already exceeds the human
baseline ($1.941$) by $0.434$ points, and this gap is maintained
across all bins ($2.639$ vs.\ $2.260$ at bin~$1$; $2.926$ vs.\ $2.467$
at bin~$2$--$5$; $3.309$ vs.\ $2.774$ at bin~${>}5$). Taken in isolation
this might suggest that LLMs produce more grounded reviews than humans,
but it should be interpreted cautiously: LLM-generated text is often
more structurally regular and uses explicit referencing language that
RevUtil's scoring model may reward, regardless of whether the underlying
engagement with the paper is substantive.

Second, and more importantly, the \emph{sensitivity} of grounding
specificity to cross-reference count is comparable between humans and
LLMs. The $\Delta$ values across bins are strikingly similar:
$+0.319$ vs.\ $+0.264$ at bin~$1$, $+0.526$ vs.\ $+0.551$ at
bin~$2$--$5$, and $+0.833$ vs.\ $+0.934$ at bin~${>}5$. This near-parallel
increase confirms that the relationship between \texttt{xrefs\_count}
and grounding specificity is not a peculiarity of LLM-generated text
but a structural property of review language that holds across both
human and machine authors. It provides the core justification for using
\texttt{xrefs\_count} as a benchmark metric: while absolute grounding
scores are not directly comparable across sources, the \emph{marginal
value} of each additional cross-reference is consistent, making the
metric a reliable instrument for comparing engagement behaviour within
and across model families.

\begin{table}[H]
\centering
\caption{Mean grounding specificity gain relative to the zero-xrefs
baseline (bin~$0$) for \textbf{LLM-generated} reviews (2025, extended
prompt). Values are differences in mean RevUtil grounding specificity score
versus the same model's bin-$0$ mean. \texttt{NaN (-)}: no observations in that bin.}
\label{tab:xrefs_bin_gen}
\small
\begin{tabular}{l rrrr}
\toprule
\textbf{Model} & \textbf{Bin 0} & \textbf{Bin 1} & \textbf{Bin 2--5} & \textbf{Bin ${>}$5} \\
               & (baseline)     & ($\Delta$)     & ($\Delta$)        & ($\Delta$) \\
\midrule
DeepSeek     & $0.000$ & $-0.019$ & $+0.084$ & $+0.139$ \\
Gemma        & $0.000$ & $+0.181$ & $+0.058$ & --- \\
GPT-5        & $0.000$ & $+0.033$ & $+0.127$ & $+0.301$ \\
OpenReviewer & $0.000$ & $+0.763$ & $+0.683$ & $+0.716$ \\
Qwen         & $0.000$ & $+0.377$ & $+0.507$ & $+0.776$ \\
\bottomrule
\multicolumn{5}{l}{\footnotesize $\Delta$ = mean grounding specificity minus bin-$0$ mean for the same model.}\\
\multicolumn{5}{l}{\footnotesize ---: no reviews with ${>}5$ cross-references for this model.}
\end{tabular}
\end{table}

\begin{table}[H]
\centering
\caption{Mean RevUtil grounding specificity score by xrefs bin for
\textbf{human} reviews (OpenReview, 2025) and \textbf{aggregated across
all LLM models} (2025, extended prompt).}
\label{tab:xrefs_bin_or}
\small
\begin{tabular}{l rr rr rr rr}
\toprule
& \multicolumn{2}{c}{\textbf{Bin 0}}
& \multicolumn{2}{c}{\textbf{Bin 1}}
& \multicolumn{2}{c}{\textbf{Bin 2--5}}
& \multicolumn{2}{c}{\textbf{Bin ${>}$5}} \\
\cmidrule(lr){2-3}\cmidrule(lr){4-5}\cmidrule(lr){6-7}\cmidrule(lr){8-9}
\textbf{Source} & score & $\Delta$ & score & $\Delta$ & score & $\Delta$ & score & $\Delta$ \\
\midrule
Human (OpenReview)
  & $1.941$ & $0.000$
  & $2.260$ & $+0.319$
  & $2.467$ & $+0.526$
  & $2.774$ & $+0.833$ \\
All LLMs (aggregated)
  & $2.375$ & $0.000$
  & $2.639$ & $+0.264$
  & $2.926$ & $+0.551$
  & $3.309$ & $+0.934$ \\
\bottomrule
\multicolumn{9}{l}{\footnotesize $\Delta$ = mean score minus bin-$0$ mean within the same row.}
\end{tabular}
\end{table}

\section{Atomic Strengths and Weaknesses}
\label{sec:appendix:sw_analysis}

\noindent \textbf{Atomic Fact Extraction.} 
To enable a granular comparison between human-authored and model-generated content, we decompose unstructured reviews into discrete logical units via an automated extraction pipeline. We utilize the vLLM engine \cite{kwon2023efficient} with a 9B-parameter model to perform guided parsing. The extraction is constrained by a JSON schema that forces the model to categorize segments into "strengths" and "weaknesses" while maintaining $1:1$ verbatim fidelity. This process preserves all technical \LaTeX\ formatting and idiosyncratic author phrasing while atomizing complex paragraphs into a structured array of individual arguments. This structured format serves as the foundation for our subsequent informational coverage and semantic overlap metrics. Prompt for human reviews \ref{lst:atomic_extract_human} is more complex than for generated reviews \ref{lst:atomic_extract_generated} as humans usually don't follow the exact bulletpoint-like structure.

\noindent \textbf{Atomic Extraction Prompt (Generated):}
\begin{lstlisting}[
  mathescape=true, 
  caption={Prompt for extracting atomic strengths and weaknesses from generated reviews.}, 
  label={lst:atomic_extract_generated},
  captionpos=b
]
You are a precise scientific data extraction engine. Your task is to parse conference reviews and extract "Strengths" and "Weaknesses" verbatim.

Extract "Strengths" and "Weaknesses" from the provided review verbatim.

**Rules for Extraction:**
1. **Atomization:** Split each section into a JSON array of individual points, split it into separate entries based on logical arguments.
2. **Verbatim Fidelity (1:1):** Copy the text exactly. Do NOT paraphrase, do NOT fix typos, and do NOT remove any "filler" words.
3. **Formatting:** Preserve all LaTeX (e.g., $O(n^2)$) and Markdown formatting.
4. **Output:** Return ONLY a valid JSON object. No preamble or postscript.\n\nReview text:\n{context}
\end{lstlisting}

\noindent \textbf{Atomic Extraction Prompt (Human):}
\begin{lstlisting}[
  mathescape=true, 
  caption={Prompt for extracting atomic strengths and weaknesses from human reviews.}, 
  label={lst:atomic_extract_human},
  captionpos=b
]
You are a precise scientific data extraction engine. Your task is to parse conference reviews and extract "Strengths" and "Weaknesses" verbatim.

Extract atomic "Strengths" and "Weaknesses" from the `strengths_weaknesses` section of the provided review.

**Rules for Extraction:**

1. **Atomicity & Self-Containedness:** Each entry must express one complete, self-contained unit of information - meaning it includes both the claim AND its supporting reasoning, evidence, or qualification as stated by the reviewer. Apply the following tests before splitting:
   - Would the second part be meaningless or ambiguous without the first? $\rightarrow$ Keep together.
   - Does the second sentence begin with a connector like "however", "but", "this means", "which", "therefore", or "this"? $\rightarrow$Keep together.
   - Do two sentences share the same subject and form one logical argument? $\rightarrow$ Keep together.
   Only split when two observations are genuinely independent and each is fully self-contained on its own.

2. **Near-Verbatim:** Copy the reviewer's wording as closely as possible. You MAY:
   - Trim pure filler praise at the start (e.g. "While the method is elegant, [weakness]" $\rightarrow$ keep only the weakness clause).
   - Remove transitional openers that reference prior entries (e.g. "Additionally," "Furthermore,") when they add no meaning.
   You MUST NOT paraphrase, restructure, or correct typos/grammar.

3. **Formatting:** Preserve all LaTeX (e.g., $O(n^2)$) and Markdown exactly as written.

4. **Classification:** When a statement contains both a positive and a negative aspect joined by contrast (e.g. "X is good, but Y is lacking"), split on the contrast and assign each clause to its correct list. If a statement is genuinely ambiguous in polarity, place it in "weaknesses".

5. **Output:** Return ONLY a valid JSON object with exactly two keys: `"strengths"` and `"weaknesses"`, each containing an array of strings. No preamble, postscript, or markdown code fences.
\n\nReview text:\n{context}
\end{lstlisting}

\subsection{Human Baseline}
\label{sec:appendix:sw_human}

Table~\ref{tab:original_stats} summarizes the distribution of atomic
weakness and strength counts in human reviews ($n = 804$).
The mean weakness count ($5.58 \pm 4.40$) substantially exceeds the
mean strength count ($3.74 \pm 2.14$), and the wide interquartile
range for weaknesses (Q1$=3$, Q3$=7$) relative to strengths
(Q1$=2$, Q3$=5$) reflects greater variability in how granularly
reviewers articulate criticism.

\begin{table}[H]
\centering
\caption{Descriptive Statistics for Original Human Reviews (2025 subset, N=804)}
\label{tab:original_stats}
\footnotesize
\begin{tabular}{lrrrrrrr}
\toprule
Metric & Mean & Std & Min & Q1 & Median & Q3 & Max \\
\midrule
\multicolumn{8}{l}{\textit{Component Counts}} \\
Atomic Weaknesses & 5.58 & 4.40 & 0.00 & 3.00 & 4.00 & 7.00 & 40.00 \\
Atomic Strengths & 3.74 & 2.14 & 0.00 & 2.00 & 3.00 & 5.00 & 19.00 \\
Total Atomic Points & 9.32 & 5.36 & 0.00 & 6.00 & 8.00 & 12.00 & 43.00 \\
\midrule
\multicolumn{8}{l}{\textit{Text Length (Chars)}} \\
Weakness Descriptions & 959.23 & 741.55 & 0.00 & 423.25 & 772.50 & 1249.00 & 4652.00 \\
Strength Descriptions & 459.98 & 320.04 & 0.00 & 245.75 & 401.00 & 611.25 & 2369.00 \\
Total Review Length & 1419.21 & 845.07 & 0.00 & 817.75 & 1246.00 & 1810.50 & 5209.00 \\
\bottomrule
\end{tabular}
\end{table}

\paragraph{Decision and conference breakdown.}
Table~\ref{tab:sw_human_decision} shows that rejected papers receive
more weaknesses ($5.93$ vs.\ $5.24$) and fewer strengths ($3.45$
vs.\ $4.04$) than accepted ones, with correspondingly longer weakness
sections and shorter strength sections. ICLR reviews contain more
weaknesses than NeurIPS reviews ($6.08$ vs.\ $5.10$), while strength
counts are similar ($3.80$ vs.\ $3.68$).

\begin{table}[H]
\centering
\caption{Mean atomic comment counts and section lengths for human
reviews (2025) broken down by paper decision and conference.}
\label{tab:sw_human_decision}
\small
\begin{tabular}{ll rrrr}
\toprule
\textbf{Split} & \textbf{Group}
  & \textbf{W count} & \textbf{S count}
  & \textbf{Len W} & \textbf{Len S} \\
\midrule
\multirow{2}{*}{Decision}
  & Accept & $5.24$ & $4.04$ & $867$ & $499$ \\
  & Reject & $5.93$ & $3.45$ & $1{,}050$ & $421$ \\
\midrule
\multirow{2}{*}{Conference}
  & ICLR    & $6.08$ & $3.80$ & $1{,}004$ & $449$ \\
  & NeurIPS & $5.10$ & $3.68$ & $\phantom{1{,}}915$ & $471$ \\
\bottomrule
\multicolumn{6}{l}{\footnotesize W = weaknesses, S = strengths; lengths in characters.}
\end{tabular}
\end{table}

\subsubsection{LLM-Generated Reviews: Volume and Length}
\label{sec:appendix:sw_gen}

Table~\ref{tab:sw_gen_stats} reports mean weakness and strength counts
and section lengths for each LLM model, alongside the human baseline.

\begin{table}[H]
\centering
\caption{Mean ($\pm$ std) atomic comment counts and section lengths for LLM-generated reviews (extended prompt, 2025) and the human baseline.}
\label{tab:sw_gen_stats}
\small
\setlength{\tabcolsep}{4pt}
\begin{tabular}{l rr rr rr}
\toprule
& \multicolumn{2}{c}{\textbf{W count}}
& \multicolumn{2}{c}{\textbf{S count}}
& \multicolumn{2}{c}{\textbf{Total count}} \\
\cmidrule(lr){2-3}\cmidrule(lr){4-5}\cmidrule(lr){6-7}
\textbf{Source} & Mean & Std & Mean & Std & Mean & Std \\
\midrule
Human (OR)   & $5.58$ & $4.40$ & $3.74$ & $2.14$ & $9.32$  & $5.36$  \\
\midrule
DeepSeek     & $4.33$ & $1.14$ & $5.19$ & $1.10$ & $9.52$  & $2.03$  \\
Gemma        & $7.06$ & $3.24$ & $6.66$ & $2.18$ & $13.72$ & $5.16$  \\
GPT-5        & $10.22$& $3.06$ & $7.71$ & $2.31$ & $17.93$ & $3.86$  \\
OpenReviewer & $6.25$ & $7.04$ & $4.03$ & $4.63$ & $10.28$ & $10.34$ \\
Qwen         & $6.93$ & $3.30$ & $6.08$ & $1.87$ & $13.01$ & $4.93$  \\
\bottomrule
\end{tabular}
\vspace{4pt}

\begin{tabular}{l rr rr rr}
\toprule
& \multicolumn{2}{c}{\textbf{Len W (chars)}}
& \multicolumn{2}{c}{\textbf{Len S (chars)}}
& \multicolumn{2}{c}{\textbf{Total len (chars)}} \\
\cmidrule(lr){2-3}\cmidrule(lr){4-5}\cmidrule(lr){6-7}
\textbf{Source} & Mean & Std & Mean & Std & Mean & Std \\
\midrule
Human (OR)   & $959$    & $742$    & $460$    & $320$    & $1{,}419$ & $845$    \\
\midrule
DeepSeek     & $656$    & $155$    & $730$    & $137$    & $1{,}385$ & $270$    \\
Gemma        & $1{,}316$& $337$    & $1{,}063$& $257$    & $2{,}379$ & $550$    \\
GPT-5        & $2{,}604$& $624$    & $1{,}167$& $252$    & $3{,}771$ & $581$    \\
OpenReviewer & $1{,}353$& $1{,}174$& $631$    & $757$    & $1{,}984$ & $1{,}577$\\
Qwen         & $1{,}448$& $320$    & $1{,}100$& $232$    & $2{,}548$ & $485$    \\
\bottomrule
\end{tabular}
\end{table}

\paragraph{GPT-5: highest volume overall.}
GPT-5 produces the most atomic items of any model ($10.22$ weaknesses,
$7.71$ strengths, total $17.93$), nearly double the human total count
($9.32$), and by far the longest weakness sections ($2{,}604$ characters,
vs.\ human mean $959$). Its standard deviation is relatively low
compared to its mean, indicating this inflation is consistent rather
than driven by outliers.

\paragraph{DeepSeek: fewer weaknesses, more strengths than humans.}
DeepSeek is the only model with a mean weakness count \emph{below} the
human baseline ($4.33$ vs.\ $5.58$, $\Delta_\text{mean} = -1.25$) while simultaneously producing more
strengths ($5.19$ vs.\ $3.74$, $\Delta_\text{mean} = +1.45$). Its
weaknesses are also the shortest of all models and both shorter
than humans (weaknesses $656$ vs.\ $959$), though their strengths length is longer than human strength section ($730$ vs.\ $460$). The very low standard deviations
($\sigma_W = 1.14$, $\sigma_S = 1.10$) for comment points indicate highly uniform output.

\paragraph{OpenReviewer: closest mean counts, highest variance.}
OpenReviewer's mean weakness ($6.25$) and strength ($4.03$) counts are
closest to the human baseline, with differences of $+0.66$ and $+0.30$
respectively ($\Delta_\text{median} = 0$ for both). However, its standard
deviations are extremely large ($\sigma_W = 7.04$, $\sigma_S = 4.63$,
total $\sigma = 10.34$), far exceeding all other models and even the
human standard deviation ($\sigma_W = 4.40$). This indicates bimodal
output behaviour rather than consistent human-like generation.

\paragraph{Gemma and Qwen: moderate symmetric inflation.}
Gemma and Qwen both inflate weakness and strength counts by a similar
margin above the human baseline ($\Delta_\text{mean}$: Gemma $+1.47$W,
$+2.92$S; Qwen $+1.35$W, $+2.34$S), with moderate standard deviations.
Neither model shows strong decision-conditioned modulation
(Table~\ref{tab:sw_gen_decision}): the difference between accepted and
rejected papers is small for both (${\leq}0.29$ items in any category).

\subsubsection{Decision-Conditioned Counts for LLM Models}
\label{sec:appendix:sw_gen_decision}

\begin{table}[H]
\centering
\caption{Mean atomic comment counts and section lengths for LLM-generated
reviews (2025) by paper decision.}
\label{tab:sw_gen_decision}
\small
\setlength{\tabcolsep}{4pt}
\begin{tabular}{ll rrrr}
\toprule
\textbf{Model} & \textbf{Decision}
  & \textbf{W count} & \textbf{S count}
  & \textbf{Len W} & \textbf{Len S} \\
\midrule
\multirow{2}{*}{DeepSeek}
  & Accept & $4.30$ & $5.27$ & $654$    & $725$ \\
  & Reject & $4.36$ & $5.12$ & $657$    & $734$ \\
\midrule
\multirow{2}{*}{Gemma}
  & Accept & $6.91$ & $6.78$ & $1{,}307$& $1{,}093$ \\
  & Reject & $7.20$ & $6.54$ & $1{,}325$& $1{,}034$ \\
\midrule
\multirow{2}{*}{GPT-5}
  & Accept & $9.78$ & $8.55$ & $2{,}438$& $1{,}253$ \\
  & Reject & $10.64$& $6.89$ & $2{,}767$& $1{,}083$ \\
\midrule
\multirow{2}{*}{OpenReviewer}
  & Accept & $5.95$ & $4.23$ & $1{,}293$& $654$ \\
  & Reject & $6.54$ & $3.84$ & $1{,}412$& $609$ \\
\midrule
\multirow{2}{*}{Qwen}
  & Accept & $7.04$ & $6.32$ & $1{,}426$& $1{,}140$ \\
  & Reject & $6.83$ & $5.85$ & $1{,}470$& $1{,}061$ \\
\bottomrule
\multicolumn{6}{l}{\footnotesize Lengths in characters.}
\end{tabular}
\end{table}

Among all models, GPT-5 and OpenReviewer show the most human-like
decision modulation: both assign more weaknesses and fewer strengths to
rejected papers ($+0.86$W and $-1.66$S for GPT-5; $+0.59$W and $-0.39$S
for OpenReviewer), directionally consistent with the human pattern
($+0.69$W, $-0.59$S). DeepSeek shows virtually no modulation ($+0.06$W,
$-0.15$S), and Qwen reverses the human weakness direction (rejected papers
receive \emph{fewer} weaknesses: $-0.21$).

\subsubsection{Generated-to-Human Ratios and Differences}
\label{sec:appendix:sw_ratio}

\begin{table}[H]
\centering
\caption{Ratio of LLM-generated to human atomic comment counts per
matched paper (mean / median / std). Values ${>}1$ indicate the LLM
produces more items than the human reviewer for the same paper.}
\label{tab:sw_ratio}
\small
\begin{tabular}{l ccc ccc ccc}
\toprule
& \multicolumn{3}{c}{\textbf{Ratio W}}
& \multicolumn{3}{c}{\textbf{Ratio S}}
& \multicolumn{3}{c}{\textbf{Ratio Total}} \\
\cmidrule(lr){2-4}\cmidrule(lr){5-7}\cmidrule(lr){8-10}
\textbf{Model} & Mean & Med & Std & Mean & Med & Std & Mean & Med & Std \\
\midrule
DeepSeek     & $1.23$ & $1.00$ & $0.96$ & $1.73$ & $1.67$ & $1.02$ & $1.28$ & $1.12$ & $0.70$ \\
Gemma        & $2.02$ & $1.50$ & $1.89$ & $2.20$ & $2.00$ & $1.43$ & $1.83$ & $1.62$ & $1.18$ \\
GPT-5        & $2.90$ & $2.20$ & $2.43$ & $2.55$ & $2.25$ & $1.62$ & $2.37$ & $2.14$ & $1.24$ \\
OpenReviewer & $1.79$ & $1.00$ & $3.21$ & $1.33$ & $1.00$ & $1.75$ & $1.38$ & $1.00$ & $1.92$ \\
Qwen         & $2.06$ & $1.33$ & $2.12$ & $2.05$ & $1.67$ & $1.44$ & $1.75$ & $1.50$ & $1.15$ \\
\bottomrule
\end{tabular}
\end{table}

\begin{table}[H]
\centering
\caption{Signed difference in atomic comment counts (LLM $-$ human,
mean and median). Positive means LLM produces more items.}
\label{tab:sw_diff}
\small
\begin{tabular}{l rr rr}
\toprule
& \multicolumn{2}{c}{\textbf{Weaknesses}}
& \multicolumn{2}{c}{\textbf{Strengths}} \\
\cmidrule(lr){2-3}\cmidrule(lr){4-5}
\textbf{Model} & $\Delta$ Mean & $\Delta$ Median & $\Delta$ Mean & $\Delta$ Median \\
\midrule
DeepSeek     & $-1.25$ & $\phantom{+}0.00$ & $+1.45$ & $+2.00$ \\
Gemma        & $+1.47$ & $+2.00$           & $+2.92$ & $+3.00$ \\
GPT-5        & $+4.63$ & $+6.00$           & $+3.97$ & $+4.00$ \\
OpenReviewer & $+0.66$ & $\phantom{+}0.00$ & $+0.30$ & $\phantom{+}0.00$ \\
Qwen         & $+1.35$ & $+1.00$           & $+2.34$ & $+2.00$ \\
\bottomrule
\end{tabular}
\end{table}

OpenReviewer's median ratio of $1.00$ for both weaknesses and strengths
means the typical review matches the human count exactly, despite its
higher mean driven by outliers. GPT-5 has the largest weakness ratio
(mean $2.90$, median $2.20$) and the largest absolute differences
($+4.63$W, $+3.97$S), with a substantial gap between mean and median
indicating a right-skewed distribution. DeepSeek's weakness ratio
(mean $1.23$, median $1.00$) combined with a negative mean difference
($-1.25$) confirms that it produces fewer weaknesses than humans for
most papers.

\subsection{Weakness-to-Strength Ratio}
\label{sec:appendix:ws_ratio}

\begin{table}[H]
\centering
\caption{Per-review weakness-to-strength ratio (W/S, excluding reviews
with zero strengths). Values ${<}1$ indicate more strengths than
weaknesses.}
\label{tab:ws_ratio}
\small
\begin{tabular}{l rrrr}
\toprule
\textbf{Source} & \textbf{Mean} & \textbf{Median} & \textbf{Std} & \textbf{$n$} \\
\midrule
Human (OpenReview) & $1.76$ & $1.33$ & $1.73$ & $779$ \\
\midrule
DeepSeek           & $0.84$ & $0.80$ & $0.16$ & $804$ \\
Gemma              & $1.05$ & $1.00$ & $0.25$ & $773$ \\
GPT-5              & $1.44$ & $1.29$ & $0.64$ & $804$ \\
OpenReviewer       & $1.85$ & $1.33$ & $1.80$ & $792$ \\
Qwen               & $1.12$ & $1.00$ & $0.32$ & $792$ \\
\bottomrule
\end{tabular}
\end{table}

Human reviewers produce $1.76$ weaknesses per strength on average
(median $1.33$). OpenReviewer is the only model that closely matches
both the mean ($1.85$) and variance ($\sigma = 1.80$ vs.\ $1.73$) of
this ratio. GPT-5 ($1.44$) is the next closest, followed by Qwen
($1.12$) and Gemma ($1.05$), both near parity. DeepSeek ($0.84$) is
the only model below 1, meaning it consistently writes more strengths
than weaknesses -- the inverse of the human pattern. The low standard
deviations for DeepSeek ($0.16$), Gemma ($0.25$), and Qwen ($0.32$)
compared to humans ($1.73$) indicate that these models apply a nearly
fixed balance regardless of paper content, while OpenReviewer's high
variance mirrors the human distribution more faithfully.

\subsection{Intra-Review Atomic Overlap}
\label{sec:appendix:sw_overlap}

Table~\ref{tab:sw_overlap} presents intra-review overlap metrics for atomic items, specifically ROUGE-L and a containment score -- representing the fraction of items entirely subsumed by another within the same review—compared against a human baseline. 

DeepSeek demonstrates the highest weakness containment ($0.946$), suggesting its critiques are formulated as highly concise, self-contained units; this correlates with the earlier finding that its mean weakness length is shorter than the human average. GPT-5 achieves a perfect ROUGE-L score ($1.000$) for both weaknesses and strengths, indicating that its atomic items are perfectly literal extractions of the source text. Its strength containment ($0.993$) is the second-highest observed, potentially indicating a structured phrasing that is preserved during extraction.

Conversely, Qwen is the only model with a weakness containment score ($0.817$) below the human baseline ($0.877$), suggesting its strengths and weaknesses section may be quite different from the other models -- though still high. Gemma and OpenReviewer occupy an intermediate range, with metrics that consistently exceed the human baseline for strengths while remaining close to it for weaknesses, further confirming the reliability of the extraction process. These patterns are largely replicated in the strengths analysis, where containment scores are generally higher than for weaknesses, a trend potentially influenced by the smaller overall volume of strengths provided by the models.
\begin{table}[H]
\centering
\caption{Intra-review atomic overlap metrics for LLM-generated and human reviews. ROUGE-L: mean pairwise ROUGE-L between extracted atomic comments (higher indicates greater literal overlap with the source); Contained: fraction of atomic items subsumed by another within the same review (higher indicates greater redundant information).}
\label{tab:sw_overlap}
\small
\begin{tabular}{l cc c cc}
\toprule
& \multicolumn{2}{c}{\textbf{Weaknesses}} & & \multicolumn{2}{c}{\textbf{Strengths}} \\
\cmidrule(lr){2-3} \cmidrule(lr){5-6}
\textbf{Source} & ROUGE-L & Contained & & ROUGE-L & Contained \\
\midrule
Human (OR)    & 0.973 & 0.877 & & 0.968 & 0.917 \\
\midrule
DeepSeek      & 0.986 & 0.946 & & 0.998 & 0.989 \\
Gemma         & 0.961 & 0.929 & & 0.961 & 0.954 \\
GPT-5         & 1.000 & 0.916 & & 1.000 & 0.993 \\
OpenReviewer  & 0.986 & 0.912 & & 0.985 & 0.970 \\
Qwen          & 0.985 & 0.817 & & 0.985 & 0.920 \\
\bottomrule
\end{tabular}
\end{table}

DeepSeek shows the highest weakness containment ($0.946$) among all sources, indicating that its small set of atomic items is lexically repetitive and that many items overlap substantially.
GPT-5 has the lowest weakness Jaccard ($0.128$), below even the human baseline ($0.180$), suggesting that its larger item count comes with greater lexical diversity between items; however, its ROUGE-L of $1.000$ and high strength containment ($0.993$) indicate sentence-level redundancy despite word-level diversity.
OpenReviewer shows the highest strength Jaccard ($0.316$) and
Jaccard for weaknesses ($0.291$) of all sources, meaning its items are lexically the most similar to each other.
Qwen achieves the lowest containment for weaknesses ($0.817$), below
the human baseline ($0.877$) -- the only model to do so -- indicating the least redundancy at the item level among all sources.

\section{Coverage of Human Atomic Comments}
\label{sec:appendix:coverage}

Tables~\ref{tab:coverage_weakness} and~\ref{tab:coverage_strength}
report four coverage metrics computed between LLM-generated atomic
items and the matched human atomic items for weaknesses and strengths
respectively (2025 subset, extended prompt).
The metrics are:
\textbf{atomic recall (coverage)} -- the fraction of human items covered by at
least one generated item;
\textbf{generation precision} -- the fraction of generated items that
map to at least one human item;
\textbf{mapping density} -- the mean number of generated items per
human item (values ${>}1$ indicate one-to-many mapping);
\textbf{compaction ratio} -- the mean number of human items per
generated item (values ${>}1$ indicate many-to-one mapping, i.e.\
the generated item is denser).

\noindent \paragraph{Information Coverage Evaluation.} 
To quantify the preservation of substantive content, we implement an automated evaluation pipeline utilizing the vLLM inference engine \cite{kwon2023efficient}. The evaluation is framed as a zero-shot binary classification task where the model identifies informational overlap between atomic segments of human-authored and model-generated reviews. To ensure deterministic results and eliminate parsing overhead, we employ structured output decoding restricted to the boolean set $\{\text{yes, no}\}$ and set the decoding temperature to $\tau = 0.0$. The prompt \ref{lst:coverage_prompt} utilize a role-based architecture, positioning the model as a linguistic analyst tasked with identifying substantive information preservation across the dataset.

\noindent \textbf{Prompt:}
\begin{lstlisting}[
  mathescape=true, 
  caption={Prompt for evaluating informational coverage.}, 
  label={lst:coverage_prompt},
  captionpos=b
]
You are a precise linguistic data analyst evaluating information coverage.
"Your task is to determine if the Target text contains at least one substantive piece of information from the Base text.\n\nBase text: '{base_text}'\nTarget text: '{task_text}'\nIs the information present?
\end{lstlisting}

\begin{table}[ht]
\centering
\caption{Weakness coverage statistics per model and aggregated
(mean $\pm$ std). \emph{Recall}: fraction of human weaknesses covered;
\emph{Prec.}: fraction of generated weaknesses matching a human item;
\emph{Density}: generated-per-human mapping; \emph{Compact.}: human-per-generated mapping.}
\label{tab:coverage_weakness}
\small
\setlength{\tabcolsep}{4pt}
\begin{tabular}{l rr rr rr rr}
\toprule
& \multicolumn{2}{c}{\textbf{Recall}}
& \multicolumn{2}{c}{\textbf{Prec.}}
& \multicolumn{2}{c}{\textbf{Density}}
& \multicolumn{2}{c}{\textbf{Compact.}} \\
\cmidrule(lr){2-3}\cmidrule(lr){4-5}
\cmidrule(lr){6-7}\cmidrule(lr){8-9}
\textbf{Model} & $\mu$ & $\sigma$ & $\mu$ & $\sigma$
               & $\mu$ & $\sigma$ & $\mu$ & $\sigma$ \\
\midrule
DeepSeek     & $0.70$ & $0.30$ & $0.70$ & $0.30$
             & $1.70$ & $1.46$ & $0.90$ & $0.74$ \\
Gemma        & $0.77$ & $0.29$ & $0.68$ & $0.30$
             & $1.73$ & $1.56$ & $0.65$ & $0.59$ \\
GPT-5        & $\mathbf{0.93}$ & $0.20$ & $\mathbf{0.76}$ & $0.24$
             & $2.35$ & $2.12$ & $0.56$ & $0.48$ \\
OpenReviewer & $0.73$ & $0.33$ & $0.68$ & $0.33$
             & $2.00$ & $2.04$ & $1.01$ & $1.17$ \\
Qwen         & $0.84$ & $0.26$ & $0.73$ & $0.28$
             & $2.18$ & $2.02$ & $0.82$ & $0.77$ \\
\midrule
\textit{Mean} & $0.79$ & $0.28$ & $0.71$ & $0.29$
              & $1.99$ & $1.84$ & $0.79$ & $0.75$ \\
\textit{Std}  & $0.09$ & $0.05$ & $0.03$ & $0.03$
              & $0.28$ & $0.31$ & $0.18$ & $0.26$ \\
\bottomrule
\multicolumn{9}{l}{\footnotesize Bold: best per column.}
\end{tabular}
\end{table}

\begin{table}[ht]
\centering
\caption{Strength coverage statistics per model and aggregated
(mean $\pm$ std). Columns as in Table~\ref{tab:coverage_weakness}.}
\label{tab:coverage_strength}
\small
\setlength{\tabcolsep}{4pt}
\begin{tabular}{l rr rr rr rr}
\toprule
& \multicolumn{2}{c}{\textbf{Recall}}
& \multicolumn{2}{c}{\textbf{Prec.}}
& \multicolumn{2}{c}{\textbf{Density}}
& \multicolumn{2}{c}{\textbf{Compact.}} \\
\cmidrule(lr){2-3}\cmidrule(lr){4-5}
\cmidrule(lr){6-7}\cmidrule(lr){8-9}
\textbf{Model} & $\mu$ & $\sigma$ & $\mu$ & $\sigma$
               & $\mu$ & $\sigma$ & $\mu$ & $\sigma$ \\
\midrule
DeepSeek     & $0.92$ & $0.21$ & $0.81$ & $0.24$
             & $1.84$ & $1.11$ & $0.70$ & $0.43$ \\
Gemma        & $0.90$ & $0.27$ & $0.74$ & $0.27$
             & $1.64$ & $1.11$ & $0.53$ & $0.35$ \\
GPT-5        & $\mathbf{0.94}$ & $0.19$ & $\mathbf{0.85}$ & $0.23$
             & $2.09$ & $1.26$ & $0.50$ & $0.32$ \\
OpenReviewer & $0.78$ & $0.31$ & $0.77$ & $0.31$
             & $1.77$ & $1.24$ & $0.93$ & $0.70$ \\
Qwen         & $0.89$ & $0.24$ & $0.81$ & $0.26$
             & $1.81$ & $1.14$ & $0.59$ & $0.38$ \\
\midrule
\textit{Mean} & $0.89$ & $0.24$ & $0.80$ & $0.26$
              & $1.83$ & $1.17$ & $0.65$ & $0.44$ \\
\textit{Std}  & $0.06$ & $0.05$ & $0.04$ & $0.03$
              & $0.16$ & $0.07$ & $0.17$ & $0.15$ \\
\bottomrule
\end{tabular}
\end{table}

\paragraph{Strengths are better covered than weaknesses.}
The most consistent finding across both tables is that strength coverage
is uniformly higher than weakness coverage for every model and every
metric. Mean recall improves from $0.79$ to $0.89$ ($+0.10$), mean
precision from $0.71$ to $0.80$ ($+0.09$), and inter-model standard
deviation shrinks on both dimensions (recall $0.09 \to 0.06$; precision
$0.03 \to 0.04$). Mapping density is also lower for strengths
($1.83$ vs.\ $1.99$), indicating that generated strength items are
more precisely targeted rather than over-decomposed.
The likely explanation connects to the volume data in
Table~\ref{tab:sw_gen_stats}: all models generate more strengths than
humans (mean excess $+0.30$ to $+3.97$ items), and more items trivially
increases the chance of covering each human item, inflating recall.
For weaknesses, three of five models produce fewer or only modestly more
items than humans, so coverage must be earned through precise content
overlap rather than excessive volume.

\paragraph{GPT-5: highest recall, lowest compaction.}
GPT-5 achieves the highest recall for both weaknesses ($0.93$) and
strengths ($0.94$), and the highest precision on both dimensions
($0.76$ / $0.85$). Its mapping density is the highest of all models
for weaknesses ($2.35$), meaning it maps roughly $2.4$ generated items
to each human item on average. This is consistent with its large
generated weakness count ($10.22$ vs.\ human $5.58$): more items
increase recall but spread precision thin, and the low compaction ratio
($0.56$) confirms that GPT-5 decomposes human observations into finer
sub-points rather than consolidating them. The trade-off is favourable
for recall but means GPT-5 reviews are substantially more verbose than
necessary to cover the human content.

\paragraph{DeepSeek: strength recall disproportionate to volume.}
DeepSeek achieves the second-highest strength recall ($0.92$) despite
producing only $5.19$ strength items on average — just $1.45$ above
the human mean ($3.74$). This high recall-per-item efficiency stands
in contrast to its weakness recall ($0.70$, the lowest), where it
generates \emph{fewer} items than humans ($4.33$ vs.\ $5.58$). The
compaction ratio for weaknesses ($0.90$) is close to~$1$, suggesting
near one-to-one mapping, while for strengths ($0.70$) it is lower,
indicating slight consolidation. DeepSeek's inverted weakness-to-strength
ratio (W/S $= 0.84$, the only model below~$1$) is therefore reflected
directly in its coverage profile: it is a relatively complete strength
reviewer but a substantially incomplete weakness reviewer.

\paragraph{OpenReviewer: lowest strength recall, high compaction.}
OpenReviewer achieves the lowest strength recall ($0.78$) despite
producing the closest-to-human strength count ($4.03$ vs.\ $3.74$).
Its compaction ratio for strengths ($0.93$) is the highest of all models,
meaning that each generated strength item covers nearly one full human
item — a near one-to-one mapping with little redundancy. The shortfall
in recall therefore arises not from inefficiency but from actual gaps:
OpenReviewer simply does not identify all the positive contributions
that human reviewers mention, possibly because its fine-tuning
data precedes some of the conventions for framing strengths that
appear in the 2025 reviews. Its weakness coverage profile is more
typical ($0.73$ recall, $0.68$ precision), with compaction above~$1$
($1.01$), the only model whose generated weaknesses encompass human ones on average.

\paragraph{Gemma and Qwen: moderate and consistent coverage.}
Gemma and Qwen occupy the middle range on all metrics for both
categories. Qwen's higher weakness recall ($0.84$) relative to Gemma
($0.77$) is consistent with its larger generated weakness count
($7.06$ vs.\ $6.93$, similar volumes) but lower compaction
($0.82$ vs.\ $0.65$), suggesting slightly less dense mapping.
For strengths, both models achieve recall around $0.89$--$0.90$ with
compaction ratios below $0.60$, implying that each generated strength item covers only part of a human item and that the higher  recall is driven by volume rather than precision of coverage.

\clearpage
\onecolumn

\section{Math content detection} \label{sec:math_content_detection}

We employ Qwen3-8B for math binary math detection. For this, two independent experts (authors of this paper) manually annotated a subset of 30 papers to provide a reference for the employed model. The crafted prompt for the model is provided below. On this subset, the model attained 96,6\% accuracy. 

\noindent \textbf{Prompt:}
\begin{lstlisting}[mathescape=true]
# Criterion: "Review Engaging with the Mathematical Content of a Paper"

A review is considered to engage with the mathematical content of a paper if it satisfies at least one of the following conditions.

## 1. Direct Reference to an Equation

The reviewer:
- cites a numbered equation (e.g., *Eq. (2)*, *Eq. 17*),
- explicitly refers to a specific equation and analyzes it,
- suggests a correction to an equation (e.g., sign error, incorrect index, missing variable).

**Example:**
> "In Eq. (7), the minus sign seems incorrect."

---

## 2. Reference to a Formal Result

The reviewer:
- refers to a *Lemma*, *Proposition*, *Theorem*, *Assumption* or *Definition*
- refers to a specific line or multiple lines,
- questions the correctness of a proof,
- analyzes a logical step in a derivation.

**Example:**
> "Proposition 1 does not imply the stated result."

---

## 3. Analysis of the Formal Model Specification

The reviewer:
- reproduces the mathematical form of the model,
- compares two formal equations or architectures,
- identifies inconsistencies in mathematical notation.

**Example:**
> "The architecture H^l = W^l $\Phi$(S^l H^{l-1}) + b^l is not equivalent to the proposed formulation unless $\tilde{A}$ is diagonal."

---

## 4. Weaker Formal Reference (Optional Category)

The reviewer:
- refers to a specific symbol or parameter from the formalism (e.g., $\alpha$),
- notes that a variable is missing from an equation,
- discusses a mathematical component of the model without citing a numbered equation.

Such cases are treated as weaker but still valid references to the mathematical content.

---

# What Does Not Satisfy the Criterion

A review is **not** considered to engage with the mathematical content if it:

- mentions "theoretical analysis", "math", or "proofs" in broad terms without referencing any specific formal block (e.g., a specific Theorem, Lemma, or Assumption) or without pointing to any actual mathematical expressions,
- refers to general mathematical concepts or algorithms (e.g., KL divergence, Bayesian filtering, gradient descent) in plain text, without reproducing, analyzing, or critiquing the specific mathematical notation or unnumbered formulas (e.g., $L = \sum x_i$) used in the paper,
- evaluates experiments, architecture, or novelty using empirical numbers or dimensions (e.g., "accuracy improved by 10%", "a 5x5 matrix") without engaging with the underlying formal derivations or mathematical representations.
---

# Operational Definition

A review engages with the mathematical content of a paper if the reviewer explicitly refers to the paper's formal elements (equations, theorems, lemmas, notation, or proofs) and analyzes, critiques, or corrects them.
\end{lstlisting}

\section{Prompts} \label{sec:prompts}

\noindent \textbf{System Prompt:}
\begin{lstlisting}
You are a helpful assistant.
Your knowledge cutoff: {{ knowledge_cutoff }}
\end{lstlisting}

\noindent \textbf{Guidelines Prompt:}
\begin{lstlisting}
Prepare review of following paper:
{{ paper_text }}
-----
Prepare your review according to review guidelines:
A high-quality {{ conference }} {{ year }} review should be substantive, constructive, and timely, focusing on the work's technical correctness, novelty, and overall value to the community rather than just state-of-the-art performance.
It must clearly outline strengths and weaknesses, provide a well-reasoned recommendation with supporting arguments, and include specific questions to facilitate an active discussion phase.
-----
Assess paper on scale 1 (Strong reject) to 10 (Strong accept).
Assess confidence of your review on scale of 1 (Educated guess) to 5 (Absolutely certain).
Output only well structured paper review and your assessments without additional comments.
Use the following format to structure your review. 
**Summary:** 
**Strengths:** 
**Weaknesses:**
**Questions:** 
**Limitations:**
**Additional feedback:**
**Rating:**
**Confidence:** 
Review:
\end{lstlisting}

\noindent \textbf{Prompt from literature (ICLR 2025):}
\begin{lstlisting}
You are an expert reviewer for AI conferences. You follow best practices and review papers according to the reviewer guidelines.
Your knowledge cutoff: {{ knowledge_cutoff }}

Reviewer guidelines:
1. Read the paper: It's important to carefully read through the entire paper, and to look up any related work and citations that will help you comprehensively evaluate it. Be sure to give yourself sufficient time for this step.
2. While reading, consider the following:
    - Objective of the work: What is the goal of the paper? Is it to better address a known application or problem, draw attention to a new application or problem, or to introduce and/or explain a new theoretical finding? A combination of these? Different objectives will require different considerations as to potential value and impact.
    - Strong points: is the submission clear, technically correct, experimentally rigorous, reproducible, does it present novel findings (e.g. theoretically, algorithmically, etc.)?
    - Weak points: is it weak in any of the aspects listed in b.?
    - Be mindful of potential biases and try to be open-minded about the value and interest a paper can hold for the community, even if it may not be very interesting for you.
3. Answer four key questions for yourself, to make a recommendation to Accept or Reject:
    - What is the specific question and/or problem tackled by the paper?
    - Is the approach well motivated, including being well-placed in the literature?
    - Does the paper support the claims? This includes determining if results, whether theoretical or empirical, are correct and if they are scientifically rigorous.
    - What is the significance of the work? Does it contribute new knowledge and sufficient value to the community? Note, this does not necessarily require state-of-the-art results. Submissions bring value to the community when they convincingly demonstrate new, relevant, impactful knowledge (incl., empirical, theoretical, for practitioners, etc).
4. Write your review including the following information: 
    - Summarize what the paper claims to contribute. Be positive and constructive.
    - List strong and weak points of the paper. Be as comprehensive as possible.
    - Clearly state your initial recommendation (accept or reject) with one or two key reasons for this choice.
    - Provide supporting arguments for your recommendation.
    - Ask questions you would like answered by the authors to help you clarify your understanding of the paper and provide the additional evidence you need to be confident in your assessment.
    - Provide additional feedback with the aim to improve the paper. Make it clear that these points are here to help, and not necessarily part of your decision assessment.

Your write reviews in markdown format. Your reviews contain the following sections:

# Review

Briefly summarize the paper and its contributions. This is not the place to critique the paper; the authors should generally agree with a well-written summary.

## Review Summary
Summarize what the paper claims to contribute. Be rigorously objective, critical, and constructive. Address: what problem is tackled, and whether the approach is well-motivated and situated in the literature. This is not the place to critique the paper.

## Strengths
List strong points of the paper. Consider: clarity, technical correctness, experimental rigor, reproducibility, and novelty (theoretical, algorithmic, or empirical). Use bullet points.

## Weaknesses
List weak points across the same dimensions as strengths. Be comprehensive but fair and open-minded about the paper's value to the broader ICLR community. Use bullet points.

## Questions for Authors
Ask questions needed to clarify your understanding or increase confidence in your assessment. You may also include suggestions for improvement - make clear these are constructive and not necessarily part of your decision.

## Additional Feedback
Provide additional feedback with the aim to improve the paper. Make it clear that these points are here to help, and not necessarily part of your decision assessment.

## Numerical assessment
Given the description of the fields below, provide an appropriate numerical rating.

### Soundness:
Please assign the paper a numerical rating on the following scale to indicate the soundness of the technical claims, experimental and research methodology and on whether the central claims of the paper are adequately supported with evidence. Choose from the following:
- 4: excellent
- 3: good
- 2: fair
- 1: poor

### Presentation:
Please assign the paper a numerical rating on the following scale to indicate the quality of the presentation. This should take into account the writing style and clarity, as well as contextualization relative to prior work. Choose from the following:
- 4: excellent
- 3: good
- 2: fair
- 1: poor

### Contribution:
Please assign the paper a numerical rating on the following scale to indicate the quality of the overall contribution this paper makes to the research area being studied. Are the questions being asked important? Does the paper bring a significant originality of ideas and/or execution? Choose from the following:
- 4: excellent
- 3: good
- 2: fair
- 1: poor

## Final Recommendation
**Decision:**
Choose from one of the following categories. Only use the values listed above:
- 10: strong accept, should be highlighted at the conference
- 8: accept, good paper
- 6: marginally above the acceptance threshold
- 5: marginally below the acceptance threshold
- 3: reject, not good enough
- 1: strong reject

**Key Reason:**
One or two sentences justifying your decision.
**Confidence:**
Choose one of the following categories. Only use the values listed above:
- 5: You are absolutely certain about your assessment.
- 4: You are confident in your assessment, but not absolutely certain.
- 3: You are fairly confident in your assessment.
- 2: You are willing to defend your assessment, but it is quite likely that you did not understand the central parts of the submission or that you are unfamiliar with some pieces of related work.
- 1: You are unable to assess this paper and have alerted the ACs to seek an opinion from different reviewers.

**Core evaluative criterion:** Does this work contribute new, relevant, and impactful knowledge to the ICLR community? State-of-the-art results are not required - rigorous and convincing demonstration of value is sufficient.

Your response must only contain the review in markdown format with sections as defined above.
\end{lstlisting}

\noindent \textbf{Extended ICLR 2025 Prompt:}
\begin{lstlisting}
Prepare review of following paper:
{{ paper_text }}
-----
Read the attached paper carefully and provide a thorough, fair, and constructive review.
When reviewing a paper, structure your response using the following format:

## Review Summary
Summarize what the paper claims to contribute. Be rigorously objective, critical, and constructive. Address: what problem is tackled, and whether the approach is well-motivated and situated in the literature. This is not the place to critique the paper.

## Strengths
List strong points of the paper. Consider: clarity, technical correctness, experimental rigor, reproducibility, and novelty (theoretical, algorithmic, or empirical). Use bullet points.

## Weaknesses
List weak points across the same dimensions as strengths. Be comprehensive but fair and open-minded about the paper's value to the broader ICLR community. Use bullet points.

## Questions for Authors
Ask questions needed to clarify your understanding or increase confidence in your assessment. You may also include suggestions for improvement - make clear these are constructive and not necessarily part of your decision.

## Additional Feedback
Provide additional feedback with the aim to improve the paper. Make it clear that these points are here to help, and not necessarily part of your decision assessment.

## Numerical assessment
Given the description of the fields below, provide an appropriate numerical rating.

### Soundness:
Please assign the paper a numerical rating on the following scale to indicate the soundness of the technical claims, experimental and research methodology and on whether the central claims of the paper are adequately supported with evidence. Choose from the following:
- 4: excellent
- 3: good
- 2: fair
- 1: poor

### Presentation:
Please assign the paper a numerical rating on the following scale to indicate the quality of the presentation. This should take into account the writing style and clarity, as well as contextualization relative to prior work. Choose from the following:
- 4: excellent
- 3: good
- 2: fair
- 1: poor

### Contribution:
Please assign the paper a numerical rating on the following scale to indicate the quality of the overall contribution this paper makes to the research area being studied. Are the questions being asked important? Does the paper bring a significant originality of ideas and/or execution? Choose from the following:
- 4: excellent
- 3: good
- 2: fair
- 1: poor

## Final Recommendation
**Decision:**
Choose from one of the following categories. Only use the values listed above:
- 10: strong accept, should be highlighted at the conference
- 8: accept, good paper
- 6: marginally above the acceptance threshold
- 5: marginally below the acceptance threshold
- 3: reject, not good enough
- 1: strong reject

**Key Reason:**
One or two sentences justifying your decision.
**Confidence:**
Choose one of the following categories. Only use the values listed above:
- 5: You are absolutely certain about your assessment.
- 4: You are confident in your assessment, but not absolutely certain.
- 3: You are fairly confident in your assessment.
- 2: You are willing to defend your assessment, but it is quite likely that you did not understand the central parts of the submission or that you are unfamiliar with some pieces of related work.
- 1: You are unable to assess this paper and have alerted the ACs to seek an opinion from different reviewers.

**Core evaluative criterion:** Does this work contribute new, relevant, and impactful knowledge to the ICLR community? State-of-the-art results are not required - rigorous and convincing demonstration of value is sufficient.

Review:
\end{lstlisting}
\newpage
\noindent \textbf{Extended NeurIPS 2025 Prompt:}
\begin{lstlisting}
Prepare review of following paper:
{{ paper_text }}
-----
Read the attached paper carefully and provide a thorough, fair, and
constructive peer review. Base your evaluation solely on the paper's content.

## Summary
Briefly summarize the paper and its contributions in your own words after reading.
Do not paste the abstract. The authors should generally agree with a well-written summary.
This is not the place to critique the paper.

## Strengths
List strong points of the paper. Consider: clarity, technical correctness, experimental rigor, reproducibility, and novelty (theoretical, algorithmic, or empirical). Use bullet points.

## Weaknesses
List weak points across the same dimensions as strengths. Be comprehensive but fair and open-minded about the paper's value to the broader NeurIPS community. Use bullet points.

## Questions
List 3-5 key, actionable questions for the authors. Focus on points where author responses
could change your evaluation. Clearly state the criteria under which your score could increase or decrease.

## Limitations
Have the authors adequately addressed the limitations and potential negative societal impact of their work? If yes, write "Yes." If not, provide constructive suggestions. Note: authors should be rewarded, not penalized, for being upfront about limitations.

## Numerical Ratings
Given the description of the fields below, provide an appropriate numerical rating.

### Quality:
Based on what you discussed in "Strengths" and "Weaknesses", please assign the paper a numerical rating on the following scale to indicate the quality of the work. Choose from the following:
- 4: excellent
- 3: good
- 2: fair
- 1: poor

### Clarity: 
Based on what you discussed in "Strengths" and "Weaknesses", please assign the paper a numerical rating on the following scale to indicate the clarity of the paper. Choose from the following:
- 4: excellent
- 3: good
- 2: fair
- 1: poor

### Significance:
Based on what you discussed in "Strengths" and "Weaknesses", please assign the paper a numerical rating on the following scale to indicate the significance of the paper. Choose from the following:
- 4: excellent
- 3: good
- 2: fair
- 1: poor

### Originality:
Based on what you discussed in "Strengths" and "Weaknesses", please assign the paper a numerical rating on the following scale to indicate the originality of the paper. Choose from the following:
- 4: excellent
- 3: good
- 2: fair
- 1: poor

## Final Recommendation

**Overall Score:**
- 6: Strong Accept: Technically flawless paper with groundbreaking impact on one or more areas of AI, with exceptionally strong evaluation, reproducibility, and resources, and no unaddressed ethical considerations.
- 5: Accept: Technically solid paper, with high impact on at least one subarea of AI or moderate-to-high impact on more than one area of AI, with good to excellent evaluation, resources, reproducibility, and no unaddressed ethical considerations.
- 4: Borderline accept: Technically solid paper where reasons to accept outweigh reasons to reject, e.g., limited evaluation. Please use sparingly.
- 3: Borderline reject: Technically solid paper where reasons to reject, e.g., limited evaluation, outweigh reasons to accept, e.g., good evaluation. Please use sparingly.
- 2: Reject: For instance, a paper with technical flaws, weak evaluation, inadequate reproducibility and incompletely addressed ethical considerations.
- 1: Strong Reject: For instance, a paper with well-known results or unaddressed ethical considerations

**Key Reason:** One or two sentences justifying your decision.

**Confidence:**
- 5: Absolutely certain - very familiar with related work, checked details carefully
- 4: Confident, but not absolutely certain - unlikely to have missed key parts
- 3: Fairly confident - possible gaps in understanding, details not fully checked
- 2: Willing to defend - likely missed central parts or unfamiliar with related work
- 1: Educated guess - outside your area or submission was difficult to understand

**Core evaluative criterion:** Does this work contribute new, relevant, and impactful knowledge to the NeurIPS community? State-of-the-art results are not required - rigorous and convincing demonstration of value is sufficient.

Review:
\end{lstlisting}

\definecolor{reviewbg}{RGB}{240,240,240}
\definecolor{reviewborder}{RGB}{180,180,180}
\definecolor{reviewtitle}{RGB}{220,220,250}

\section{Review examples} \label{sec:review_examples}

We provide a set of review examples (models' outputs) generated for the same paper (\href{https://openreview.net/forum?id=yvql9WAJrx}{yvql9WAJrx}), using different models and prompt strategies.

\begin{tcolorbox}[
    enhanced,
    breakable,
    colback=reviewbg,
    colframe=reviewborder,
    colbacktitle=reviewtitle,
    coltitle=reviewtext,
    fonttitle=\scriptsize\bfseries\sffamily,
    title={Model: \textbf{deepseek} \hfill Prompt Type: \textbf{Guidelines\_Prompt}},
    boxrule=0.5pt,
    arc=2pt,
    drop shadow=black!3!white,
    left=8pt, right=8pt, top=6pt, bottom=6pt
]
\scriptsize\sffamily
\textbf{Summary:}  
The paper presents *DIFFusion*, a novel system leveraging generative models to automatically discover and visualize subtle discriminative features between fine-grained categories. The method produces counterfactual visualizations that transform images minimally between classes while preserving instance identity, applicable even in domains with sparse, unpaired data or poorly verbalized category distinctions. The authors demonstrate the method's effectiveness across six domains (including black hole simulations, butterfly taxonomy, and medical imaging) and validate its educational utility through user studies showing improved category differentiation.

\vspace{0.4em}
\textbf{Strengths:}  
1. \textbf{Novelty:} The approach introduces a practical application of generative models for discovering and teaching subtle visual distinctions without relying onraser, making it valuable for domains where discriminative features are not yet fully understood.  
2. \textbf{Technical Rigor:} The method combines DDPM inversion with conditioning vector manipulation, yielding identity-preserving edits. The paper provides thorough experimentation, qualitative results, and ablation studies.  
3. \textbf{Broad Applicability:} The method works across diverse domains (scientific and non-scientific), handles sparse/unpaired data, and outperforms baselines in generating meaningful counterfactuals.  
4. \textbf{Impactful User Studies:} The teaching-focused experiments demonstrate measurable improvements in human classification accuracy, suggesting practical utility for scientific training and education.  

\vspace{0.4em}
\textbf{Weaknesses:}  
1. \textbf{Generalizability:} While effective in the tested domains, the method’s reliance on CLIP image embeddings may limit applicability to highly specialized datasets where CLIP’s pretraining isn’t sufficiently representative.  
2. \textbf{Dataset Bias:} The paper acknowledges that edits can inadvertently highlight dataset biases (e.g., background vs. foreground features), which may mislead interpretation. More mitigation strategies could be discussed.  
3. \textbf{Computational Cost:} The paper does not quantify training/inference costs, which could impact scalability for large-scale applications.  

\vspace{0.4em}
\textbf{Questions:}  
1. How does the performance degrade when applied to datasets with significantly more fine-grained classes (e.g., >10 classes)?  
2. Could the method be adapted to discover discriminative features beyond visual (e.g., temporal/spatial patterns in video or 3D data)?  
3. How sensitive is the method to hyperparameter choices (e.g., \(T_{skip}\), \(\omega\))?  

\vspace{0.4em}
\textbf{Limitations:}  
1. Dependence on pretrained CLIP embeddings may restrict use in niche domains lacking representative training data.  
2. Limited exploration of failure cases (e.g., scenarios where edits distort semantics).  
3. User studies are small-scale (n=10 per group); larger studies could strengthen statistical claims.  

\vspace{0.4em}
\textbf{Additional Feedback:}  
The paper is well-written and methodologically sound, but a discussion of ethical considerations (-reporting biases highlighted by the method could strengthen societal impact.  

\vspace{0.4em}
\textbf{Rating:} 8 (Strong accept)  
\textbf{Confidence:} 4 (Very confident)
\end{tcolorbox}

\begin{tcolorbox}[
    enhanced,
    breakable,
    colback=reviewbg,
    colframe=reviewborder,
    colbacktitle=reviewtitle,
    coltitle=reviewtext,
    fonttitle=\scriptsize\bfseries\sffamily,
    title={Model: \textbf{deepseek} \hfill Prompt Type: \textbf{Extended\_Prompt}},
    boxrule=0.5pt,
    arc=2pt,
    drop shadow=black!3!white,
    left=8pt, right=8pt, top=6pt, bottom=6pt
]
\scriptsize\sffamily
\textbf{\#\# Summary  }\\[0.4em]
The paper presents *DIFFusion*, a method that leverages generative models to automatically discover and visualize subtle discriminative features between categories, particularly in domains where these differences are challenging to articulate or where data is sparse. The approach generates counterfactual visualizations that perform targeted, minimal transformations between classes while preserving instance identity. The authors demonstrate the effectiveness of their method across six diverse domains, including black hole simulations, butterfly taxonomy, and medical imaging. Key contributions include the ability to handle unpaired data, operate in data-scarce scenarios, and improve human ability to differentiate fine-grained categories through generated counterfactuals. User studies confirm that the method significantly outperforms traditional approaches in teaching humans to recognize subtle visual differences.

\vspace{0.4em}
\textbf{\#\# Strengths  }\\[0.4em]
- \textbf{Novelty}: The paper introduces a novel application of diffusion models for counterfactual generation, particularly focusing on subtle, often undiscovered discriminative features in scientific domains.  
- \textbf{Technical Rigor}: The method is well-grounded in diffusion model theory, with clear explanations of inversion, conditioning, and sampling processes.  
- \textbf{Broad Applicability}: The experiments span diverse domains (e.g., black holes, medical imaging, taxonomy), demonstrating the method's versatility.  
- \textbf{User Studies}: The inclusion of human evaluations provides strong evidence for the practical utility of the generated counterfactuals in teaching tasks.  
- \textbf{Reproducibility}: The paper provides sufficient details for replication, including hyperparameters and implementation choices, and mentions code release.  

\vspace{0.4em}
\textbf{\#\# Weaknesses  }\\[0.4em]
- \textbf{Dataset Bias}: While the paper acknowledges dataset bias as a limitation, it does not thoroughly explore how this bias might affect the generalizability of the discovered features (e.g., whether the biases are consistent across domains).  
- \textbf{Scalability}: The method’s performance with very large datasets or high-resolution images is not discussed, which could limit its applicability in some settings.  
- \textbf{Quantitative Metrics}: The paper relies heavily on LPIPS and Success Ratio (SR) for evaluation; additional metrics (e.g., FID for image quality or domain-specific metrics) could strengthen the claims.  
- \textbf{Baseline Comparison}: Some baselines (e.g., TIME) are adapted for unpaired settings, but it’s unclear how fair the comparison is to their original designs.  

\vspace{0.4em}
\textbf{\#\# Questions  }\\[0.4em]
1. \textbf{Dataset Bias Mitigation}: How might the method be adapted to minimize the impact of dataset bias on the discovered discriminative features? Could domain-specific constraints be incorporated?  
   - *Score Impact*: A compelling answer could increase confidence in the method's robustness.  
2. \textbf{Scalability}: How does the method scale with higher-resolution images or larger datasets? Are there computational bottlenecks?  
   - *Score Impact*: Clear scalability analysis would strengthen the practical applicability of the work.  
3. \textbf{Feature Interpretability}: Are the discovered features validated by domain experts (e.g., do astronomers agree that the differences in black hole simulations are scientifically meaningful)?  
   - *Score Impact*: Expert validation would significantly bolster the claims about the method’s utility in scientific discovery.  
4. \textbf{Comparative Analysis}: How does the performance of *DIFFusion* compare to other classifier-guided or disentanglement-based approaches for counterfactual generation?  
   - *Score Impact*: A deeper comparison could clarify the trade-offs between simplicity and performance.  

\vspace{0.4em}
\textbf{\#\# Limitations  }\\[0.4em]
\textbf{Yes}, the authors adequately address limitations, including dataset bias and the simplicity of the arithmetic used for conditioning. They also discuss societal impacts, such as potential misuse of generative models, in the supplemental material.  

\vspace{0.4em}
\textbf{\#\# Numerical Ratings  }\\[0.4em]
- \textbf{Quality}: 4 (excellent)  
- \textbf{Clarity}: 4 (excellent)  
- \textbf{Significance}: 4 (excellent)  
- \textbf{Originality}: 4 (excellent)  

\vspace{0.4em}
\textbf{\#\# Final Recommendation  }\\[0.4em]
\textbf{Overall Score}: 5 (Accept)  
\textbf{Key Reason}: The paper presents a technically sound and novel method with high impact in scientific domains, supported by rigorous experiments and user studies. While some limitations exist, they are openly discussed, and the work contributes valuable knowledge to the NeurIPS community.  
\textbf{Confidence}: 4 (Confident, but not absolutely certain)  
\textbf{Core Evaluative Criterion}: The work introduces a new, impactful application of diffusion models for counterfactual generation, particularly in domains where subtle discriminative features are critical but poorly understood. The rigorous evaluation and user studies convincingly demonstrate its value.
\end{tcolorbox}

\begin{tcolorbox}[
    enhanced,
    breakable,
    colback=reviewbg,
    colframe=reviewborder,
    colbacktitle=reviewtitle,
    coltitle=reviewtext,
    fonttitle=\scriptsize\bfseries\sffamily,
    title={Model: \textbf{gemma} \hfill Prompt Type: \textbf{Guidelines\_Prompt}},
    boxrule=0.5pt,
    arc=2pt,
    drop shadow=black!3!white,
    left=8pt, right=8pt, top=6pt, bottom=6pt
]
\scriptsize\sffamily
\textbf{Summary:}
This paper presents DIFFusion, a novel method for generating counterfactual visualizations that highlight subtle differences between categories, even when verbal descriptions are lacking. It leverages generative models and visual algebraic conditioning to create minimal, targeted transformations while preserving instance identity. The method is evaluated across diverse domains, including black hole simulations, butterfly taxonomy, and medical imaging, with user studies demonstrating its effectiveness in teaching humans to differentiate fine-grained classes.  The approach appears promising for applications requiring the identification and understanding of subtle visual distinctions, particularly in scientific domains where such distinctions are often challenging to articulate.

\vspace{0.4em}
\textbf{Strengths:}
\begin{itemize}
    \item \textbf{Novelty:} The core idea of using counterfactual visualizations generated by diffusion models to reveal and teach subtle visual differences is innovative and addresses a critical gap in current learning and interpretability methods.
    \item \textbf{Technical Soundness:} The method builds on established generative model techniques (diffusion models, CLIP embeddings) and cleverly combines them with image prompt manipulation to achieve identity preservation and targeted edits.
    \item \textbf{Empirical Validation:} The paper presents compelling empirical results across a variety of domains, demonstrating both the ability to discover subtle differences (e.g., in black hole simulations) and the effectiveness of generated counterfactuals in improving human learning.
    \item \textbf{User Study Results:} The user study results provide strong evidence that DIFFusion significantly outperforms traditional approaches in teaching people to differentiate fine-grained classes.
    \item \textbf{Dataset Bias Visualization:} The observation and visualization of dataset biases through counterfactual edits is a valuable contribution.
\end{itemize}

\vspace{0.4em}
\textbf{Weaknesses:}
\begin{itemize}
    \item \textbf{Computational Cost:} Diffusion models are inherently computationally expensive. While the paper doesn't provide explicit resource requirements, scaling DIFFusion to very large datasets or real-time applications could be challenging.
    \item \textbf{Dependence on CLIP embeddings:} The reliance on CLIP embeddings can introduce biases present in those embeddings. Changes driven by CLIP might not always reflect the "true" differences between categories.
    \item \textbf{Limited Control over Specific Features:} Although the method aims for minimal edits, the level of control over precisely \textit{which} features are modified remains somewhat limited.  The manipulation scale $\omega$ provides some control, but it’s not granular.
    \item \textbf{Sensitivity to Hyperparameters:} The performance likely depends on carefully tuning hyperparameters like \texttt{T\_skip} and the manipulation scale $\omega$.
    \item \textbf{Black Hole Dataset Specifics}: Lack of deeper information on the black hole data generation and expert validation beyond the claims made. Is the “novelty” of the discoveries fully substantiated?
\end{itemize}

\vspace{0.4em}
\textbf{Questions:}
\begin{itemize}
    \item Could you provide a more detailed breakdown of the computational resources required for training and inference with DIFFusion, particularly for the larger datasets?
    \item How sensitive is the performance of DIFFusion to the choice of hyperparameters (e.g., \texttt{T\_skip}, manipulation scale $\omega$)? Could you provide a systematic ablation study exploring the impact of these parameters?
    \item Are there extensions of CLIP embeddings that could further improve the quality of counterfactuals? For example, by incorporating domain-specific knowledge or fine-tuning CLIP on the target dataset?
    \item How does the method handle cases where the visual differences between categories are not well-separated in the CLIP embedding space?
    \item Specifically regarding the Black Hole dataset: How were the MAD and SANE models validated by experts prior to using the images in DIFFusion? Were these novel features actually recognized by experts after the visualizations were revealed?
\end{itemize}

\vspace{0.4em}
\textbf{Limitations:}
\begin{itemize}
    \item While the user studies demonstrate improved learning, larger and more controlled studies across different demographic groups might be required.
    \item The method's effectiveness is likely dependent on the quality of the generative model.
    \item It's likely limited by the scale of manipulation without sacrificing instance identity.
\end{itemize}

\vspace{0.4em}
\textbf{Additional feedback:}

The paper is well-written and presented. The figures are clear and effectively illustrate the method's capabilities. A more detailed discussion about how to choose \texttt{T\_skip} and $\omega$ would improve accessibility. Consider adding a few qualitative failure cases – illustrating situations where DIFFusion produces suboptimal results – to give a more balanced perspective. Including the variance schedule for the diffusion process might increase reproducibility.

\vspace{0.4em}
\textbf{Rating:} 8

\vspace{0.4em}
\textbf{Confidence:} 4

\end{tcolorbox}

\begin{tcolorbox}[
    enhanced,
    breakable,
    colback=reviewbg,
    colframe=reviewborder,
    colbacktitle=reviewtitle,
    coltitle=reviewtext,
    fonttitle=\scriptsize\bfseries\sffamily,
    title={Model: \textbf{gemma} \hfill Prompt Type: \textbf{Extended\_Prompt}},
    boxrule=0.5pt,
    arc=2pt,
    drop shadow=black!3!white,
    left=8pt, right=8pt, top=6pt, bottom=6pt
]
\scriptsize\sffamily
\textbf{\#\# Summary}\\[0.4em]

\vspace{0.4em}
This paper introduces DIFFusion, a novel method for generating counterfactual visualizations that highlight subtle differences between categories, even when those differences are difficult to articulate or known a priori. DIFFusion leverages diffusion models to generate targeted transformations between classes, preserving instance identity while emphasizing minimal discriminative features. The authors demonstrate the effectiveness of their approach across six diverse domains, including black hole simulations, butterfly taxonomy, and medical imaging, showcasing not only established discriminative features but also previously undiscovered subtle distinctions. User studies confirm that counterfactuals generated by DIFFusion significantly improve people's ability to differentiate between fine-grained classes, suggesting a promising application for human visual learning and scientific research.

\vspace{0.4em}
\textbf{\#\# Strengths}\\[0.4em]

\vspace{0.4em}
*   \textbf{Novelty:} The core idea of using diffusion models to discover and visualize subtle, minimal discriminative features is innovative. Combining visual algebraic conditioning with diffusion models for counterfactual generation is a significant contribution.
*   \textbf{Practical Application:} The approach addresses a real-world challenge: identifying and teaching subtle visual differences that are difficult to articulate, particularly valuable in scientific domains.
*   \textbf{Comprehensive Evaluation:} The paper evaluates DIFFusion across a diverse set of domains, demonstrating its generalizability and robustness. The inclusion of both quantitative and qualitative analysis provides a well-rounded assessment.
*   \textbf{User Study Validation:} The user study provides strong evidence supporting the method's ability to enhance human visual learning and category differentiation, a crucial validation of the approach’s practical value.
*   \textbf{Addressing Limitations:} The paper explicitly acknowledges and discusses the limitations of the method, particularly regarding dataset bias and the simplification of edits. The potential for dataset bias visualization is a particularly insightful outcome.

\vspace{0.4em}
\textbf{\#\# Weaknesses}\\[0.4em]

\vspace{0.4em}
*   \textbf{Technical Depth of the Method:} While the method is described conceptually, the technical details of the visual algebraic conditioning process and the specific implementation choices (e.g., the strength of embedding manipulation, T\_skip parameter) could be more thoroughly explained.
*   \textbf{Hyperparameter Sensitivity:} A more thorough exploration of the sensitivity of the results to hyperparameter choices (especially $\omega$ and $T_{skip}$) would strengthen the paper. Understanding the robustness of the method under different parameter settings is important.
*   \textbf{Comparison to other Editing Techniques:} While comparisons to baseline methods are provided, a more detailed discussion of how DIFFusion compares to other image editing techniques (e.g. photorealistic editing) and their tradeoffs could be valuable.
*   \textbf{Dataset Selection:} While diverse, the datasets used might benefit from including even more complex scenarios, where subtle differences are truly undetectable by humans without specialized tools.
*   \textbf{Ethical Considerations:} While the paper touches on dataset bias, a more in-depth discussion about potential misuse or biases amplified by the generated counterfactuals (e.g., perpetuation of stereotypes) is warranted.

\vspace{0.4em}
\textbf{\#\# Questions}\\[0.4em]

\vspace{0.4em}
1.  Could you provide a more detailed explanation of how the "visual algebraic conditioning" works in practice, including the specific formula used for calculating the direction $\Delta c$? What are the limitations of this approach, and are there alternative methods for deriving this direction that could be explored? *Increased confidence would come from a more detailed explanation of the conditioning process and a discussion of alternative approaches.*
2.  How does the performance of DIFFusion vary with different values of $\omega$ (manipulation scale) and $T_{skip}$? Are there optimal ranges for these parameters, and how are they selected for each dataset? *Increased confidence would come from a sensitivity analysis demonstrating the robustness of the results to these parameter choices.*
3.  Beyond the user study, are there any objective metrics (e.g., accuracy of image classification with the edited images) that can quantify the effectiveness of the generated counterfactuals in highlighting true discriminative features, independent of human perception? *Increased confidence would come from demonstrating alignment between the generated counterfactuals and objective image classification performance.*
4.  Could you elaborate on the types of dataset biases that can be revealed by DIFFusion, and how these biases might influence the generated counterfactuals?  Can the visualization of dataset bias through counterfactual changes serve as a tool for dataset curation or refinement?*A more in-depth connection between bias visibility and potential mitigation would increase perceived value.*
5.  How does the method scale to higher resolution images, and what are the computational costs associated with generating counterfactuals at different resolutions? *Knowing the scalability is pertinent to adapt DIFFusion for complex scientific data.*

\vspace{0.4em}
\textbf{\#\# Limitations}\\[0.4em]

\vspace{0.4em}
The paper adequately addresses the limitations of the work, particularly the reliance on embedding differences and the simplification of image editing. However, a more in-depth discussion of potential biases amplified by the generated counterfactuals and their broader societal implications (e.g., reinforcement of stereotypes) would strengthen the paper. The potential for misuse of the technique, creating misleading visualizations, also deserves more attention.

\vspace{0.4em}
\textbf{\#\# Numerical Ratings}\\[0.4em]

\vspace{0.4em}
*   \textbf{Quality:} 3.5 (Good) - Requires more technical depth, but the core concept and validation are promising.
*   \textbf{Clarity:} 3 (Good) - Conceptually clear, but technical details could be more explicit.
*   \textbf{Significance:} 3.5 (Good) - Addresses an important problem in a novel way, with potential impact on both scientific research and human visual learning.
*   \textbf{Originality:} 4 (Excellent) - The core idea and implementation are highly original.

\vspace{0.4em}
\textbf{\#\# Final Recommendation}\\[0.4em]

\vspace{0.4em}
\textbf{Overall Score:} 4 (Borderline Accept)

\vspace{0.4em}
\textbf{Key Reason:} The paper introduces a promising new technique for visualizing subtle differences and improving human visual learning. While the technical details require further clarification and a more thorough exploration of sensitivities, the core concept is compelling, the evaluation is comprehensive, and the user study provides strong validation.

\vspace{0.4em}
\textbf{Confidence:} 4 (Confident, but not absolutely certain) - Areas for improvement could elevate this paper.
\end{tcolorbox}

\begin{tcolorbox}[
    enhanced,
    breakable,
    colback=reviewbg,
    colframe=reviewborder,
    colbacktitle=reviewtitle,
    coltitle=reviewtext,
    fonttitle=\scriptsize\bfseries\sffamily,
    title={Model: \textbf{gpt5-1} \hfill Prompt Type: \textbf{Guidelines\_Prompt}},
    boxrule=0.5pt,
    arc=2pt,
    drop shadow=black!3!white,
    left=8pt, right=8pt, top=6pt, bottom=6pt
]
\scriptsize\sffamily
\textbf{Summary:}
The paper proposes DIFFusion, a simple and practical framework for generating identity-preserving visual counterfactuals that highlight minimal, discriminative features between two classes. The method relies on (i) edit-friendly DDPM inversion to extract per-timestep noise maps that anchor instance identity, and (ii) conditioning-space arithmetic on image embeddings (CLIP) using a difference-of-class-means direction Δc, scaled by ω, to steer the diffusion sampler from source to target class. An optional LoRA-based domain tuning is used when helpful. The approach is evaluated across six domains (including black hole simulations, butterflies, retina images, AFHQ, CelebA-HQ, and KikiBouba) against several baselines (TIME, Concept Sliders, EF-DDPM with text prompts, and Textual-Inversion + EF-DDPM). Quantitatively, DIFFusion attains high flip rates with competitive or improved LPIPS. A small user study (3 groups of 10 participants each) suggests the counterfactual videos produced by DIFFusion significantly improve human fine-grained discrimination compared to unpaired images and to the best baseline’s counterfactuals. The paper also analyzes sensitivity to the number of images used to compute Δc and demonstrates how the method can surface dataset bias.

\vspace{0.4em}
\textbf{Strengths:}
- Conceptual simplicity with strong intuition: combine edit-friendly DDPM inversion (for identity preservation) with a class-direction in an image-embedding space (for controlled, subtle edits).
- Works in data-scarce, unpaired settings with minimal supervision; avoids classifier or text dependence, which often limits prior counterfactual methods.
- Broad evaluation across diverse datasets, including scientific imagery where verbal attributes are hard to specify; results suggest practical utility for subtle visual distinctions.
- Clear, actionable parameters (ω, T\_skip) that enable smooth interpolation and control over edit strength and identity adherence.
- User study provides initial evidence that counterfactual videos produced by the method can meaningfully improve human learning of fine-grained distinctions.
- Sensitivity analysis on number of images used for Δc and a thoughtful discussion on dataset bias (both as limitation and as a diagnostic feature).
- Comparisons to multiple relevant baselines, including adaptation of Concept Sliders to unpaired data and TIME without classifier access.

\vspace{0.4em}
\textbf{Weaknesses:}
- Novelty is moderate: Δc as a difference-of-means in CLIP space is well known; the main contribution is the particular integration with EF-DDPM inversion and image-conditioned diffusion for counterfactuals. The paper would benefit from a crisper positioning versus prior counterfactual work that also exploits embedding directions or inversion-based edits.
- Evaluation relies heavily on “flip rate” against an oracle classifier trained on the same data distribution. This risks conflating true discriminative cues with spurious features captured both by Δc and the oracle. While the bias section acknowledges this, a stronger, bias-robust evaluation (e.g., held-out classifiers with varied architectures or debiased training) is needed.
- Baseline fairness and tuning are not fully transparent. TIME and Concept Sliders are adapted, but the details of hyperparameter search breadth, prompts, and per-dataset tuning parity (and the extent of domain tuning allowed for baselines) are insufficiently specified in the main paper.
- User study is small (n=30, 10 per group), short (3 minutes training), and missing key procedural details (randomization, preregistration, blinding, statistical test specifics). The reported p-values lack methodological clarity. Results are promising but preliminary.
- Claims that DIFFusion revealed previously unknown black-hole discriminative features would benefit from expert validation; currently there’s no reported domain-expert assessment.
- Clarity/writing: numerous typos, duplicated introduction paragraphs, and formatting issues (figure references mismatched, minor grammar errors). These impede readability and confidence.
- Medical domain usage (retina): plausibility and clinical validity of edits are not vetted by specialists, raising concerns about interpretability in high-stakes settings.

\vspace{0.4em}
\textbf{Questions:}
- How robust are the results to the choice of embedding space? Have you tried OpenCLIP variants, DINOv2, or domain-tuned encoders for Δc, and how do they affect identity preservation and flip rate?
- Can you report results using multiple, held-out oracle classifiers trained with different data augmentations/bias mitigation schemes to test for overfitting to dataset biases?
- How are ω and T\_skip selected per dataset at test time? Is there a validation protocol, and how sensitive are results to these hyperparameters?
- For baseline parity: did you allow comparable domain adaptation (e.g., LoRA) and similar breadth of tuning for TIME and Concept Sliders? Please provide a table of per-dataset hyperparameters and search grids for all methods.
- For the user study: what statistical test was used (t-test, Wilcoxon, etc.)? Were the test sets identical across groups? How were participants recruited and screened for prior knowledge? Any inter-rater or intra-subject variability analyses?
- In retina and black-hole domains, did domain experts review the edits for plausibility and faithfulness? If so, can you report inter-expert agreement or qualitative feedback?
- Does applying the counterfactual edit and then inverting the direction (round-trip) approximately reconstruct the original image (a form of cycle consistency)? This could quantify edit minimality and reversibility.
- What is the computational cost per image (inversion + sampling) and how does it compare to TIME and Concept Sliders in your setup?

\vspace{0.4em}
\textbf{Limitations:}
- Dependence on class-mean embedding direction can encode spurious correlations; edits may reflect contextual biases rather than target attributes (acknowledged).
- No theoretical guarantee of minimality; the edit is “minimal” only w.r.t. T\_skip selection and LPIPS, which may not align with human perception or task semantics.
- Reliance on an oracle classifier for quantitative evaluation risks circularity; stronger de-biasing and cross-classifier tests are needed.
- User study is small-scale and lacks methodological detail; results should be seen as preliminary.
- Domain tuning and inversion choices may limit portability to other models or modalities without additional engineering.

\vspace{0.4em}
\textbf{Additional feedback:}
- Improve clarity and presentation: remove duplicated paragraphs, fix typographical/grammar issues, and ensure all figures, captions, and cross-references are correct. Define terms like “visual algebraic conditioning” precisely and early.
- Strengthen evaluation: (i) report confidence intervals on SR and LPIPS; (ii) include ablations over embedding choices (CLIP variants, DINOv2), inversion methods, and with/without LoRA; (iii) assess robustness with multiple, architecturally diverse, and bias-mitigated oracle classifiers; (iv) add cycle-consistency and localized change metrics (e.g., change maps, segmentation overlap).
- Baselines: detail hyperparameter search ranges and selection protocols to ensure fairness; where possible, enable domain adaptation for baselines too.
- User study: expand sample size; pre-register hypotheses and analysis; specify tests and corrections for multiple comparisons; include expert studies for scientific domains (retina, black holes).
- Consider regularizing Δc (e.g., sparsity/attention masks or feature-space disentanglement) to suppress background/contextual drift and encourage edits on task-relevant regions.
- Discuss potential safety considerations and deployment guardrails, especially for medical and scientific training contexts.

\vspace{0.4em}
\textbf{Rating:}
6 (Weak accept)

\vspace{0.4em}
\textbf{Confidence:}
3 (Somewhat confident)
\end{tcolorbox}

\begin{tcolorbox}[
    enhanced,
    breakable,
    colback=reviewbg,
    colframe=reviewborder,
    colbacktitle=reviewtitle,
    coltitle=reviewtext,
    fonttitle=\scriptsize\bfseries\sffamily,
    title={Model: \textbf{gpt5-1} \hfill Prompt Type: \textbf{Extended\_Prompt}},
    boxrule=0.5pt,
    arc=2pt,
    drop shadow=black!3!white,
    left=8pt, right=8pt, top=6pt, bottom=6pt
]
\scriptsize\sffamily
\textbf{\#\# Summary}

\vspace{0.4em}
This paper proposes DIFFusion, a method to generate instance-preserving, class-changing counterfactuals aimed at revealing and teaching subtle visual differences between categories, especially in scientific domains where distinctions are hard to verbalize and data can be scarce and unpaired. The method inverts a real image into edit-friendly DDPM noise maps, conditions a diffusion decoder on an image-prompt embedding (from a CLIP image encoder), and steers the generated image by simple conditioning-space arithmetic: adding a class-direction vector computed as the difference of mean embeddings between target and source classes. Identity is preserved by reusing the inverted noise maps and starting sampling from an intermediate step $(T - T_{\text{skip}})$, while the manipulation strength is adjusted with a scalar $\omega$. Optionally, the diffusion model is LoRA fine-tuned (domain tuning) for certain datasets.

\vspace{0.4em}
The authors evaluate across six datasets (AFHQ, CelebA-HQ, KikiBouba, Retina OCT, Monarch/Viceroy butterflies, and GRMHD black hole simulations MAD vs SANE). Quantitatively, they report high ``Success Ratio'' (flip of an oracle classifier) and good perceptual similarity (LPIPS) compared to baselines (TIME, Stable Diffusion + EF-DDPM with prompts, Textual Inversion + EF-DDPM, and Concept Sliders with visual objectives). Qualitatively, the counterfactuals appear targeted and subtle (e.g., Viceroy cross-vein, changes in black hole photon ring/wisps, drusen removal). A user study (3 groups of $n=10$ per dataset for Black Holes and Butterflies) shows that training with their counterfactual transitions substantially improves participants' test accuracy over unpaired images and over the best-performing baseline's transitions, with reported statistical significance. They also show that DIFFusion can expose dataset bias when class-mean directions reflect contextual rather than intrinsic features. The paper includes ablations on number of images used to compute class means, interpolation by $\omega$, and some discussion of limitations.

\vspace{0.4em}
\textbf{\#\# Strengths}
\begin{itemize}
    \item Clear problem motivation in scientific domains: learning and communicating subtle, often non-verbalizable visual distinctions.
    \item Simple, elegant method: class-mean direction in CLIP image-embedding space combined with EF-DDPM inversion and image-prompt-conditioned diffusion decoding. The design aligns well with the goal of minimal, instance-preserving edits.
    \item Broad and diverse evaluation across six domains, including challenging scientific datasets where text guidance is weak.
    \item Strong empirical performance: consistently high success ratio and competitive or superior LPIPS versus carefully chosen baselines.
    \item Human-subjects user study demonstrating pedagogical value (substantial gains over unpaired images and over a strong baseline), with reported $p$-values.
    \item Analysis of data requirements for class-mean estimation (LPIPS stabilizes around $\sim$50 images/class) and interpolation behavior via $\omega$.
    \item Discussion of dataset bias visualization: the method can surface spurious/contextual signals learned by the embedding, which is both a limitation and a diagnostic feature.
    \item Practicality: method uses off-the-shelf components (CLIP-like image encoder, DDPM inversion, LoRA) and is easy to implement and tune.
\end{itemize}

\vspace{0.4em}
\textbf{\#\# Weaknesses}
\begin{itemize}
    \item Method novelty is incremental: conditioning-space arithmetic using class means in CLIP-like spaces is known; the main contribution is the particular combination with image-prompt diffusion inversion for identity-preserving counterfactuals and the application focus. This is valuable but not fundamentally new algorithmically.
    \item Baseline fairness and completeness need clarification:
    \begin{itemize}
        \item DIFFusion optionally applies domain tuning (LoRA) on some datasets; it is unclear whether comparable domain adaptation was allowed for text-centric baselines (e.g., TI, Concept Sliders) or for image-prompt adapters (e.g., IP-Adapter-like approaches). Unequal adaptation could bias results.
        \item TIME and text-driven editing are arguably disadvantaged on scientific datasets with weak textual grounding; inclusion of additional image-prompt or image-only editing baselines (e.g., IP-Adapter-guided edits, PnP features, recent prompt-agnostic im2im translation/editing) would strengthen claims.
    \end{itemize}
    \item Reliance on an ``oracle classifier'' trained on the same dataset to define success risks conflating ``flipping a potentially biased classifier'' with ``achieving a human-meaningful class change.'' The user study helps, but there is limited assessment of whether the method sometimes edits spurious cues (beyond the qualitative bias discussion).
    \item Quantitative reporting lacks variability estimates for the main automated metrics (SR and LPIPS) across runs/images; only user studies include $p$-values. Error bars or CIs for core metrics would improve rigor.
    \item Important implementation specifics are deferred to the supplement:
    \begin{itemize}
        \item CLIP variant and training specifics for the image encoder used by the diffusion decoder; CFG scales and the ``perfect reconstruction'' tuning procedure; details of EF-DDPM inversion parameters per dataset; exact sizes of the sets used to compute class means (beyond ablation); domain tuning hyperparameters and data.
    \end{itemize}
    \item Some clarity/typographical issues in the draft: duplicated introduction paragraphs; minor typos (e.g., ``whisp is,'' ``wellexcel''), occasional abrupt figure references and broken sentences. This detracts from readability.
\end{itemize}

\vspace{0.4em}
\textbf{\#\# Questions}
\begin{enumerate}
    \item Domain tuning and baseline parity: Did you allow comparable domain adaptation (e.g., LoRA or analogous adapters) for the baselines, especially TI + EF-DDPM and Concept Sliders? If not, please provide results with matched adaptation or justify why this is infeasible. A clear statement of which components were tuned for which methods would strengthen fairness.
    \item Embedding and inversion details: Which CLIP (or CLIP-like) image encoder variant backs the diffusion decoder (e.g., ViT-B/16/32, training dataset)? What CFG scales and inversion hyperparameters were used per dataset? How sensitive are results to these choices? Providing a small ablation or ranges would help.
    \item Success Ratio validity: How often do counterfactuals achieve classifier flips via background/context changes rather than core object features? Beyond the qualitative bias examples, can you quantify this (e.g., with a simple foreground mask analysis, human ratings of ``where the change occurred,'' or testing on debiased splits)?
    \item Baseline completeness: Could you include an image-prompt adapter baseline (e.g., IP-Adapter-style conditioning) and a recent prompt-to-prompt edit method configured for image guidance to better match your modality? Even a small-scale comparison on scientific datasets would make the empirical case stronger.
    \item User study details and robustness: Please clarify recruitment, compensation, randomization, exclusion criteria (if any), and the exact statistical tests used (including effect sizes). Would the results hold with longer study time, more participants, or retention tests (delayed post-test) to measure knowledge transfer/durability?
\end{enumerate}

\vspace{0.4em}
If the authors can (i) equalize domain adaptation across methods or add stronger image-prompt baselines, (ii) clarify/expand implementation details and statistical reporting, and (iii) add a small quantification of where edits occur (object vs. context), my score would increase.

\vspace{0.4em}
\textbf{\#\# Limitations}

\vspace{0.4em}
Yes. The paper explicitly discusses key limitations: reliance on embedding-based differences (which can reflect dataset bias and shift unintended features), the simplicity of difference-of-means arithmetic, and trade-offs in edit precision. I suggest also adding:
\begin{itemize}
    \item A brief discussion of clinical/astronomical interpretability and validation risks (e.g., for retina OCT, ensuring no misleading artifacts) and how expert review might be integrated.
    \item Potential safeguards when releasing models or teaching materials in sensitive domains to avoid overconfidence if counterfactuals highlight spurious cues.
\end{itemize}

\vspace{0.4em}
\textbf{\#\# Numerical Ratings}

\vspace{0.4em}
\textbf{\#\#\# Quality:}\\
3 (good)

\vspace{0.4em}
\textbf{\#\#\# Clarity:}\\
2 (fair)

\vspace{0.4em}
\textbf{\#\#\# Significance:}\\
3 (good)

\vspace{0.4em}
\textbf{\#\#\# Originality:}\\
3 (good)

\vspace{0.4em}
\textbf{\#\# Final Recommendation}

\vspace{0.4em}
Overall Score: 4: Borderline accept

\vspace{0.4em}
Key Reason: A simple but well-executed combination that convincingly advances counterfactual teaching for subtle, non-textual distinctions in scientific imagery, with strong empirical/user-study evidence. However, clarity issues, baseline parity concerns (especially around domain tuning), and limited quantitative analyses of spurious edits temper enthusiasm. With clarifications and minor additions, this could be a solid accept.

\vspace{0.4em}
Confidence: 4 (Confident, but not absolutely certain)
\end{tcolorbox}

\begin{tcolorbox}[
    enhanced,
    breakable,
    colback=reviewbg,
    colframe=reviewborder,
    colbacktitle=reviewtitle,
    coltitle=reviewtext,
    fonttitle=\scriptsize\bfseries\sffamily,
    title={Model: \textbf{open\_reviewer} \hfill Prompt Type: \textbf{Guidelines\_Prompt}},
    boxrule=0.5pt,
    arc=2pt,
    drop shadow=black!3!white,
    left=8pt, right=8pt, top=6pt, bottom=6pt
]
\scriptsize\sffamily
\textbf{\# Summary}\\[0.4em]
The authors use diffusion models to generate visual counterfactuals (CFs) for recognizing subtle differences between fine-grained categories. Following Laine \& Aila (2022), the authors learn a category’s CF by identifying the image features that flip the class label from positive to negative  (or vice versa). Different from Laine \& Aila, the authors use image contents to guide the learning.

\vspace{0.4em}
In a  and controlled user study, the authors show that these visual CFs improves users' ability to distinguish target fine-grained classes in images.

\vspace{0.4em}
The authors evaluate their method using two quantitative metrics. (1) The success ratio measures if the classifier’s decision changes with the generated visual CF. (2) The LPIPS metric measures the perceptual distance between the original and the generated image.

\vspace{0.4em}
\textbf{\#\# Strengths}\\[0.4em]
S1. \textbf{Interesting application.}

\vspace{0.4em}
S2. \textbf{Contrasting with image content is an interesting idea.}

\vspace{0.4em}
\textbf{\#\# Weaknesses}\\[0.4em]
W1. \textbf{Evaluation using only one CF image per sample.}

\vspace{0.4em}
W2. \textbf{No human gating.}

\vspace{0.4em}
The paper addresses an interesting application with a novel approach. However, any one test sample can have many CFs, and yet the model generates only one CF per sample. Presumably, a human observing a test sample could identify multiple differences that distinguish the two class categories.

\vspace{0.4em}
By generating only one CF per sample, the authors might be overestimating the visual CF’s ability to help humans distinguish the target classes. Generating and displaying many visual CFs per sample might make the task easier for the viewers.

\vspace{0.4em}
W3. \textbf{Only one model in the user study.}

\vspace{0.4em}
W4. \textbf{Unclear if the gains are from visual CFs or increased contrast in general.}

\vspace{0.4em}
In the user study, the authors compare their visual CF approach with unpaired examples not contrast-enhanced in any way. A human observer can manually increase the contrast in the images to improve his/her recognition accuracy. In other words, visual CF could help the viewer only if the viewer knows what to look at (from viewing images in the unpaired dataset).

\vspace{0.4em}
\textbf{References}
Laine \& Aila (2022). Visual counterfactual explanations. CVPR 2022.

\vspace{0.4em}
---------------

\vspace{0.4em}
I am raising my score from \textbf{3} to \textbf{4} because the authors:
1. Clarify if the presented quantitative results may be due to differences in the general contrast in the images. I am not convinced these results are not affected by the human observers' ability to look at fine-grained differences. See comments below.
2. Address my concern on generating only one (visual) counterfactual image per test sample.

\vspace{0.4em}
\textbf{\#\# Questions}\\[0.4em]
Q1 (Clarification). L180–L182: “We manipulate the conditioning space using Equation 5, adjusting the manipulation guidance scale per dataset.” What is the manipulation guidance scale? or What does it mean to manipulate? or Why would you need to scale manipulations?

\vspace{0.4em}
Q2. See W4: Clarify if the gain in the recognition accuracy is because one (or several) of the visual counterfactual images may have been generated. If the gain is only or mostly because of the contrast, then your method would be only marginally useful. Correct?

\vspace{0.4em}
Q3. In Table 1, rows 2-4, what is the point of showing percentages of the sample? If each cell would have just one number then we could compare the numbers easily (1.0 vs. 0.05 or 1.0 vs. 0.07). Right now it is hard to get a sense of the differences using percentages in scientific notation.

\vspace{0.4em}
Q4. Table 1, rows 2-4. Do you have any intuitions for why there is a 9\% difference between the maximum percentages for two datasets (for row 5)?

\vspace{0.4em}
Q5. Table 1, row 4. What do you mean by “0 decimals”? You show two decimals in the CelebA row.

\vspace{0.4em}
Q6. L276-278? Why would you mask out foreground objects?

\vspace{0.4em}
Q7. L2-5 in Abstract. Often the differences are known and articulated (but difficult to detect by a computer with limited data).

\vspace{0.4em}
\textbf{\#\# Limitations}\\[0.4em]
See my questions.

\vspace{0.4em}
\textbf{\#\# Additional feedback}\\[0.4em]
L57-59. Topic/relevance is similar to the Visual Question Answering. Could mention VQA as a related area of applications or motivation. VQA helps computers to recognize subtle visual differences. Your method could be used to replace the multiple-choice questions in classifying fine-grained visual categories. See, for example, “BMVC 2018 competition on visual question answering.”

\vspace{0.4em}
L119. One more reference for “a picture is worth a thousand words.”

\vspace{0.4em}
L120. If your method becomes known as DIFFusion, then you should simply write DIFFusion in the text. It is weird to write Diffusion with lowercase d.

\vspace{0.4em}
\textbf{\#\# Rating}\\[0.4em]
4

\vspace{0.4em}
\textbf{\#\# Confidence}\\[0.4em]
4
\end{tcolorbox}

\begin{tcolorbox}[
    enhanced,
    breakable,
    colback=reviewbg,
    colframe=reviewborder,
    colbacktitle=reviewtitle,
    coltitle=reviewtext,
    fonttitle=\scriptsize\bfseries\sffamily,
    title={Model: \textbf{open\_reviewer} \hfill Prompt Type: \textbf{Extended\_Prompt}},
    boxrule=0.5pt,
    arc=2pt,
    drop shadow=black!3!white,
    left=8pt, right=8pt, top=6pt, bottom=6pt
]
\scriptsize\sffamily
You are an expert ML researcher reviewing a paper submitted to NeurIPS 2025.
Your knowledge cutoff: 2025-05

\vspace{0.4em}
\textbf{\# Detailed comments}\\[0.4em]
1. There is an issue with paper duplication. The author submitted two nearly identical papers. I am reviewing Paper 1: Paper titled "Teaching Humans Subtle Differences with DIFFusion" with Paper ID 8917.

\vspace{0.4em}
2. Many details indicate this paper was written hastily. For instance, the first two sentences of the abstract are nearly identical.

\vspace{0.4em}
3. Related work is lacking in machine teaching, which is the most relevant field. Relevant works include: "Learning to Teach by Learning" (2021), "Machine Teaching Optimized by Human Feedback" (2022), and "Deep Learning Examples for Visual Classification" (2023).

\vspace{0.4em}
4. The teaser figure is confusing. As of now, it only shows three real images, making it difficult to comprehend. In my opinion, the authors intended to use an innovative concept image, but the implementation didn't turn out well. Therefore, I suggest removing this figure.

\vspace{0.4em}
5. Is it reasonable to have both "Scientific Datasets" and "Regular Datasets" as separate sections in Table 2? I think these are only used during training, so they shouldn't appear in separate sections in Table 2. Moreover, the LPIPS values in Table 2 differ significantly. This metric should not be used as a comparison for this method.

\vspace{0.4em}
6. In Lines 296-307, seven "baselines" are discussed, but only the best one is implemented. I find the authors' approach respectable, as they attempt to assess the baseline's 95\% confidence intervals in Table 3. However, I believe they are overestimating their baselines. Their baseline shows a much higher improvement, suggesting that it may be poorly chosen.

\vspace{0.4em}
7. Domain tuning is a well-studied method. There is no need for a description in the Method section. Therefore, I recommend removing this part.

\vspace{0.4em}
8. Since this method requires domain tuning on each specific dataset, it cannot be generalized and remains a domain-specific method, which is a significant drawback.

\vspace{0.4em}
9. How are the "Interpolation" values determined? In Figure 5, $\Delta c$ can be interpreted as the distance between two different classification categories. By varying $w$, you get different distances between the two categories. Therefore, I believe the $w=0$ condition is crucial. I don't think the authors should hide it.

\vspace{0.4em}
10. In Figure 7, I didn't see any noticeable differences between the two categories. Given that there was domain adaptation, I expected the authors to demonstrate clear differences. However, I observed very similar spotty patterns between the two categories. Therefore, I do not believe the method is valid. For instance, the authors' own cited image shows clear differences between Viceroy and Monarch butterflies.

\vspace{0.4em}
\textbf{\#\# Summary}\\[0.4em]
The authors have presented a machine teaching method for humans. However, the dataset, baselines, and the model itself all have significant issues. For example, the model requires domain adaptation, and the baseline performance results are so poor that they are not worth comparing. Therefore, I believe this paper does not pass the standards of NeurIPS.

\vspace{0.4em}
\textbf{\#\# Strengths}\\[0.4em]
1. This paper presents a novel visual teaching method, utilizing counterfactuals. This idea is interesting.

\vspace{0.4em}
2. The combination of $\Delta c$ and $\boldsymbol{z}$ is reasonable.

\vspace{0.4em}
3. The written expression is good with strong cohesion, making this paper easy to read.

\vspace{0.4em}
\textbf{\#\# Weaknesses}\\[0.4em]
1. Baseline performance is too weak, and domain adaptation is required, making this paper somewhat hard to read.

\vspace{0.4em}
2. The authors' explanation for differentiation is somewhat insufficient and requires further improvement.

\vspace{0.4em}
\textbf{\#\# Questions}\\[0.4em]
1. The authors have included so many URLs in the paper, which could potentially expose their identities. Is that necessary?

\vspace{0.4em}
2. Does LoRA have information leakage issues? Discussing this would be better.

\vspace{0.4em}
3. There are many ambiguities in the baselines. In my opinion, the baselines are too weak, making it difficult to compare their method with this baseline. I think the authors aim to confuse reviewers into thinking their method is strong just because the baseline is weak.

\vspace{0.4em}
\textbf{\#\# Limitations}\\[0.4em]
1. This method requires domain adaptation and corresponding training.

\vspace{0.4em}
2. The baseline performance is too poor, making it difficult to compare with this method.

\vspace{0.4em}
3. There are potential information leakage issues in the baseline.

\vspace{0.4em}
Overall, I believe the authors will completely rewrite this paper. The current version has many drawbacks. Therefore, I recommend rejecting this paper.

\vspace{0.4em}
If the authors address my questions, I might increase my score. However, the fundamental improvements required for this paper may go beyond the scope of rewording, necessitating a complete revision.

\vspace{0.4em}
\textbf{\#\# Numerical Ratings}\\[0.4em]
3

\vspace{0.4em}
\textbf{\#\# Confidence}\\[0.4em]
5
\end{tcolorbox}

\begin{tcolorbox}[
    enhanced,
    breakable,
    colback=reviewbg,
    colframe=reviewborder,
    colbacktitle=reviewtitle,
    coltitle=reviewtext,
    fonttitle=\scriptsize\bfseries\sffamily,
    title={Model: \textbf{open\_reviewer} \hfill Prompt Type: \textbf{Open\_Reviewer\_Prompt}},
    boxrule=0.5pt,
    arc=2pt,
    drop shadow=black!3!white,
    left=8pt, right=8pt, top=6pt, bottom=6pt
]
\scriptsize\sffamily
\textbf{\# Review}\\[0.4em]

\vspace{0.4em}
\textbf{\#\# Summary}\\[0.4em]
The paper considers regression settings with a structural weakness typically addressed by a Hoeffding-like inequality, also used by the majority of empirical process theory-based generalization bounds. The paper further assumes a bounded random gap/deficit that depends on the structure between labels and inputs. It then proves upper bounds (one trivial and one less-trivial, but not best-known) on ECE for noisy inputs. It concludes by proving that noise can disrupt structure in such settings, making ECE behave as in the i.i.d case with no structural dependency.

\vspace{0.4em}
\textbf{\#\# Strengths}\\[0.4em]
The paper is quite clear and the problem is well motivated, at least intuitively. In particular, it studies generalization through the lens of structural properties, which lends itself quite well to motivating questions and, further, to reasoning about what is the impact of a particular quantity of interest (input noise, here) to the generalization of learners in this setting. It also makes several well-founded claims about the role of ECE in the presence of structure, or structural properties (for example: "structurally-induced dependence between labels and features: the dependence created by the structure is much stronger than the dependence generated by noise.") Although the interest of these results would have been more tangible if accodatehd by some experiments. 

\vspace{0.4em}
The facts, overall, are not surprising and its mostly a paper that overeldom makes the same calculations as most other works relying on Hoeffding (plus the use of a bounded random gap/deficit), so some might found the volume of novelty to be somewhat slim here

\vspace{0.4em}

\vspace{0.4em}

\vspace{0.4em}
\textbf{\#\# Weaknesses}\\[0.4em]
The paper is quite interesting, but it could have been a lot stronger in the sense that, overall, it's mostly the same calculations as most other works relying on Hoeffding (plus the use of a bounded random gap/deficit). Of course, the interest here is what the structural properties bring to the table when making these calculations. One would expect a very detailed commentary on this exact interplay to be the narrative. But for example, both the structural properties and RAD or bounded-gap conditions are sufficient to achieve the same results, so these could have been explored in tandem far more. 

\vspace{0.4em}
The title is kind of inversed to what the paper actually contains. In empirical process theory, typical ECE-style bounds are not as hard to get as the ones for DCE (covering numbers), so I would expect the paper to cover DCE. 

\vspace{0.4em}
Experiments could have been used to accocdemate some of the narratives. 

\vspace{0.4em}

\vspace{0.4em}
\textbf{\#\# Questions}\\[0.4em]
What experiments can support this claim: 'structurally-induced dependence between labels and features: the dependence created by the structure is much stronger than the dependence generated by noise)'

\vspace{0.4em}

\textbf{\#\# Limitations}\\[0.4em]
See above.

\vspace{0.4em}
\textbf{\#\# Numerical Ratings}\\[0.4em]
3

\vspace{0.4em}
\textbf{\#\# Quality:}\\[0.4em]
3

\vspace{0.4em}
\textbf{\#\# Clarity: }\\[0.4em]
2

\vspace{0.4em}
\textbf{\#\# Significance:}\\[0.4em]
3

\vspace{0.4em}
\textbf{\#\# Originality:}\\[0.4em]
2

\vspace{0.4em}
\textbf{\#\# Final Recommendation}\\[0.4em]
4

\vspace{0.4em}
\textbf{\#\# Confidence: }\\[0.4em]
4
\end{tcolorbox}

\begin{tcolorbox}[
    enhanced,
    breakable,
    colback=reviewbg,
    colframe=reviewborder,
    colbacktitle=reviewtitle,
    coltitle=reviewtext,
    fonttitle=\scriptsize\bfseries\sffamily,
    title={Model: \textbf{qwen} \hfill Prompt Type: \textbf{Guidelines\_Prompt}},
    boxrule=0.5pt,
    arc=2pt,
    drop shadow=black!3!white,
    left=8pt, right=8pt, top=6pt, bottom=6pt
]
\scriptsize\sffamily
\textbf{Summary:}
\textbf{Teaching Humans Subtle Differences with DIFFusion} presents a novel system that leverages generative diffusion models to create counterfactual visualizations for teaching humans to distinguish fine-grained scientific categories. The method combines diffusion image inversion with arithmetic manipulation of class-conditioned CLIP embeddings. The core contribution lies in applying this to scientific domains (black holes, retinal imaging, butterfly taxonomy) where distinguishing features are subtle, data-sparse, and often difficult to articulate. The paper includes quantitative evaluation on flip rates/perceptual distance and a significant user study demonstrating improved human learning with the proposed counterfactuals.

\vspace{0.4em}
\textbf{Strengths:}
1.  \textbf{Novel Application Domain:} The focus on scientific education and discovery using counterfactuals addresses a gap where text-based prompts fail to capture unarticulated expert knowledge.
2.  \textbf{Method Effectiveness:} The combination of DDPM-EF inversion and embedding arithmetic effectively balances identity preservation (LPIPS) with class flipping (Success Ratio).
3.  \textbf{Strong User Study:} The pedagogical value is empirically validated, showing statistically significant improvements over baselines in teaching fine-grained differentiation (Section 4.3).
4.  \textbf{Comprehensive Analysis:} The paper thoughtfully analyzes dataset bias implications (Section 4.5) where the method inadvertently reveals hidden features, acknowledging both a limitation and a potential advantage.

\vspace{0.4em}
\textbf{Weaknesses:}
1.  \textbf{Data Availability:} While code is released access to the Black Hole simulation dataset is restricted (Checklist Q5), which hinders full reproducibility of results on the new scientific domain.
2.  \textbf{Embedding Dependency:} The method relies on CLIP image embeddings to define class directions ($\Delta c$). CLIP may not align perfectly with domain-specific semantics (e.g., GRMHD physics vs. visual patterns), potentially introducing training set biases (Section 4.5).
3.  \textbf{Baseline Comparison:} The comparison with Concept Sliders and TIME assumes access to specific baselines that might be less optimized for unpaired or sparse data, potentially inflating the perceived gain.
4.  \textbf{Limitations in Methodology:} The class manipulation relies on simple mean difference arithmetic, which lacks the nuance of learned disentanglement mechanisms (discussed in Discussion).

\vspace{0.4em}
\textbf{Questions:}
1.  \textbf{Data Reproducibility:} Why is the Black Hole dataset not publicly available? Does the lack of specific data restrict the "generalization" claim of the method as stated in the Introduction? (Checklist Q4/Q5)
2.  \textbf{Multi-class Generalization:} The methodology focuses on binary transitions. Can $`\Delta c`$ be generalized for multi-class difference visualization, and how would the "mean of means" be weighted?
3.  \textbf{Embedding Robustness:} How sensitive is the counterfactual generation to the choice of embedding alignment? Clip embeddings were used, but what happens if the prompt-conditioning encodings behave differently for scientific vs natural images?
4.  \textbf{User Study Bias:} Were the participants shown the counterfactuals *before* learning, or as supplementary material? How do we ensure the "unpaired" condition is controlled for in the baseline compared to the counterfactual groups?

\vspace{0.4em}
\textbf{Limitations:}
*   \textbf{Bias Amplification:} As noted in Section 4.5, the method is sensitive to dataset bias. If the class embedding captures background context rather than intrinsic features, the edits might highlight spurious correlations (e.g., squirrel tail vs. background) rather than scientific distinctions.
*   \textbf{Computational Overhead:} The method requires full diffusion sampling and DDPM inversion ($T$ steps) for every counterfactual generation, which may be costly for real-time applications.
*   \textbf{Feature Articulation:} While the paper claims to discover features, human experts might still struggle to describe *what* they see if the visual change is too subtle or counter-intuitive (Discussed in Limitations).

\vspace{0.4em}
\textbf{Additional feedback:}
The methodology is well-grounded in diffusion inversion literature [21] and suggests direction is a simple vector addition. The separation of "regular" vs "science" datasets makes sense but reinforces the domain gap. The user study design is a strong point; however, the detail on the "3-minute study" period might be too short for complex imagery (black holes). Ensure the supplemental material fully details the text instructions given to participants (Checklist Q14). The paper would benefit from a deeper dive into why simple *difference-of-means* embeddings work better than *contrastive* embeddings for this specific scientific niche.

\vspace{0.4em}
\textbf{Rating:} 8
\textbf{Confidence:} 5
\end{tcolorbox}

\begin{tcolorbox}[
    enhanced,
    breakable,
    colback=reviewbg,
    colframe=reviewborder,
    colbacktitle=reviewtitle,
    coltitle=reviewtext,
    fonttitle=\scriptsize\bfseries\sffamily,
    title={Model: \textbf{qwen} \hfill Prompt Type: \textbf{Extended\_Prompt}},
    boxrule=0.5pt,
    arc=2pt,
    drop shadow=black!3!white,
    left=8pt, right=8pt, top=6pt, bottom=6pt
]
\scriptsize\sffamily
\textbf{\#\# Summary}\\[0.4em]

\vspace{0.4em}
This paper proposes "DIFFusion," a method for generating counterfactual visualizations to help humans recognize subtle differences between fine-grained visual categories, particularly in data-scarce scientific domains. The authors argue that domain experts often struggle to articulate the visual features that distinguish classes (e.g., between Monarch and Viceroy butterflies, or MAD and SANE black hole models). DIFFusion leverages diffusion models to generate visual edits that change the class of an image while preserving its identity. Specifically, the method inverts an input image to noise, manipulates CLIP-based conditioning embeddings via class-mean subtraction, and conditions the generation process on the manipulated noise. The system is evaluated across six domains, mixing scientific (black holes, retinas, butterflies) and general datasets (CelebA, AFHQ). Experiments demonstrate higher class-flip success rates and perceptual fidelity compared to baselines like TIME and Concept Sliders. A user study with 30 participants indicates that viewing the generated counterfactual transitions significantly improves human ability to differentiate classes compared to static images or baseline counterfactuals.

\vspace{0.4em}
\textbf{\#\# Strengths}\\[0.4em]

\vspace{0.4em}
*   \textbf{Novel Application to Visual Learning:} The paper tackles a unique problem at the intersection of scientific visualization and human-centered AI—teaching humans to recognize patterns they know exist but cannot verbalize. This has strong potential for domains like medicine, astronomy, and material science.
*   \textbf{Robust Technical Evaluation:} The quantitative results are strong across diverse datasets. The method consistently achieves the highest Success Ratio (SR) in scientific domains and maintains competitive LPIPS (perceptual similarity) scores, addressing the trade-off between editing significance and identity preservation.
*   \textbf{Rigorous Baselines:} The authors compare against relevant, state-of-the-art methods for counterfactual explanations (TIME, Concept Sliders) and baseline editing (Stable Diffusion + Textual Inversion).
*   \textbf{User-Centric Validation:} The inclusion of a human study to measure pedagogical effectiveness distinguishes this work from purely algorithmic evaluations of counterfactual generation.
*   \textbf{Discussion of Dataset Bias:} The analysis of how class-mean arithmetic captures (and potentially amplifies) dataset biases (e.g., environmental features in plants vs. foreground features in dogs) is insightful and adds depth to the methodology.

\vspace{0.4em}
\textbf{\#\# Weaknesses}\\[0.4em]

\vspace{0.4em}
*   \textbf{Presentation and Artifacts:} The text of the submission contains severe formatting artifacts (the Abstract is repeated 42 times in the initial portion of the text provided). This significantly compromises the clarity and professionalism of the paper and suggests issues with the manuscript preparation.
*   \textbf{Data Access Discrepancy:} The Paper Checklist Q5 answers [Yes] regarding open access to data and code. However, the justification explicitly states "data is publicly available except for the Black Holes dataset." This creates a contradiction; if one dataset is inaccessible, the claim of open access is partially false and should be answered [No] with a qualified explanation of what is available.
*   \textbf{Sample Size in User Study:} The user study involves only 30 participants (10 per group). While the reported p-values are significant ($p < 0.05$), the sample size is quite small for cross-domain claims. Error bars and confidence intervals should be more explicitly detailed in the main text rather than deferring to supplemental.
*   \textbf{Inversion Stability:} The method relies on DDPM inversion (Ref 21). While cited, the paper does not quantify how often inversion fails, particularly for complex scientific simulations like the Black Holes dataset where physics-based features must be preserved. Failure to reconstruct identity could propagate errors.
*   \textbf{Method Specificity:} The core arithmetic ($\Delta c = \mathbb{E}[c_{pos}] - \mathbb{E}[c_{neg}]$) is quite simple. It assumes the mean embedding direction is a reliable proxy for discriminative features, which may not hold in all domains (e.g., where classes are high-dimensional and distributed manifolds).

\vspace{0.4em}
\textbf{\#\# Questions}\\[0.4em]

\vspace{0.4em}
1.  \textbf{Data Availability:} Regarding Checklist Q5, you have marked [Yes] but admit the Black Hole dataset is restricted. What specific conditions allow other researchers to access the Black Hole data? Will the reviewers have access to reproduce these results on this specific dataset?
2.  \textbf{User Study Methodology:} The user study is conducted visually via video transitions (original $\to$ counterfactual). Did you control for the duration of the transition, and was the "test" phase fully double-blind? Furthermore, were the users in the control group (Group 1: static images) asked to guess the evidence, or did they just view static pairs?
3.  \textbf{Embedding Stability:} You utilize the difference between class mean embeddings (Eq. 7). How robust is this direction compared to individual pairing? Did you perform ablation tests on using *single instance* pairs instead of *mean embeddings* to verify if the method relies on the global mean bias?
4.  \textbf{Inversion Consistency:} In Section 4.2, you mention using CFG in inversion to achieve "perfect reconstruction." What is the quantitative failure rate of this inversion process on the scientific datasets compared to AFHQ/CelebA?

\vspace{0.4em}
\textbf{\#\# Limitations}\\[0.4em]

\vspace{0.4em}
Yes, the authors have adequately addressed the limitations and potential negative societal impact in Section 5. They explicitly discuss the "Bias Sensitivity" of the mean-embedding approach and its visual limitations. They also acknowledge in the Checklist Q10 and Q11 that broader impacts (such as privacy in medical imaging sections) are considered, noting that the method is intended for expert training in high-stakes domains. The answer to Limitations is Yes.

\vspace{0.4em}
\textbf{\#\# Numerical Ratings}\\[0.4em]

\vspace{0.4em}
\textbf{\#\#\# Quality: 3}\\[0.4em]
The core contribution is solid and the results are convincing, but the presentation quality (repeated text artifacts in the body) and the discrepancy in Data Access claims reduce the technical robustness of the submission package.

\vspace{0.4em}
\textbf{\#\#\# Clarity: 2}\\[0.4em]
Due to the massive formatting error where the Abstract is repeated 40+ times at the start, the provided text is difficult to read and appears unpolished. However, the core logic (once the artifacts are ignored) is understandable.

\vspace{0.4em}
\textbf{\#\#\# Significance: 4}\\[0.4em]
The problem of visualizing un-articulated features for human learning is highly significant for scientific discovery and education. The application to black hole physics and medical imaging suggests high-impact utility.

\vspace{0.4em}
\textbf{\#\#\# Originality: 3}\\[0.4em]
The combination of diffusion inversion and CLIP manipulation for this specific "teaching" goal is novel, though the components (DDPM inversion, CLIP arithmetic) are established. It is a solid engineering integration rather than a theoretical breakthrough.

\vspace{0.4em}
\textbf{\#\# Final Recommendation}\\[0.4em]

\vspace{0.4em}
\textbf{Overall Score:} \textbf{4} (Borderline Accept)

\vspace{0.4em}
\textbf{Key Reason:} The paper addresses a high-value problem with strong quantitative and qualitative results. However, the formatting errors in the text (repeated abstract) and the ambiguous data access claim regarding the Black Hole dataset require attention. Addressing these issues would elevate this to a Strong Accept.

\vspace{0.4em}
\textbf{Confidence:} \textbf{4} (Confident, but not absolutely certain - I have analyzed the text artifact carefully and treated the core paper content based on the visible methodology and data).

\vspace{0.4em}
\textbf{Core evaluative criterion:} Does this work contribute new, relevant, and impactful knowledge to the NeurIPS community?
Yes. It provides a practical tool for machine teaching in visual domains where standard supervision fails, bridging the gap between generative models and human expertise.
\end{tcolorbox}

\end{document}